\documentclass[10pt,journal,compsoc]{IEEEtran}
\ifCLASSOPTIONcompsoc
  \usepackage[nocompress]{cite}
\else
  \usepackage{cite}
\fi
\ifCLASSINFOpdf
\else
\fi
\usepackage{graphicx}
\usepackage{float}
\usepackage{cite}
\usepackage{amsmath}
\usepackage{setspace}
\usepackage{tabularx,ragged2e}
\usepackage{color}
\usepackage{forest}
\usepackage[T1]{fontenc}
\usepackage[utf8]{inputenc}
\usepackage{tikz}
\usepackage{calc}  
\usepackage{lipsum}
\usepackage{pdflscape}
\usepackage{environ}
\usepackage{multirow}
\usepackage{pifont}
\usepackage{rotating}
\usepackage{array}
\usepackage{amssymb}
\usepackage{hyperref}
\usepackage[tableposition = top, skip=0pt]{caption}
\usepackage[export]{adjustbox}
\usepackage{changes}
\usepackage{scalerel}
\usepackage{cleveref}
\usepackage{subcaption} 
\usepackage{hhline}
\usepackage{algorithm}
\usepackage{amsthm}
\newtheorem{theorem}{Theorem}

\crefformat{section}{\S#2#1#3} 
\crefformat{subsection}{\S#2#1#3}
\crefformat{subsubsection}{\S#2#1#3}

\usetikzlibrary{svg.path}
\definecolor{orcidlogocol}{HTML}{A6CE39}
\tikzset{
  orcidlogo/.pic={
    \fill[orcidlogocol] svg{M256,128c0,70.7-57.3,128-128,128C57.3,256,0,198.7,0,128C0,57.3,57.3,0,128,0C198.7,0,256,57.3,256,128z};
    \fill[white] svg{M86.3,186.2H70.9V79.1h15.4v48.4V186.2z}
                 svg{M108.9,79.1h41.6c39.6,0,57,28.3,57,53.6c0,27.5-21.5,53.6-56.8,53.6h-41.8V79.1z M124.3,172.4h24.5c34.9,0,42.9-26.5,42.9-39.7c0-21.5-13.7-39.7-43.7-39.7h-23.7V172.4z}
                 svg{M88.7,56.8c0,5.5-4.5,10.1-10.1,10.1c-5.6,0-10.1-4.6-10.1-10.1c0-5.6,4.5-10.1,10.1-10.1C84.2,46.7,88.7,51.3,88.7,56.8z};
  }
}
\newcommand\orcidicon[1]{\href{https://orcid.org/#1}{\mbox{\scalerel*{
\begin{tikzpicture}[yscale=-1,transform shape]
\pic{orcidlogo};
\end{tikzpicture}
}{|}}}}
\usepackage{hyperref}

\definechangesauthor[color=purple]{EA}

\definecolor{bcolor}{RGB}{50, 125, 50}

\newcommand{\cmark}{\ding{51}}%
\newcommand{\xmark}{\ding{55}}%
\newcommand{\figref}[1]{Fig. #1}
\newcommand{\equref}[1]{Eq. (#1)}

\graphicspath{{Images/}} 
\newcommand{\blockcomment}[1]{}

\begin{document}
%
\title{One Metric to Measure them All: Localisation Recall Precision (LRP) for Evaluating Visual Detection Tasks}
%
%
%
%

\author{Kemal~Oksuz$^\dagger$ \orcidicon{0000-0002-0066-1517}\ ,
        Baris~Can~Cam \orcidicon{0000-0001-8480-4636}\ ,
        Sinan~Kalkan$^\ddagger$ \orcidicon{0000-0003-0915-5917}\ , and
        Emre~Akbas$^\ddagger$ \orcidicon{0000-0002-3760-6722}\ 
\thanks{All authors are at the Dept. of Computer Engineering, Middle East Technical University (METU), Ankara, Turkey. E-mail:\{kemal.oksuz@metu.edu.tr, can.cam@metu.edu.tr, skalkan@metu.edu.tr, emre@ceng.metu.edu.tr\}\protect}
\thanks{$^\dagger$ Corresponding author.}
\thanks{$^\ddagger$ Equal contribution for senior authorship.}
}

%
%

\markboth{IEEE Transaction of Pattern Analysis and Machine Intelligence}%
{Oksuz \MakeLowercase{\textit{et al.}}: One Metric to Measure them All: Localisation Recall Precision (LRP) for Evaluating Visual Detection Tasks}
%



\IEEEtitleabstractindextext{%
\begin{abstract}
Despite being widely used as a performance measure for visual detection tasks, Average Precision (AP) is limited in (i) reflecting localisation quality, (ii) interpretability and (iii) robustness to the design choices regarding its computation, and its  applicability to outputs without confidence scores. Panoptic Quality (PQ), a measure proposed for evaluating panoptic segmentation (Kirillov et al., 2019), does not suffer from these limitations but is limited to panoptic segmentation. In this paper, we propose Localisation Recall Precision (LRP) Error as the average matching error of a visual detector computed based on both its localisation and classification qualities for a given confidence score threshold. LRP Error, initially proposed only for object detection by Oksuz et al. (2018), does not suffer from the aforementioned limitations and is applicable to all visual detection tasks. We also introduce Optimal LRP (oLRP) Error as the minimum LRP Error obtained over confidence scores to evaluate visual detectors and obtain optimal thresholds for deployment. We provide a detailed comparative analysis of LRP Error with AP and PQ, and use nearly 100 state-of-the-art visual detectors from seven visual detection tasks (i.e. object detection, keypoint detection, instance segmentation, panoptic segmentation, visual relationship detection, zero-shot detection and generalised zero-shot detection) using ten datasets to empirically show that LRP Error provides richer and more discriminative information than its counterparts. Code available at: \url{https://github.com/kemaloksuz/LRP-Error}.
\end{abstract}

}

\maketitle

\IEEEdisplaynontitleabstractindextext

%
\IEEEpeerreviewmaketitle
\section{Introduction}
\label{sec:intro}

Many vision applications require identifying objects and object-related information from images. Such identification can be performed at different levels of detail,  which are addressed by  different detection tasks such as ``object detection'' for identifying labels of objects and boxes bounding them, ``keypoint detection'' for finding keypoints on objects,  ``instance segmentation'' for identifying the classes of objects and localising them with masks, and ``panoptic segmentation'' for classifying both background classes and objects by providing detection ids and labels of pixels in an image.
Accurately evaluating  performances of these methods  is  crucial for developing better solutions. 
\blockcomment{
Today ``average precision’’ (AP), the area under the Precision-Recall (PR) curve, is the de facto standard for evaluating performance on many visual detection tasks and competitions  \cite{COCO,ILSVRC,PASCAL,Objects365,Cityscapes,OpenImages,LVIS}. AP not only enjoys vast acceptance but  also appears to be unchallenged. There has been only a few attempts on developing an alternative to AP  \cite{LRP,PanopticSegmentation,PDQ}.  Despite its popularity, AP has many limitations as we discuss below. 
}

\subsection{Important features for a performance measure}
\label{subsec:EvaluationMeasure}
To facilitate our analysis, we define three important features for performance measures of visual detection methods:  

\textbf{Completeness.} Arguably, three most important  performance aspects that an evaluation measure should take into account in a visual detection task are false positive (FP) rate, false negative (FN) rate and localisation error.  We call a performance measure ``complete'' if it precisely  takes into account all three quantities.  

\textbf{Interpretability.} Interpretability of a performance measure is related to its ability to provide insights on the strengths and weaknesses of the detector being evaluated. To provide such insight, the evaluation measure should ideally comprise interpretable components.  
  
\textbf{Practicality.} Any issue that arises during  practical use of a performance measure diminishes its practicality. This could be, for example, any discrepancy between the well-defined theoretical description of the evaluation measure and its actual application in practice, or any shortcoming that limits the applicability of the measure to certain cases.

\begin{figure*}
    \centerline{
        \includegraphics[width=0.85\textwidth]{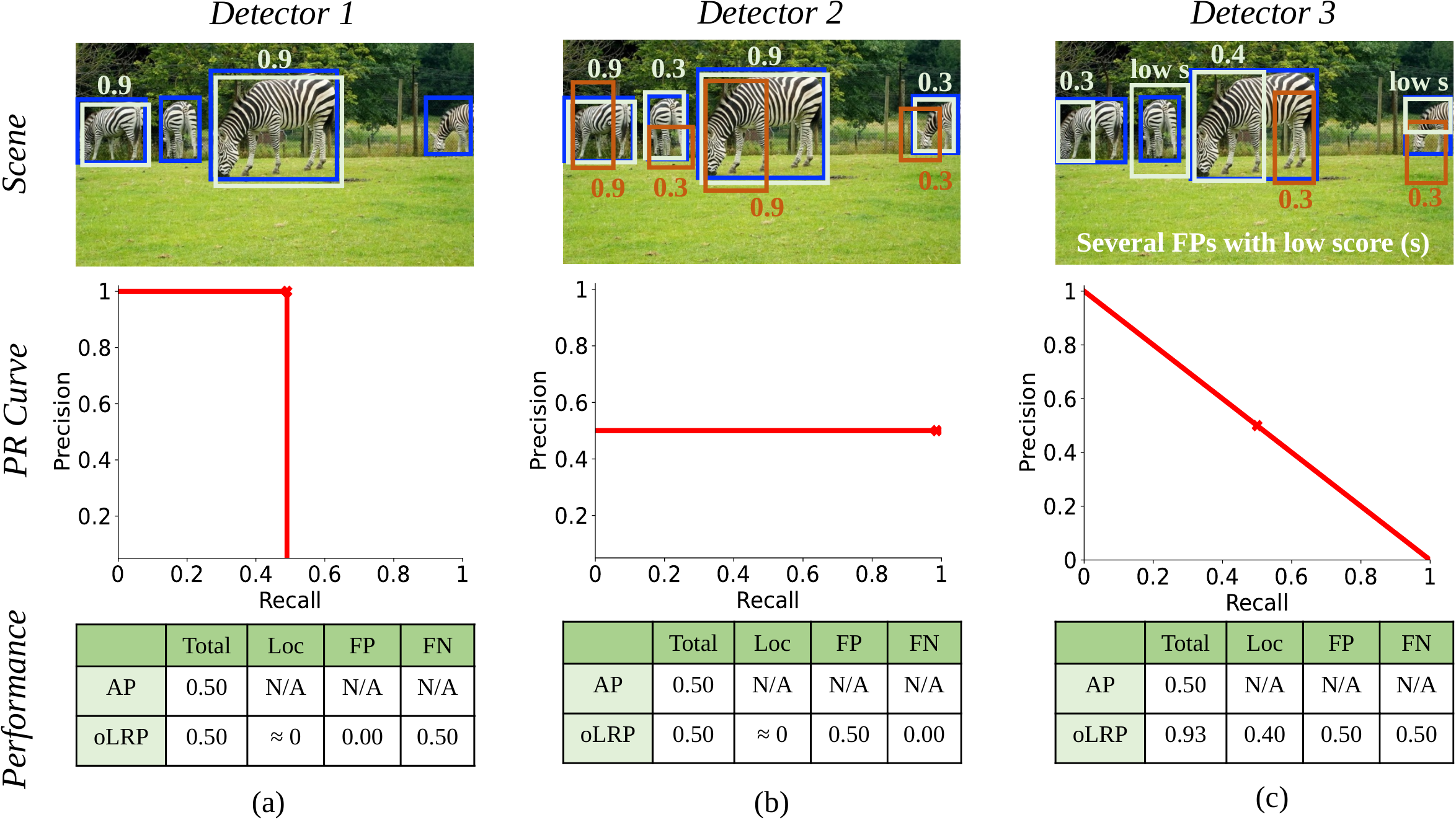}
    }
\caption{Three different object detection results (for an image from COCO \cite{COCO}) with very different PR curves but the same AP. \textbf{First Row}: Blue, white and orange colors denote ground-truth, TPs and FPs respectively. Numbers are confidence scores, $s$, of the detections. \textbf{Second row}: PR curves for the corresponding detections. Red crosses indicate optimal points designated by oLRP Error. \textbf{Third row}: AP and oLRP Error results of the detection results. Comparing LRP Error and AP, (i) In terms of ``completeness'' (Section \ref{subsec:EvaluationMeasure}): While the localisation quality of the TPs in (c) is worse than (a) and (b), AP does not penalize (c) more, while oLRP Error does. (ii) In terms ``interpretability'' (Section \ref{subsec:EvaluationMeasure}): AP is unable to identify the difference among (a), (b) and (c) despite they have very different problems. On the other hand, oLRP Error, as an interpretable metric, demonstrates the strengths and weaknesses of each scenario with its components corresponding to each performance aspect.
}
\label{fig:differentdetectors}
\end{figure*}

\subsection{Overview of Average Precision and Its Limitations}
\label{subsec:APLimitations}
Today ``average precision’’ (AP) is the de facto standard for evaluating performance on many visual detection tasks and competitions  \cite{COCO,ILSVRC,PASCAL,Objects365,Cityscapes,OpenImages,LVIS}. Computing AP for a class involves a set of  detection results with confidence scores and a set of ground-truth items (e.g. bounding boxes in the case of object detection). First, detections are matched to ground-truth items (GT) based on a predefined spatial overlap criterion such as Intersection over Union (IoU)\footnote{While IoU is computed between the boxes for the object detection task, it is computed between the masks for segmentation tasks.} being larger than $0.50$. Each GT can only match one detection and if there are multiple detections that satisfy the overlap criterion, the one with the highest confidence score is matched. A detection that is matched to a GT is counted as a true positive (TP). Unmatched detections are FPs and unmatched GTs are FNs. Given a specific confidence threshold $s$, detections with a lower confidence score than $s$ are discarded, and numbers of TP, FP, FN (denoted by $\mathrm{N_{TP}}$, $\mathrm{N_{FP}}$ and $\mathrm{N_{FN}}$ respectively) are calculated with the remaining detections. By systematically changing $s$, we compute precision (i.e. $\mathrm{N_{TP}}/(\mathrm{N_{TP}}+\mathrm{N_{FP}})$) and recall (i.e. $\mathrm{N_{TP}}/(\mathrm{N_{TP}}+\mathrm{N_{FN}})$) pairs to obtain a Precision-Recall (PR) curve. The area under the PR curve determines the AP for a class and the detector's  performance over all classes is obtained simply by averaging per-class AP values.

AP not only enjoys vast acceptance but  also appears to be unchallenged. There has been only a few attempts on developing an alternative to AP  \cite{LRP,PanopticSegmentation,PDQ}. Despite its popularity, AP has many limitations as we discuss below based on our important features. 
 
\textbf{Completeness.} Localisation quality is only loosely taken into account in AP. Detections that meet a certain localisation  criterion (e.g., IoU over $0.50$) are treated equally regardless  of their actual localisation quality. Further increase in  localisation quality, for example, increasing the IoU of a TP detection, does not change AP (compare Detector 1 or 2 with 3 in \figref{\ref{fig:differentdetectors}}). 

\textbf{Interpretability.} The AP score itself does not provide any insight in terms of the important performance aspects, namely, FP rate, FN rate and localisation error. One needs to inspect the PR curve and make additional measurements (e.g. average-recall (AR) or some kind of localisation quality) in order to comment on the weaknesses or strengths of a detector in terms of these aspects. The PR curves and their APs in  \figref{\ref{fig:differentdetectors}} provide a compelling example in which all detectors have the same AP but different weaknesses. 
\textbf{Practicality.}  We identify three major issues related to the practical use of AP: (i)  Evaluating hard-prediction tasks, i.e. tasks that involve outputs without confidence scores, such as panoptic segmentation \cite{PanopticSegmentation}, is not trivial. (ii) AP cannot be used for model selection, and (iii) design choices in the computation of AP (e.g. interpolating the PR curve or approximating the area under PR curve) limit its reliability.  

\blockcomment{
\textbf{Practicality.}  We identify three major issues related to the practical use of AP: 

(i) Using AP to evaluate hard-prediction tasks, i.e. tasks that involve outputs without confidence scores, such as panoptic segmentation \cite{PanopticSegmentation}, is problematic because hard predictions yield only a single point on the PR curve. Assumptions are needed to compute the area under the PR curve consisting of just one point. 

(ii) AP cannot be used for model selection. For example, when a detector is to be deployed for a certain problem, an optimal detection threshold is needed. AP does not offer any help in finding such optimal thresholds.

(iii) Lastly, one needs to interpolate the PR curve before computing AP, which, as we will show, is a problem with classes with few examples.

}

We provide a detailed discussion on the limitations of AP regarding each important feature in Section \ref{sec:AveragePrecision} and provide an empirical analysis in Section \ref{subsec:SoftPredictionExperiments}.

\subsection{Localisation Recall Precision (LRP) Error} 
LRP Error is a new performance metric for visual detection tasks. We originally proposed LRP Error  \cite{LRP}   for the object detection task. It can be compactly written as:
\begin{align}
\label{eq:LRPdefintro}
\frac{1}{\mathrm{N_{TP}}+\mathrm{N_{FP}} + \mathrm{N_{FN}}} \left( \sum \limits_{i=1}^{\mathrm{N_{TP}}} \mathcal{E}_{Loc}(i)+\mathrm{N_{FP}} + \mathrm{N_{FN}} \right),
\end{align}
where  $\mathrm{N_{TP}}$, $\mathrm{N_{FP}}$ and $\mathrm{N_{FN}}$ are identified as done in AP (Section \ref{subsec:APLimitations}) and $\mathcal{E}_{Loc}(i)$ is the normalised (i.e. between $0$ and $1$) localisation error for the $i^\mathrm{th}$ TP. We  showed that LRP Error can equivalently be expressed as a weighted combination of the average localisation qualities of TPs, precision error (1-precision) and recall error (1-recall) -- the three components which we coin as the components of LRP Error. With this definition and extensions presented in this paper, LRP Error addresses all limitations of AP: (i) LRP Error considers precisely all three important performance aspects, thus it is complete (\textit{cf.} AP and LRP Error in \figref{\ref{fig:differentdetectors}}). (ii) LRP Error is  interpretable through its components, hence it provides insights regarding each performance aspect (\textit{cf.} AP and LRP Error in \figref{\ref{fig:differentdetectors}}). (iii) LRP Error is not limited by the practicality issues of AP.
Also,  LRP Error is a metric, for which, however, we do not demonstrate any theoretical or practical benefits. 


\subsection{Other alternatives to AP} 
While AP is still de facto performance measure for many visual detection tasks, recently proposed visual detection tasks have  preferred not employing AP, but instead introduced novel performance measures:

\textbf{Panoptic Quality (PQ):} Panoptic segmentation task \cite{PanopticSegmentation} requires the background classes to be labeled and localised by masks in addition to the objects. Since this task is a combination of the instance segmentation and semantic segmentation tasks, AP can be used to  evaluate performance. However, arguing the inconsistency between machines and humans in terms of perceiving the objects due to the confidence scores in the outputs, Kirillov et al. \cite{PanopticSegmentation} preferred to discard these scores for evaluation and proposed PQ as a new performance measure to evaluate the results of the panoptic segmentation task. Similar to LRP Error \cite{LRP}, PQ combines all important performance aspects of visual detection, however, its extension to other visual detection tasks has not been explored. We provide a detailed analysis on PQ in Section \ref{sec:PanopticQuality} and discuss empirical results in Section \ref{subsec:HardPredictionExperiments}. Note that PQ was proposed later than LRP Error. 

\textbf{Probability-based Detection Quality (PDQ):} Unlike conventional object detection, probabilistic object detection (POD) \cite{PDQ} takes into account the spatial and semantic uncertainties of the objects, and accordingly for each detection, requires (i) a probability distribution over the class labels (i.e. instead of a single confidence score as in soft predictions) and (ii) a probabilistic bounding box represented by Gaussian distributions. Similar to Kirillov et al. \cite{PanopticSegmentation}, Hall et al. \cite{PDQ} also did not prefer an AP-based performance measure for POD, instead proposed a new performance measure called PDQ to evaluate probabilistic outputs. In this paper, we limit our scope to deterministic approaches to visual detection tasks. Therefore, we do not delve into a detailed discussion on PDQ as we do for AP and PQ; instead, we provide a guidance on how LRP Error can be extended for different visual detection tasks in Section \ref{subsec:UsageExtensions}.


\begin{figure}
        \captionsetup[subfigure]{}
        \centering
        \begin{subfigure}[b]{0.2\textwidth}
        \includegraphics[width=\textwidth]{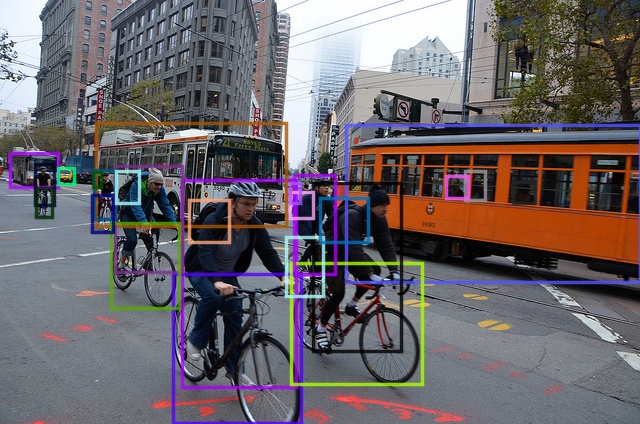}
        \caption{\footnotesize{Object Detection}}
        \end{subfigure}
        \begin{subfigure}[b]{0.2\textwidth}
        \includegraphics[width=\textwidth]{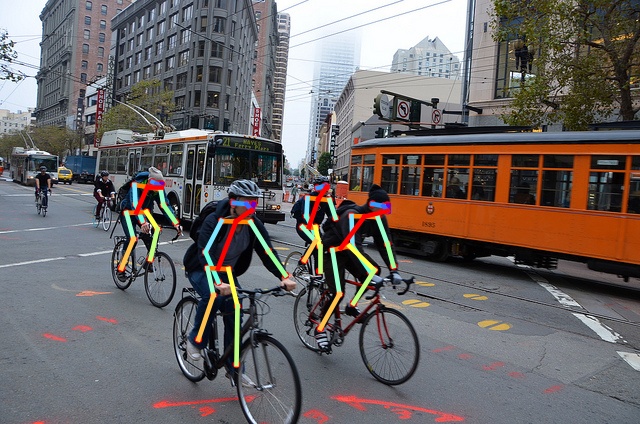}
        \caption{\footnotesize{Keypoint Detection}}
        \end{subfigure}
        
        \begin{subfigure}[b]{0.2\textwidth}
        \includegraphics[width=\textwidth]{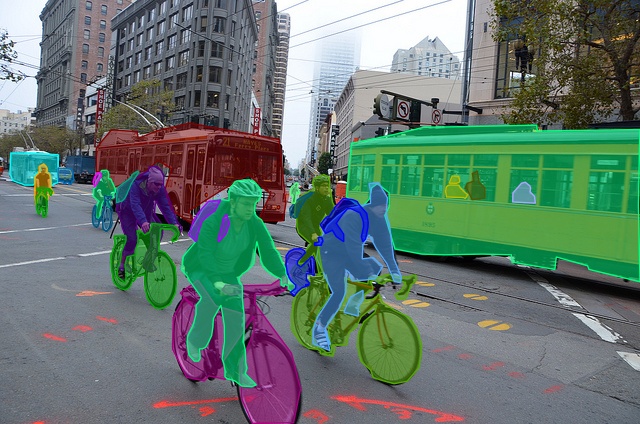}
        \caption{\footnotesize{Instance Segmentation}}
        \end{subfigure}
        \begin{subfigure}[b]{0.2\textwidth}
        \includegraphics[width=\textwidth]{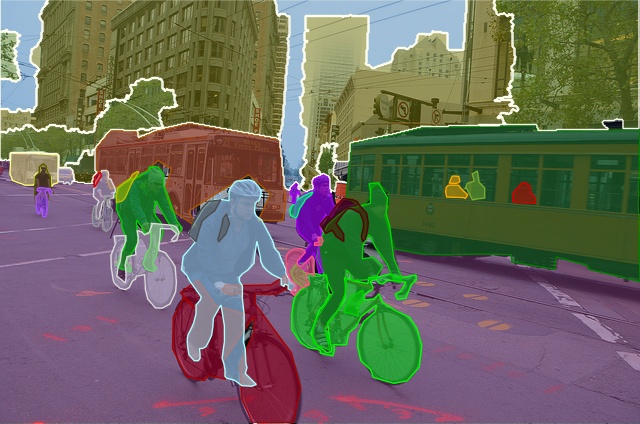}
        \caption{\footnotesize{Panoptic Segmentation}}
        \end{subfigure}
        \caption{Example visual detection tasks considered in the paper. Keypoint detection is illustrated for the `person' class only. Image: COCO \cite{COCO}. Ground truth plots: Detectron2 \cite{Detectron2}.}
        \label{fig:tasks}
\end{figure}

\subsection{Contributions of the Paper}
Our contributions are as follows:

\noindent (1) We thoroughly analyse AP and PQ.\\

\noindent (2) We present LRP Error and describe its use for  \textit{all} visual detection tasks (\figref{\ref{fig:tasks}}). LRP Error can evaluate all visual detection tasks with soft or hard predictions\footnote{Soft predictions: outputs with class labels and confidence scores, such as in   object detection; hard predictions:  outputs with class labels only, such as in panoptic segmentation.} by alleviating the drawbacks of AP and PQ. In particular, we empirically present the usage of LRP Error on seven important visual detection tasks, namely, object detection, keypoint detection, instance segmentation, panoptic segmentation, visual relationship detection, zero-shot detection (ZSD), generalised ZSD and discuss its potential extensions to other tasks.

\noindent (3) While LRP Error can directly be used for hard predictions, to evaluate soft predictions we propose Optimal LRP Error (oLRP) Error as the minimum achievable LRP Error over the confidence scores.

\noindent (4) We show that LRP Error is an upper bound for the error versions of precision, recall and PQ (Section \ref{sec:Comparison}). Hence, minimizing LRP Error is guaranteed to minimize the other measures. 

\noindent (5) We show that the performances of visual detectors are sensitive to thresholding, and based on oLRP Error, we propose ``LRP-Optimal Threshold'' to reduce the number of detections in an optimal manner.

\textbf{Main Results:} We compare LRP Error with its counterparts on  $\sim$100 state-of-the-art visual detectors and provide examples, observations and various analyses with LRP and oLRP Errors at the detector- and class-level to present its evaluation capabilities. We show that: (i) LRP Error can consistently evaluate and unify the evaluation of all visual detection tasks in any desired output type (i.e. soft predictions or hard predictions), (ii) visual detectors need to be thresholded in a class-specific manner for optimal performance, and (iii) oLRP Error provides class-specific optimal thresholds  by considering all performance aspects.

\textbf{Comparison with Our Previous Work:} The current paper extends our previous work \cite{LRP} to \textit{all} visual detection tasks. Hence, besides object detection, here, we present the usage of LRP Error on other six common visual detection tasks. While extending LRP Error for other detection tasks, we present how LRP Error can be employed to evaluate hard predictions, soft predictions and its potential extensions. Moreover, the experimental analysis is performed from scratch to cover all seven detection tasks, to evaluate more recent methods, to use LRP Error for evaluating datasets with different characteristics and for tuning hyperparameters, to dwell more on the analysis of oLRP Error and to investigate computational latency. 

\subsection{Outline of the Paper}
The paper is organized as follows. Section \ref{sec:Related} presents the related work. Sections \ref{sec:AveragePrecision} and \ref{sec:PanopticQuality} present a thorough analysis of AP and PQ respectively. Section \ref{section:LRP} defines the LRP Error and oLRP Error. Section \ref{sec:Comparison} compares LRP Error with AP and PQ. Section \ref{section:experiments} presents several experiments to analyse LRP Error. Finally, Section \ref{section:conclusions} concludes the paper.

\section{Related Work}
\label{sec:Related}


\textbf{Evaluation in visual detection tasks.} As discussed in Section \ref{sec:intro}, except for the panoptic segmentation task which uses PQ \cite{PanopticSegmentation}, the performances of visual detection methods are conventionally evaluated using AP. Sections \ref{sec:AveragePrecision} and \ref{sec:PanopticQuality} discuss and present an analysis of these performance measures. 

Another measure, PDQ \cite{PDQ} has recently been proposed for evaluating the probabilistic object detection task, where the label of a detection is represented by a discrete probability distribution over classes and the bounding boxes are encoded by Gaussian distributions. To compute PDQ, first, pairwise PDQ (pPDQ) score is computed over all detection-ground truth pairs and the optimal matchings are identified following the Hungarian Algorithm \cite{Hungarian}. Then, determining TPs, FPs and FNs, and using the pPDQs of optimal matchings, PDQ score of a detection set can be computed by normalizing the sum of these pPDQs by the total number of TPs, FPs and FNs. To evaluate each pair, pPDQ combines localisation and classification performances by its spatial quality and label quality components. The spatial quality evaluates a pair in a pixel-based manner (i.e. not box-based) by exploiting the segmentation mask of the ground truth. And, the label quality of a pair is the probability of the ground truth label in the label distribution of the detection. Therefore, computing PDQ requires (i) segmentation masks which are normally not provided for the conventional object detection task, and (ii) the outputs to be in the described probabilistic form. In this paper, we show that LRP Error can be used for all common visual detection tasks having the conventional deterministic representation, and provide a guideline on how it can be employed by other tasks.

\textbf{Analysis tools for visual detection tasks.} Over the years, diagnostic tools have been proposed for providing detailed insights on the performances of detectors. For example, Hoiem et al. \cite{Analyzer} selected the top-k FPs based on  confidence scores and analysed them in terms of common error types (i.e. localisation error, confusion with similar objects, confusion with other objects and confusion with background). However, the tool of Hoiem et al. \cite{Analyzer} requires additional analysis for FNs. Another toolkit, the COCO toolkit \cite{COCO}, is based on this analysis tool, but instead plots the considered error types on PR curves progressively to present how much AP difference is accounted by each error type. Recently, Bolya et al. \cite{TIDE} showed that the COCO toolkit can yield inconsistent outputs when the order of progressive contribution of the error types to the AP is interchanged. Moreover, this analysis by the COCO toolkit yields superimposed numerous PR curves which are time-consuming to examine and hard to digest. Based on these observations, Bolya et al. \cite{TIDE} proposed TIDE, a toolkit addressing the limitations of the previous analysis tools. TIDE introduces six different error types, each of which is summarized by a single score in the analysis result. Although such tools are useful for providing detailed insights on the types of errors detectors are making, they are not performance measures, and as a result they do not yield a single performance value as the detection performance.

\textbf{Point multi-target tracking performance metrics.} The evaluation of the detection tasks is very similar to that of multi-target tracking in that there are multiple instances of objects/targets to detect, and the localisation, FP and FN errors are common criteria for success. Currently, component-based performance metrics are the accepted way of evaluating point multi-target tracking methods. One of the first metric to combine the localisation and cardinality (including both FP and FN) errors is the Optimal Subpattern Assignment (OSPA) \cite{OSPA}. Among the successors of OSPA, our LRP Error \cite{LRP} was inspired by the Deficiency Aware Subpattern Assignment metric \cite{DASA}, which combines the three important performance aspects.


\textbf{Summary.} We observe  that, with similar error definitions, point multi-target tracking literature utilizes component-based performance metrics commonly, which has not been explored thoroughly in the visual detection literature. While a recent attempt, Panoptic Quality, is an example of that kind, it is limited to panoptic segmentation (Section \ref{sec:PanopticQuality}). The analysis tools also aim to provide insights on the detector, however, a single performance value for the detection performance is not provided by these methods. In this paper, we propose a single metric that  ensures important features (i.e. completeness, interpretability and practicality) while evaluating the performance of methods for visual detection tasks. We also show that our metric, considering all performance aspects, is able to pinpoint a class-wise optimal threshold for the visual detectors, from which several applications can benefit in practice.   


\section{An Analysis of Average Precision}
\label{sec:AveragePrecision}
\label{subsec:APAnalysis}
In the following, we provide an analysis of AP by discussing its limitations introduced in Section \ref{subsec:APLimitations} in detail. Later, Section \ref{subsec:SoftPredictionExperiments} provides  empirical analysis on these limitations. 

\textbf{Completeness.} \textit{AP does not explicitly evaluate localisation performance (\figref{\ref{fig:differentdetectors}}) and therefore, violates completeness.} To circumvent this issue, researchers typically use the following methods, neither of which ensures completeness: 
\begin{itemize}
    \item \textit{Quantitatively using COCO-style AP ($\mathrm{AP^C}$)\footnote{To include the localisation quality,  the COCO-style AP, $\mathrm{AP^C}$, computes 10 $\mathrm{AP_\tau}$ where the TP validation threshold, $\tau$, is increased between 0.50 and 0.95 with a step size of 0.05, and these 10 $\mathrm{AP_\tau}$ values are averaged.} or $\mathrm{AP_\tau}$ with large $\tau$:}  AP variants do not include the precise localisation quality of a detection, hence the contribution of the localisation performance to these AP variants is always loose. Specifically, regardless of the  threshold $\tau$, AP does not discriminate between a detection with perfect localisation (e.g. IoU=1) and  a detection with localisation quality barely exceeding $\tau$; hence,  the localisation quality of a TP detection is not precisely quantified. As a result, the methods  specifically proposed to improve the localisation quality \cite{KLLoss,OffsetBin,aLRPLoss,CascadeRCNN,GIoULoss} have been struggling to present their contributions quantitatively. While some of them \cite{CascadeRCNN,GIoULoss} present only $\mathrm{AP^C}$, $\mathrm{AP_{75}}$ and $\mathrm{AP_{50}}$, others \cite{KLLoss,OffsetBin,aLRPLoss,DynamicRCNN} additionally resort to APs with  larger $\tau$ values such as $\mathrm{AP_{80}}$ or $\mathrm{AP_{90}}$. However, as we demonstrate in Section \ref{subsec:Localisation}, all of these AP variants may fail to appropriately compare methods in terms of localisation quality.
    
    \item \textit{Presenting qualitative examples \cite{SSD,FasterRCNN,RFCN,DetectToTrack,AssociationLSTM,APLoss}:} In this case, note that it is very likely for the selected examples  to be very limited and biased. 
\end{itemize}

\textbf{Interpretability.} \textit{The resulting AP value does not provide any insight on the strengths or weaknesses of the detector.} As illustrated in \figref{\ref{fig:differentdetectors}}, different detectors may yield different PR curves, highlighting different types of performance issues. However, being the area under the PR curve, AP fails to distinguish between the underlying issues of different detectors. This is mainly because both precision and recall performances of a detector are vaguely combined into a single performance value as an AP value. Besides, interpreting the COCO-style AP, $\mathrm{AP^C}$, is more difficult since the localisation quality is also integrated in an indirect and loose manner, resulting in an ambiguous contribution of important performance aspects, where it is not clear how much each  aspect affects the resulting single performance value. As a result, AP variants fail to satisfy interpretability-related criteria. For example, (i) in contrast to what Bernardin and Stiefelhagen \cite{Bernardin2008} expect from useful metrics, AP is not clear and easily understandable, (ii) AP does not have a meaningful physical interpretation as opposed to the criteria suggested by Schuhmacher et al. \cite{OSPA}, and (iii) Bolya et al. \cite{TIDE} criticized AP for not being able to isolate error types. To alleviate this, the COCO toolkit \cite{COCO} can output PR curves with an error analysis, which requires manual inspection of several superimposed PR curves in order to understand the strengths and weaknesses. This is, however, time-consuming and impractical with large number of classes such as the LVIS \cite{LVIS} with $\sim 1000$ classes (also see the discussion on analysis tools in Section \ref{sec:Related}). 



\textbf{Practicality.} One can also face some practical challenges while employing AP in certain important use-cases:
\begin{itemize}
    \item \textit{Evaluation of hard predictions with AP, though possible, is problematic.} A hard prediction (i.e. an output without confidence score) corresponds to a single point on the PR space, hence yields a step PR curve resulting in $\mathrm{AP}=\mathrm{Precision} \times \mathrm{Recall}$. However, AP intends to prioritize and rank the detections with respect to their confidence scores, which are not included in hard predictions. As a result, in a recent study, Kirillov et al. \cite{PanopticSegmentation} proposed a new performance measure called Panoptic Quality for the panoptic segmentation task (e.g. instead of using $\mathrm{Precision} \times \mathrm{Recall}$ as AP), which can evaluate  hard predictions. Therefore, the usage of AP on hard predictions does not fit into its ranking-based definition.
    \item \textit{AP does not offer an optimal threshold for a detector.} Being defined as the area under the PR curve, any thresholding on  detections decreases this area. Hence, performance with respect to AP increases, when the confidence score threshold approaches to $0$ (i.e. the case of ``no-thresholding''). As a result, it is not clear how the large number of object hypotheses can be reduced properly with AP when a visual detector is to be deployed in a practical application.
    \item \textit{AP is sensitive to design choices, degrading its reliability}. Regarding this sensitivity, we discuss three points. 
    
        (i)\textit{Interpolating the PR curve:}  The procedure to obtain the PR curve (Section \ref{subsec:APLimitations}) usually results in a non-monotonic curve, that is, the precision may go up and down as recall is increased. Conventionally \cite{COCO,PASCAL,LVIS,mmdetection,Detectron2}, in order to decrease these wiggles,  the PR curve is interpolated as follows: denoting the precision at a recall $r_i$ before and after interpolation by $\mathrm{p}(r_i)$ and $\mathrm{\hat{p}}(r_i)$ respectively, $\mathrm{\hat{p}}(r_i) = \max \limits_{r_j>r_i} \mathrm{p}(r_j)$ \cite{PASCAL}. However, few examples yield a sparse set of PR pairs, and in this case interpolating the line segments spanning  larger intervals will change the AUC more, which can especially have an effect for long-tailed datasets such as LVIS with a median of only 9 instances per class in the COCO 2017 val subset (5k images) which Gupta et al. \cite{LVIS} used for analysis. To conclude, using AP for classes with few examples is  problematic owing to interpolating the PR curve.
    
     
     (ii) \textit{Approximating the area under the PR curve:} While some of the datasets calculate the area under the PR curve (e.g. standard Pascal evaluation \cite{PASCAL}), some approximate this area, e.g. in COCO \cite{COCO}  the recall axis is divided by 101 evenly-spaced points, on which precision values are averaged.  We observed that this design choice can have a significant effect on AP. 
     
     (iii) \textit{Limiting the number of detections:} In order to compare the methods equally, the number of detections to be considered during performance evaluation is usually limited (e.g. COCO \cite{COCO} allows 100 detections from each image). As a practical drawback of AP, Dave et al. \cite{devil} showed that imposing a limit based on images demotes the classes with less examples when AP is used to evaluate long-tailed datasets, and instead, introduced limiting the number of detections  per class, coined as \textit{fixed AP}. However, still, fixed AP is sensitive to the number of detections per class. 
\end{itemize}{}
\section{Panoptic Quality}
\label{sec:PanopticQuality}
Here, we first provide a definition of PQ and then analyse it based on our important features (Section \ref{subsec:EvaluationMeasure}).


\subsection{Definition of PQ}
The PQ measure is proposed to evaluate the performance of panoptic segmentation methods \cite{PanopticSegmentation}. Given hard predictions (i.e. outputs without confidence scores), first, the detections are labelled as TP, FP and FN using an $\mathrm{IoU}$-based criterion, and then the numbers of TPs ($\mathrm{N_{TP}}$), FPs ($\mathrm{N_{FP}}$), FNs ($\mathrm{N_{FN}}$) and the localisation quality of TP detections in terms of IoU (i.e. $\mathrm{IoU}(g_i, d_{g_i})$ is the IoU between the masks of the ground truth $g_i$ and its associated detection, $d_{g_i}$, with $g_i$) are computed. Based on these quantities, PQ between a ground truth set $\mathcal{G}$ and a detection set $\mathcal{D}$ is defined as:
\begin{align}
\label{eq:PQ}
\mathrm{PQ}(\mathcal{G},\mathcal{D}) = \frac{1 }{{\mathrm{N_{TP}}+ \frac{1}{2} \mathrm{N_{FP}} + \mathrm{\frac{1}{2} N_{FN}}}} \left( \sum \limits_{i=1}^{\mathrm{N_{TP}}} \mathrm{IoU}(g_i, d_{g_i}) \right).
\end{align}
PQ is a ``higher is better'' measure with a range between $0$ and $1$. To provide more insight on the segmentation performance, PQ is split into two components: (i) Segmentation Quality (SQ), defined as the average IoU of the TPs, is a measure of the localisation performance; (ii) Recognition Quality (RQ), as a measure of classification performance based on  the F-measure. Using SQ and RQ, PQ can equally be expressed as:   $\mathrm{PQ}(\mathcal{G},\mathcal{D}) =\mathrm{SQ}(\mathcal{G},\mathcal{D}) \mathrm{RQ}(\mathcal{G},\mathcal{D}) $.

\begin{table}
\centering
\caption{A comparison of LRP Error and PQ for the detectors (i.e. Detector 1 and Detector 2) in scenarios (a) and (b) in \figref{\ref{fig:differentdetectors}} (Since PQ and LRP Error do not need confidence scores, scores are simply ignored for computing PQ and LRP Error in these scenarios.) (a) PQ cannot identify the difference between these two scenarios, and yields exactly the same results for both of its components (i.e. SQ and RQ). (b) With a component for each performance aspect, LRP Error can discriminate between these results using FP and FN components. While for PQ and components higher is better; for LRP Error, lower is better.}
\includegraphics[width=0.5\textwidth]{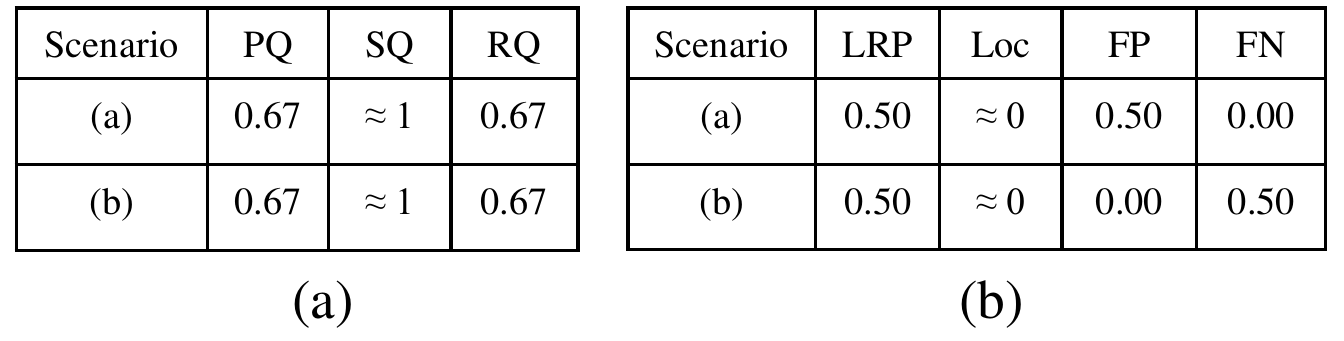}
\label{tab:PQComponents}
\end{table}
\blockcomment{
\begin{table}
\centering
\caption{An example depicting PQ yield nonmonotonic results along an arbitrary line in its input space. Note that LRP Error has a linear change. $\epsilon$ is a small value to ensure that the detections are TP (i.e. $\mathrm{lq(\cdot,\cdot)} > 0.50$). While for PQ higher is better; for LRP Error, lower is better. \label{tab:nonmonotonic}}
\setlength{\tabcolsep}{0.25em}
\renewcommand{\arraystretch}{0.6}\footnotesize
    \begin{tabular}{|c|c|c|c|c|c|c|}
         \hline
         mean lq()&0.50+$\epsilon$&0.60&0.70&0.80&0.90&1.00\\ \hline
         $\mathrm{N_{TP}}$&20&18&16&14&12&10\\ \hline
         $\mathrm{N_{FP}}+\mathrm{N_{FN}}$&0&4&8&12&16&20\\ \hline \hline
         PQ&0.50+$\epsilon$&0.54&0.56&0.56&0.54&0.50\\ \hline
         LRP&1.00-$\epsilon$&0.84&0.73&0.68&0.66&0.67\\ \hline
    \end{tabular}
\end{table}
}
\subsection{An Analysis of PQ}
\label{subsec:PQAnalysis}
We analyse PQ based on the important features (Section \ref{subsec:EvaluationMeasure}): 

\textbf{Completeness.} In contrast to AP, PQ precisely takes into account all performance aspects (i.e. FP rate, FN rate and localisation error - see ``completeness'' in Section \ref{subsec:EvaluationMeasure}) that are critical for visual detectors. 

\textbf{Interpretability.} Another advantage of PQ compared to AP is that PQ is more interpretable than AP owing to its SQ and RQ components. On the other hand, the RQ component, essentially the F-measure, is unable to distinguish  different recall and precision performances (Table \ref{tab:PQComponents}(a)) because both precision and recall errors are combined into a single component, RQ (i.e. the error types are not isolated \cite{TIDE}). Therefore, overall, PQ is superior than AP in terms of interpretability, but having a component for each performance aspect would provide more useful insights.  

\begin{figure*}
\centering
\includegraphics[width=0.72\textwidth]{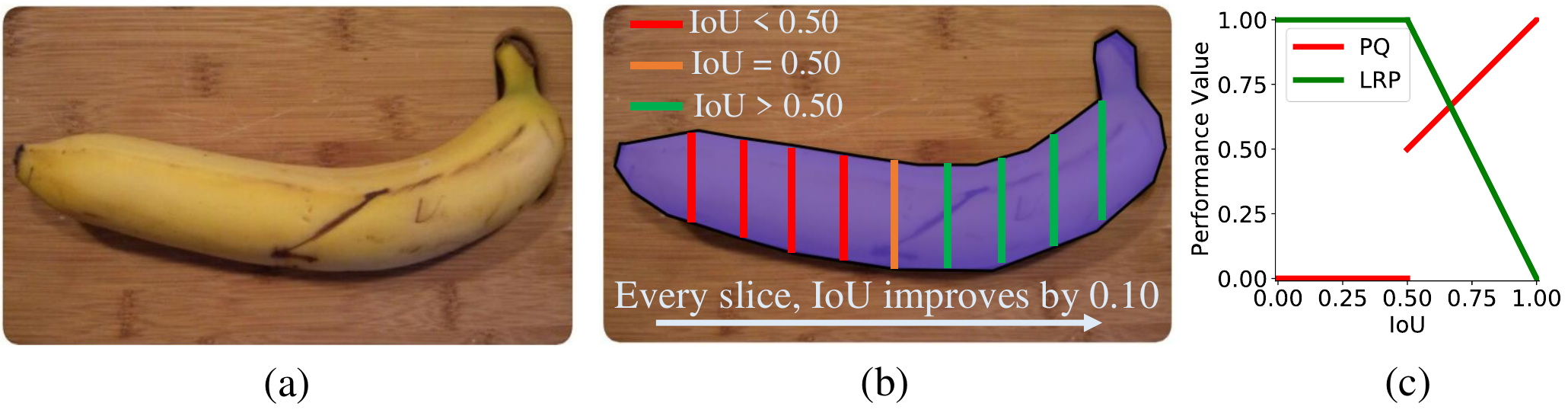}
\caption{An illustration that shows how a transition from a FP to TP is handled differently by PQ and LRP Error. (a) An example image from COCO \cite{COCO}. (b) Segmentation masks for the ground truth with different $\mathrm{IoU}$. The ground truth is split into 10  approximately equal slices. Orange line is the threshold where the detection is still a FP, hence a single pixel added makes the detection a TP. (c) How PQ and LRP Error changes for different $\mathrm{IoU}$. While LRP Error is zeroth-order continuous, PQ is a discontinuous function and allows large jumps.}
\label{fig:PQComparison}
\end{figure*}

\textbf{Practicality.}  Being designed for panoptic segmentation, we limit our discussion of PQ to panoptic segmentation, and omit its generalizability over all detection tasks:
\begin{itemize}
    \item \textit{Kirillov et al. \cite{PanopticSegmentation} did not discuss how PQ can evaluate and threshold soft predictions (i.e. the outputs with confidence scores).} Kirillov et al.  preferred hard predictions for panoptic segmentation  to eliminate the inconsistency between machines and humans in terms of perceiving the objects. Accordingly, proposed for panoptic segmentation,  PQ is designed to evaluate hard predictions, and its possible extensions on soft predictions (and also other visual detection tasks) are not discussed and analysed by its authors. 
    \item \textit{PQ overpromotes classification performance compared to localisation performance inconsistently.} We observe the following for PQ: (i) \figref{\ref{fig:PQComparison}} illustrates how small shifts, induced by a TP, can cause large changes in PQ. Due to this promotion of a TP via a jump in the performance value, the effect of the localisation quality is decreased since the localisation quality can contribute between $\mathrm{PQ} \in [0.50, 1.00]$ (\figref{\ref{fig:PQComparison}}), (ii) While one can prefer classification error to have a larger effect on the overall performance, the formulation of PQ is inconsistent in terms of how localisation and classification performances are combined. In order to provide a comparative analysis with our performance metric, we discuss this inconsistency in Section \ref{sec:Comparison}. (iii) This inconsistent combination also makes PQ violate the triangle inequality property of metricity (see Appendix \ref{sec:PQMetricity} for a proof).
\end{itemize}




\section{Localisation Recall Precision Error}
\label{section:LRP}
In this section, we describe and analyse the Localisation Recall Precision (LRP) Error (Sections \ref{subsec:LRP} and \ref{subsec:LRPAnalysis}) and present  Optimal LRP (oLRP) Error as the extension of LRP Error for  evaluating and thresholding soft-prediction-based  visual object detectors (Section \ref{subsec:oLRP}). We also present a guideline for other potential extensions of LRP Error  (Section \ref{subsec:UsageExtensions}).

\subsection{LRP Error: The Performance Metric}
\label{subsec:LRP}
\noindent \textbf{Definition:} LRP Error is an error metric that considers both localisation and classification. To compute $\mathrm{LRP}(\mathcal{G},\mathcal{D})$ given a set of detections ($\mathcal{D}$ - each $d_{i} \in \mathcal{D}$ is a tuple of class-label and location information), and a set of ground truth items  ($\mathcal{G}$), first, the detections are assigned to  ground truth items based on the matching criterion (e.g. IoU) defined for  the corresponding visual detection task. Once the assignments are made, the following values are computed: (i) $\mathrm{N_{TP}}$, the number of true positives; (ii) $\mathrm{N_{FP}}$, the number of false positives; (iii) $\mathrm{N_{FN}}$, the number of false negatives and (iv) the localisation qualities of TP detections, i.e. $\mathrm{lq}(g_i, d_{g_i})$ for all $d_{g_i}$ where $d_{g_i}$ is a TP matching with ground truth $g_i$. 
Using these quantities, the LRP Error is defined as:
\begin{align}
\label{eq:LRPdefcompact}
\mathrm{LRP}(\mathcal{G},\mathcal{D}):= \frac{1}{Z} \left( \sum \limits_{i=1}^{\mathrm{N_{TP}}}  \frac{1-\mathrm{lq}(g_i, d_{g_i})}{1-\tau}+\mathrm{N_{FP}} + \mathrm{N_{FN}} \right),
\end{align}
\normalsize
where $Z=\mathrm{N_{TP}}+\mathrm{N_{FP}} + \mathrm{N_{FN}}$ is the normalisation constant and $\tau$ is the TP validation threshold ($\tau=0.50$ unless otherwise stated). \equref{\ref{eq:LRPdefcompact}} can be interpreted as  the ``average matching error'', where the term in parentheses is the ``total matching error'', and  $Z$ is the ``maximum possible value of the total matching error''. A TP contributes to the total matching error by its localization error normalized by $1-\tau$ to ensure that the value is in interval [0,1] and LRP Error is a zeroth-order continuous function (\figref{\ref{fig:PQComparison}}(c)). Each FP or FN contributes to the total matching error by 1. Finally, normalisation by $Z$ ensures  $\mathrm{LRP}(\mathcal{G},\mathcal{D}) \in [0,1]$. We prove LRP Error is a metric if $1-\mathrm{lq}(\cdot, \cdot)$ is a metric (Appendix \ref{section:appendix}).  

\noindent \textbf{Components:} In order to provide additional information on the characteristics of the detector, LRP Error can be equivalently defined in a weighted form as:
\begin{align}
\label{eq:LRPdef}
\mathrm{LRP}(\mathcal{G},\mathcal{D}) :=& \frac{1}{Z} \left( \mathrm{w_{Loc}} \mathrm{LRP_{Loc}}(\mathcal{G},\mathcal{D})+ \mathrm{w_{FP}} \mathrm{LRP_{FP}}(\mathcal{G},\mathcal{D}) \right.  \nonumber \\
&\left. + \mathrm{w_{FN}} \mathrm{LRP_{FN}}(\mathcal{G},\mathcal{D}) \right),
\end{align}
with the weights $w_{Loc}=\frac{\mathrm{N_{TP}}}{1-\tau}$, $w_{FP}=|\mathcal{D}|$, and $w_{FN}=|\mathcal{G}|$ intuitively controlling the contributions of the terms as the upper bound of the contribution of a component (or performance aspect) to the ``total matching error''. These weights ensure that  each component corresponding to a performance aspect (Section \ref{subsec:EvaluationMeasure}) is easy to interpret, intuitively balances the components to yield \equref{\ref{eq:LRPdefcompact}} and prevents the total error from being undefined whenever the denominator of a single component is $0$. The first component in \equref{\ref{eq:LRPdef}}, $\mathrm{LRP_{Loc}}$, represents the localisation error of TPs as follows:
\begin{align}
\label{eq:Loc}
\mathrm{LRP_{Loc}}(\mathcal{G},\mathcal{D}):=\frac{1}{\mathrm{N_{TP}}}\sum \limits_{i=1}^{\mathrm{N_{TP}}} (1-\mathrm{lq}(g_i, d_{g_i})).
\end{align}
The second component, $\mathrm{LRP_{FP}}$, measures the FP rate:
\begin{align}
\label{eq:Type1}
\mathrm{LRP_{FP}}(\mathcal{G},\mathcal{D}):= 1-\mathrm{Precision}=1- \frac{\mathrm{N_{TP}}}{|\mathcal{D}|}=\frac{\mathrm{N_{FP}}}{|\mathcal{D}|},
\end{align}
and the FN rate is  measured by $\mathrm{LRP_{FN}}$:
\begin{align}
\label{eq:Type2}
\mathrm{LRP_{FN}}(\mathcal{G},\mathcal{D}):= 1-\mathrm{Recall}=1- \frac{\mathrm{N_{TP}}}{|\mathcal{G}|}=\frac{\mathrm{N_{FN}}}{|\mathcal{G}|}.
\end{align}
When necessary, the individual importance of localisation, FP, FN errors can be changed for different applications (Section \ref{sec:Comparison} and Appendix \ref{app:weight}). 

\subsection{An Analysis of LRP Error} 
\label{subsec:LRPAnalysis}
As we did for AP (Section \ref{subsec:APAnalysis}) and PQ (Section \ref{subsec:PQAnalysis}), in the following we analyse LRP Error in terms of important features for a performance measure.

\textbf{Completeness:} Both definitions of LRP Error above (which are equivalent to each other) clearly take into account  all  performance aspects precisely, and ensure completeness (Section \ref{subsec:EvaluationMeasure}).

\textbf{Interpretability:} LRP Error and its components are in $[0,1]$, and a lower value implies better performance. LRP Error describes the ``average matching error'' (see Section \ref{subsec:LRP}), and each  component summarizes the error for a single performance aspect, thereby providing insights on the strengths and weaknesses of a detector (compare with PQ in Table \ref{tab:PQComponents}(b)). Hence, LRP Error ensures interpretability (Section \ref{subsec:EvaluationMeasure}). In the extreme cases;  $\mathrm{LRP}=0$ implies each ground truth is detected with perfect localisation. If $\mathrm{LRP}=1$, no detection matches any ground truth (i.e., $|\mathcal{D}|=\mathrm{N_{FP}}$). 


\textbf{Practicality:} Since \equref{\ref{eq:LRPdefcompact}} requires a thresholded detection set (i.e. does not require confidence scores), LRP Error can directly be employed to evaluate hard predictions, and can be computed exactly without requiring any interpolations or approximations. In the next section, we discuss how LRP Error can be extended to evaluate soft predictions using Optimal LRP Error and show that it can also be computed exactly. Also, in order to prevent the over-represented classes in the dataset to dominate the performance, similar to AP and PQ, LRP Error is computed class-wise and these class-wise LRP Errors are averaged to assign the LRP Error of a detector. One practical issue of LRP Error is that localisation and FP components are undefined when there is no detection, and the FN component is undefined when there is no ground truth. However, even when some components (not all) are undefined, the LRP Error is still defined (\equref{\ref{eq:LRPdefcompact}}). $\mathrm{LRP}$ is undefined only when the ground truth and detection sets are both  empty (i.e., $\mathrm{N_{TP}}+\mathrm{N_{FP}} + \mathrm{N_{FN}}=0$), i.e., there is nothing to evaluate.  When a component is undefined, we ignore the value while averaging it over classes.


\begin{figure}
    \centerline{
        \includegraphics[width=0.5\textwidth]{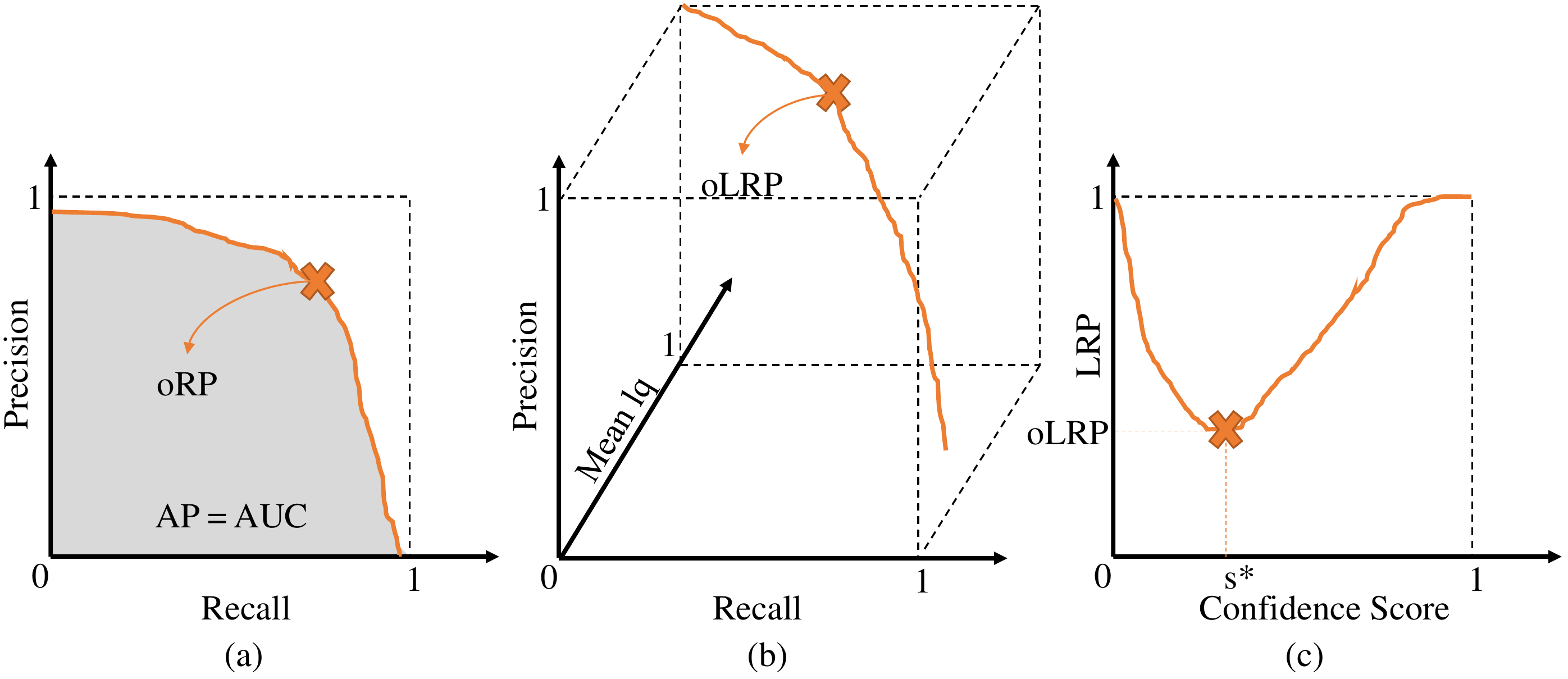}
    }
\caption{A visual comparison of AP and LRP Error. (a) A PR curve. The cross marks a hypothetical optimal-recall-precision point (oRP) (e.g. the point where F1-measure is maximized). (b) A localisation, recall and precision curve, where  ``Mean lq'' is the average localisation quality of TPs. Unlike any performance measure obtained via a PR curve (e.g. AP), LRP Error intuitively combines these performance aspects (\equref{\ref{eq:LRPdefcompact}}), and instead of area under the curve, uses the minimum of LRP Error values, defined as oLRP Error, as the performance metric. (c) An s-LRP curve. Its minimum is oLRP Error.
}
\label{fig:LRPVariants}
\end{figure}

\subsection{Optimal LRP (oLRP) Error: Evaluating and Thresholding Soft Predictions}
\label{subsec:oLRP}
\noindent \textbf{Definition:} Soft predictions (i.e. outputs with confidence scores) can be evaluated by, first, filtering the detections from a confidence score threshold and then, calculating LRP Error. 
We define Optimal LRP (oLRP) Error as the minimum achievable LRP Error  over the detection thresholds or equivalently, the confidence scores\footnote{Another way  to evaluate soft predictions is the Average LRP Error (aLRP), the average of the LRP Errors over the confidence scores. While in a recent study \cite{aLRPLoss}, we showed that aLRP can be used as a loss function, we discuss in Appendix \ref{app:aLRP} why we preferred oLRP over aLRP as a performance measure.}:
\begin{align}
\label{eq:OptimalLRP}
\mathrm{oLRP}:= \min_{s \in \mathcal{S}} \mathrm{LRP}(\mathcal{G}, \mathcal{D}_s),
\end{align}
where $\mathcal{D}_s$ is the set of detections thresholded at confidence score $s$ (i.e. those detections with larger or equal to the confidence scores than $s$ are kept, and others are discarded). \equref{\ref{eq:OptimalLRP}} implies searching over a set of confidence scores, $\mathcal{S}$, to find the best balance for competing precision, recall and localisation errors.


\noindent \textbf{Components:} The components of LRP Error for $\mathrm{oLRP}$ are coined as localisation@oLRP ($\mathrm{oLRP_{Loc}}$), FP@oLRP ($\mathrm{oLRP_{FP}}$), and FN@oLRP ($\mathrm{oLRP_{FN}}$).  $\mathrm{oLRP_{Loc}}$ describes the average localisation error of TPs, and $\mathrm{oLRP_{FP}}$ and $\mathrm{oLRP_{FN}}$ together indicate the point on the PR curve where optimal LRP Error is achieved. More specifically, one can infer the shape of the PR curve using the $(1-\mathrm{oLRP}_{FP}, 1-\mathrm{oLRP}_{FN})$ pair defining the optimal point on the PR curve. 



\noindent \textbf{Computation:} Note that, theoretically, computing oLRP Error requires infinitely many thresholding operations since $\mathcal{S}=[0, 1]$. However, given that $\mathcal{S}$ is discretised by the scores of the detections, in order to compute oLRP Error exactly, it is sufficient to threshold the detection set only at the confidence scores of the detections. More formally, for two successive detections $d_i$ and $d_j$ (in terms of confidence scores) with confidence scores $s_i$ and $s_j$ where $s_i > s_j$, $\mathrm{LRP}(\mathcal{G}, \mathcal{D}_s) = \mathrm{LRP}(\mathcal{G}, \mathcal{D}_{s_j})$ if $s_j \leq s<s_i$. Then, oLRP Error for a class can be computed \textbf{exactly} by minimizing the finite LRP Errors on the detections, and one can average oLRP Error and its components over classes to obtain the performance of  the detector.



\noindent \textbf{LRP-Optimal Thresholds:} Conventionally, visual object detectors yield numerous detections \cite{FocalLoss,FasterRCNN,FCOS}, most of which have smaller confidence scores. In order to deploy an object detector for a certain problem, the detections with ``smaller'' confidence scores need to be discarded to provide a clear output from the visual detector (i.e. model selection). While it is common to use a single class-independent threshold for the detector (e.g., Association-LSTM \cite{AssociationLSTM} uses SSD \cite{SSD} detections for all classes with confidence score above $0.80$), we show in Section \ref{subsec:ThresholdingExperiments} that (i) the performances of the detectors are sensitive to thresholding, and (ii) the thresholding needs to be handled in a class-specific manner. Note that balancing the competing performance aspects in an optimal manner, oLRP Error satisfies these requirements. In particular, we define the confidence score threshold corresponding to the oLRP Error as the ``LRP-Optimal Threshold'' ($s^*$ - see \figref{\ref{fig:LRPVariants}}). Different from the common approach, (i) $s^*$ is a class-specific optimal threshold, and (ii) $s^*$ considers all performance aspects of visual detection tasks (\figref{\ref{fig:LRPVariants}}). See Appendix \ref{app:weight} for a further discussion on thresholding object detectors.

\noindent \textbf{$s$-LRP curves:} An $s$-LRP curve (\figref{\ref{fig:LRPVariants}}(c)) presents the performance distribution of a detector in terms of $\mathrm{LRP}(\mathcal{G}, \mathcal{D}_s)$ (\equref{\ref{eq:OptimalLRP}}) for a class over confidence scores. The minimum LRP Error on this curve determines both the LRP-Optimal confidence score ($s^*$) and the corresponding oLRP Error for a class. The shape of an $s$-LRP curve provides information on the sensitivity of the detector wrt. model selection (i.e. the choice of $s^*$): While a relatively flat curve implies that the threshold choice is not very critical (e.g. see Cascade R-CNN in \figref{\ref{fig:LRPConfScoreCurves}}(a)), a bell-like shape suggests the importance of the usage of LRP-Optimal thresholds for that detector (e.g. see ATSS in (\figref{\ref{fig:LRPConfScoreCurves}}(a) and Section \ref{subsec:ThresholdingExperiments}). Also note that obtaining the oLRP Error using the underlying $s$-LRP curve is different from how AP is computed from the PR curve. In particular, while AP is the area under the PR curve (\figref{\ref{fig:LRPVariants}}(a)), oLRP Error is the minimum LRP Error value on $s$-LRP curve (\figref{\ref{fig:LRPVariants}}(c)). As a result, while  including very-low-precision detections   (i.e. in the tail part of PR curve)  increases AP by ensuring high recall, such detections do not have an effect on oLRP Error.

\subsection{Potential Extensions of LRP Error}
\label{subsec:UsageExtensions}
We discuss potential extensions of LRP Error in three levels:

\textbf{Extension to Other Localisation Quality Functions:} Any localisation function that satisfies the following two constraints can be used within LRP Error: 
(i) $\mathrm{lq}(\cdot,\cdot)$ should be a higher-better function, and (ii) $\mathrm{lq}(\cdot,\cdot) \in [0, 1]$. In addition, choosing a $\mathrm{lq}(\cdot,\cdot)$ such that $1-\mathrm{lq}(\cdot,\cdot)$ is a metric, guarantees the metricity of LRP Error. In case constraint (ii) is violated by a prospective $\mathrm{lq}(\cdot,\cdot)$, then one can normalize the range of the function (and also TP validation threshold, $\tau$) to satisfy this constraint. For example, as a recently proposed IoU variant to measure the spatial similarity between two bounding boxes, Generalized IoU (GIoU) \cite{GIoULoss} has a range of $[-1,1]$. In this case, choosing $\mathrm{lq}(\cdot,\cdot)= \mathrm{GIoU}(\cdot,\cdot)/2 + 0.50$ will allow the use of GIoU within LRP Error.

\textbf{Extension to New Detection Tasks:} While adopting for new detection tasks, one should only consider the localisation quality function (see above). Following this, LRP Error can easily be adapted to new or existing detection tasks such as 3D object detection and rotated object detection. 

\textbf{Extension to Other Fields:} LRP Error can  be extended for any problem with the following two properties in terms of evaluation: (i) the similarity between a TP and its matched ground truth can be measured by using a similarity function (preferably a metric to ensure the metricity of LRP Error), and (ii) at least one of the classification errors (i.e. FP error or FN error) matters for performance. Then, to use LRP Error, it is sufficient to ensure the similarity function satisfy the constraints for $\mathrm{lq}(\cdot,\cdot)$ (see extension to other localisation quality functions). If either FP or FN error is not included in the task, then one can set the number of errors originating from the missing component (i.e. $\mathrm{N_{FP}}$ or $\mathrm{N_{FN}}$) to $0$ and proceed with \equref{\ref{eq:LRPdefcompact}}.

\section{A Comparison of LRP with AP and PQ}
\label{sec:Comparison}

\begin{figure*}
        \captionsetup[subfigure]{}
        \centering
        \begin{subfigure}[b]{0.3\textwidth}
        \includegraphics[width=\textwidth]{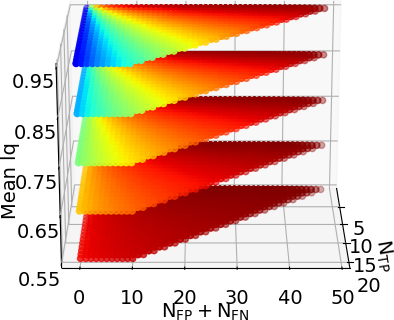}
        \caption{LRP Error}
        \end{subfigure}
        \begin{subfigure}[b]{0.3\textwidth}
        \includegraphics[width=\textwidth]{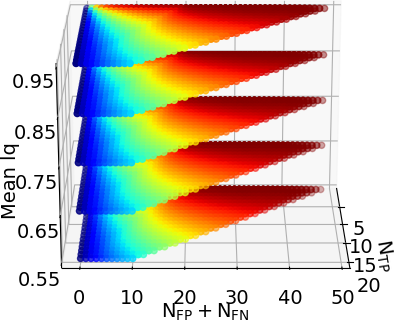}
        \caption{PR Error (1-(Precision $\times$ Recall))}
        \end{subfigure}
        \begin{subfigure}[b]{0.3\textwidth}
        \includegraphics[width=\textwidth]{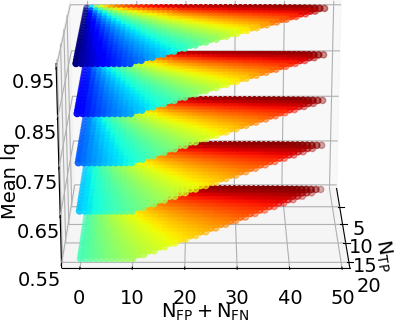}
        \caption{PQ Error (1-PQ)}
        \end{subfigure}
        \begin{subfigure}[b]{0.04\textwidth}
        \includegraphics[width=\textwidth]{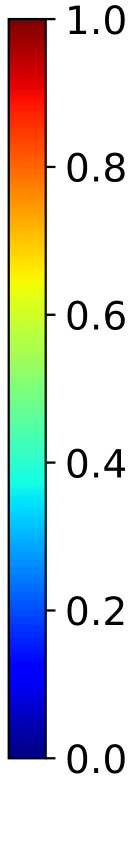}
        \vspace{0mm}
        \end{subfigure}
        \caption{How LRP Error, PR Error  (i.e. 1-(Precision $\times$ Recall)) and PQ Error (i.e. 1-PQ) behave over different inputs. Mean lq is the avg. localisation qualities of TPs. PR Error ignores localisation and PQ overpromotes classification compared to localisation. The  space is uniformly discretized. Error combinations of up to 20 ground truths and 50 detections are shown.}
        \label{fig:ErrorSpace}
\end{figure*}

\begin{figure*}
        \captionsetup[subfigure]{}
        \centering
        \begin{subfigure}[b]{0.3\textwidth}
        \includegraphics[width=\textwidth]{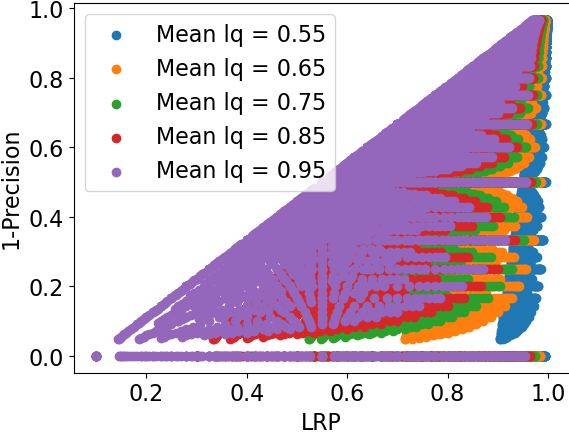}
        \caption{LRP Error vs Precision Error}
        \end{subfigure}
        \begin{subfigure}[b]{0.3\textwidth}
        \includegraphics[width=\textwidth]{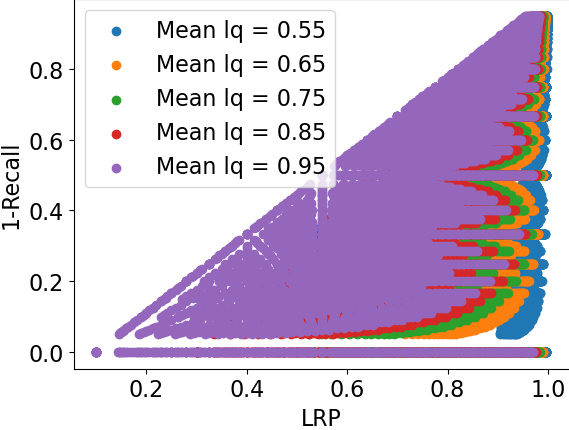}
        \caption{LRP Error vs Recall Error}
        \end{subfigure}
        \begin{subfigure}[b]{0.3\textwidth}
        \includegraphics[width=\textwidth]{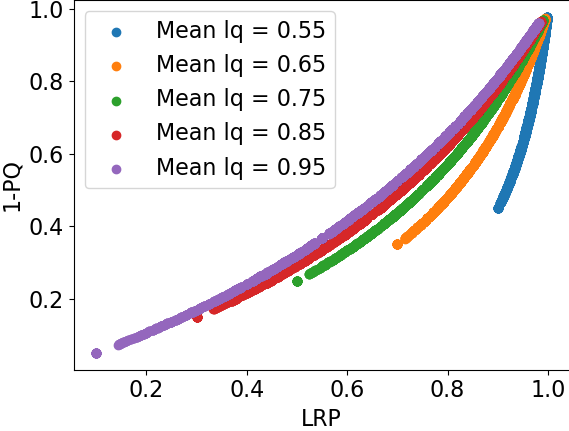}
        \caption{LRP Error vs PQ Error}
        \end{subfigure}
        \caption{The relation of LRP Error with (a) precision error (1-precision), (b) recall error (1-recall) and (c) PQ error (1-PQ) using the examples from \figref{\ref{fig:ErrorSpace}. LRP Error is an upper bound for precision, recall and PQ errors. Since LRP Error includes recall (precision) and localisation in addition to precision (recall) error, the correlation between precision (recall) error and LRP Error is not strong. On the other hand, with similar definitions LRP Error and PQ evaluate similarly, but still  their difference increases when Mean lq decreases since PQ suppresses the effect of localisation by promoting classification more.}}
        \label{fig:ErrorSpaceTwo}
\end{figure*}

To better understand the behaviours of the studied performance measures (AP, PQ and LRP Error) and make comparisons, we plot them with respect to (wrt.) Mean $\mathrm{lq}$, $N_\mathrm{TP}$ and $N_\mathrm{FP}+N_\mathrm{FN}$ (\figref{\ref{fig:ErrorSpace}}). To facilitate comparison, we represent AP and PQ by their ``error'' versions, that is, for AP, we use ``PR Error'' which is  1- Precision $\times$ Recall; and for PQ, we use ``PQ Error'' which is 1-PQ. 
The figure shows that PR Error stays the same as you move parallel to the ``Mean lq'' axis as expected (\figref{\ref{fig:ErrorSpace}}(b)). This is because PR Error, hence AP, uses the localisation quality just to validate TPs, and it does not take into account the quality above the TP-validation threshold. Moreover, PQ Error is lower than LRP Error  at low ``Mean lq'', e.g. 0.55 and 0.65, low $\mathrm{N_{FP}}+\mathrm{N_{FN}}$ and large $\mathrm{N_{TP}}$ (\figref{\ref{fig:ErrorSpace}}(a) and (c)). This is due to the fact that PQ Error prefers to emphasize  classification over localisation (as discussed in Section \ref{sec:PanopticQuality}). On the other hand, as hypothesized in Section \ref{sec:PanopticQuality}, the way how PQ overpromotes classification  is inconsistent. To show this, we first demonstrate that LRP Error and PQ Errors have quite similar definitions. PQ Error can  be written as (see Appendix \ref{app:similarity} for the derivation):
\begin{align}
        \label{eq:PQvsLRP}
        1- \mathrm{PQ} = \frac{1}{\hat{Z}} \left( \sum \limits_{i=1}^{\mathrm{N_{TP}}}  \frac{1-\mathrm{lq}(g_i, d_{g_i})}{1-0.50}+\mathrm{N_{FP}} +\mathrm{N_{FN}} \right) ,
\end{align}
where $\hat{Z} = {\textcolor{red}{2} \mathrm{N_{TP}}+\mathrm{N_{FP}} +\mathrm{N_{FN}}}$. Note that setting $\tau = 0.50$ and removing the coefficient of $\mathrm{N_{TP}}$ in $\hat{Z}$ (in red) results in $1-\mathrm{PQ} = \mathrm{LRP}$ (\equref{\ref{eq:LRPdefcompact}}), which implies very similar definitions for PQ and LRP Errors (and note that LRP Error was proposed before PQ): \equref{\ref{eq:PQvsLRP}} presents that (i) the ``total matching error'' of PQ and LRP Errors are equal (Section \ref{subsec:LRP} for total matching error), and (ii) PQ Error prefers doubling  $\mathrm{N_{TP}}$ in the normalisation constant instead of normalizing the total matching error directly by its maximum value (i.e. ${\mathrm{N_{TP}}+\mathrm{N_{FP}} +\mathrm{N_{FN}}}$) as done by LRP Error. Therefore, keeping the total matching error the same, the normalisation constant of PQ Error grows inconsistently. In other words, the rates of the change of the total matching error and its maximum possible value are different. As suggested in our previous work \cite{LRP}, a consistent prioritization of a performance aspect can be achieved by including its coefficient to both total matching error (i.e. nominator) and its maximum value (i.e. denominator) as follows:
\begin{align}
\label{eq:LRPGeneralized}
\frac{1}{Z} \left( \sum \limits_{i=1}^{\mathrm{N_{TP}}} \mathrm{\alpha_{TP}} \frac{1-\mathrm{lq}(g_i, d_{g_i})}{1-\tau}+ \mathrm{\alpha_{FP}} \mathrm{N_{FP}} + \mathrm{\alpha_{FN}} \mathrm{N_{FN}} \right),
\end{align}
where $Z = \mathrm{\alpha_{TP}} \mathrm{N_{TP}}+ \mathrm{\alpha_{FP}}\mathrm{N_{FP}}+ \mathrm{\alpha_{FN}}\mathrm{N_{FN}}$. Following the interpretation of LRP Error (Section \ref{subsec:LRP}), these coefficients imply duplicating each error source, hence the consistency between the total matching error and its maximum value is preserved (see Appendix \ref{app:similarity} for more discussion).


\begin{table*}
\centering
\caption{Comparison of AP, PQ and LRP Error in terms of desired properties. While LRP Error ensures all three, AP turns out to be limited in these properties. \label{tab:desired}}
\renewcommand{\arraystretch}{0.6}
    \begin{tabular}{|c|c|c|c|}
         \hline
         \textbf{Measure} &\textbf{Completeness}&\textbf{Interpretability}&\textbf{Practicality}\\ \hline
         AP&\xmark&\xmark&\begin{minipage}[c][1.3cm][c]{0.5\textwidth}\centering \begin{itemize}
         \item limited to soft predictions
         \item does not offer an optimal confidence score threshold
         \item sensitive to design choices
\end{itemize}
 \end{minipage}
\\ \hline
         PQ&\cmark&\cmark&\begin{minipage}[c][1cm][c]{0.5\textwidth}\centering \begin{itemize}
         \item limited to panoptic segmentation
         \item overpromotes classification perform. inconsistently 
\end{itemize}
 \end{minipage}\\ \hline
         LRP Error&\cmark&\cmark&\cmark\\ \hline
    \end{tabular}
\end{table*}

In \figref{\ref{fig:ErrorSpaceTwo}}, we  present the relationship of LRP Error with precision, recall and PQ Errors,  which show that  \textbf{LRP Error is an upper bound for all other error measures}. As a result, improving LRP Error can be considered a  more challenging task than improving the other two error measures.

The comparison of LRP Error with AP and PQ in terms of the important features (Section \ref{subsec:EvaluationMeasure}) of a performance measure for visual object detectors is summarized in Table \ref{tab:desired}. Please refer to Sections \ref{sec:AveragePrecision}, \ref{sec:PanopticQuality} and \ref{section:LRP} for further discussion.

\section{Experimental Evaluation}
\label{section:experiments}
In this section, we first present the usage and discriminative abilities of the LRP Error on visual detection tasks in comparison to AP variants (Section \ref{subsec:SoftPredictionExperiments}) and PQ (Section \ref{subsec:HardPredictionExperiments}). Then, we show that LRP Error can be used for datasets with different characteristics and for different visual detection tasks (Section \ref{subsec:DifferentDatasetsTasks}). Finally, we show that the performances of object detectors are sensitive to thresholding (Section \ref{subsec:ThresholdingExperiments}). Also, we describe how LRP Error can be used for tuning hyperparameters, discuss how manually manipulating sources of errors (e.g. by setting N\_FP=0) affects LRP Error on an example visual detector, provide a use-case of LRP-Optimal Thresholds, analyse the additional overhead of LRP Error computation and the behaviour of LRP Error under different TP validation thresholds
in Appendix \ref{subsec:ThresholdingExperiments}\ In this section, our main motivation is to present insights on LRP Error and represent its evaluation capabilities rather than choosing which detection method is better.


\subsection{Evaluated Models, Datasets and Performance Measures}
\label{subsec:ModelDataset}
\noindent \textbf{Evaluated Models and Datasets:} We evaluate around  different 100 models on 7 different visual detection tasks (object detection, keypoint detection, instance segmentation, panoptic segmentation, visual relationship detection, zero-shot detection and generalised zero-shot detection) using 10 different datasets (COCO object detection \cite{COCO},  COCO keypoint detection \cite{COCO}, COCO instance segmentation \cite{COCO}, COCO-stuff \cite{COCOStuff}, V-COCO \cite{v-coco}, COCO split for zero-shot detection, LVIS \cite{LVIS}, Open Images \cite{OpenImages}, Pascal \cite{PASCAL} and ILSVRC \cite{ILSVRC}). In general, we do not retrain the models but use the already trained instances provided in the commonly used repositories (e.g. mmdetection \cite{mmdetection}, detectron2 \cite{Detectron2}). For reproducibility, Appendix \ref{app:repos} provides the corresponding repository for each model.

\noindent \textbf{Performance Measures:} On tasks with hard-predictions (i.e. panoptic segmentation), we compare LRP Error with PQ. On the remaining tasks, all of which are soft-prediction tasks, we compare oLRP Error with AP. Since AP does not explicitly have performance components, we include the following measures to facilitate comparison: (i)
$\mathrm{AP_{\tau}}$, where $\tau$ is the TP validation threshold. With $\tau=0.75$, $\mathrm{AP_{75}}$ is a popular measure to represent the localisation accuracy and with $\tau=0.50$, $\mathrm{AP_{50}}$, to represent the classification component. We also use Average Recall, $\mathrm{AR^C_{r}}$ where $r$ is the number of top-scoring detections to include in the computation of AR. Note that $\mathrm{AR^C_{r}}$ is the  COCO-style version (i.e. averaged over 10 $\tau$ thresholds - see the definition of COCO-style AP, denoted by $\mathrm{AP^{C}}$, in Section \ref{sec:AveragePrecision}).

\renewcommand{\arraystretch}{0.5}
\begin{table*}
\setlength{\tabcolsep}{0.2em}
\caption{Performance comparison of methods for soft-prediction tasks (i.e. object detection, keypoint detection and instance segmentation) on COCO 2017 val.  
For $\mathrm{AR^C_{r}}$, $r = 100$ except for keypoint detection in which $r = 20$. \label{tbl:perf_comparison}} 
 \centering
\begin{tabular}{|l|c|c||c|c|c|c||c|c|c|c|}
\hline
& & &\multicolumn{4}{|c||}{AP \& AR}&\multicolumn{4}{|c|}{oLRP Error \& Components}\\\cline{4-11}
Method&Backbone&Epoch&$\mathrm{AP^{C}} \uparrow$&$\mathrm{AP_{50}} \uparrow$&$\mathrm{AP_{75}} \uparrow$&$\mathrm{AR^C_{r}} \uparrow$&oLRP $\downarrow$&$ \mathrm{oLRP_{Loc}} \downarrow$&$ \mathrm{oLRP_{FP}} \downarrow$&$ \mathrm{oLRP_{FN}} \downarrow$\\
\hhline{===========}
\textbf{Object Detection:}& & & & & & & & &&\\
\textit{One Stage Methods:}& & & & & & & & &&\\
SSD-300 \cite{SSD}&VGG16&120&$25.6$&$43.8$&$26.3$&$37.5$&$78.4$&$20.6$&$37.1$&$57.9$\\
SSD-512 \cite{SSD}&VGG16&120&$29.4$&$49.3$&$31.0$&$42.5$&$75.4$&$19.7$&$32.8$&$53.6$\\
RetinaNet \cite{FocalLoss}&R50&12&$35.7$&$54.7$&$38.5$&$52.0$&$71.0$&$17.0$&$29.1$&$50.0$\\
RetinaNet \cite{FocalLoss}&R50&24&$35.7$&$54.9$&$38.2$&$51.4$&$70.6$&$17.1$&$28.4$&$49.6$\\
RetinaNet \cite{FocalLoss}&X101&24&$39.2$&$59.2$&$41.8$&$53.5$&$67.5$&$16.1$&$24.5$&$46.3$\\
ATSS \cite{ATSS}&R50&12& $39.4$&$57.6$&$42.8$&$58.3$&$68.6$&$15.4$&$30.3$&$46.6$\\
RetinaNet \cite{FocalLoss}&X101&12&$39.8$&$59.5$&$43.0$&$54.8$&$67.6$&$16.1$&$25.3$&$46.2$\\
NAS-FPN \cite{NASFPN}&R50&50 &$40.5$&$58.4$&$43.1$&$55.6$&$66.7$&$14.8$&$26.6$&$46.3$\\
GHM  \cite{gradientharmonizing}&X101&12 &$41.4$&$60.9$&$44.2$&$57.7$&$66.3$&$15.6$&$27.1$&$44.2$\\
FreeAnchor \cite{FreeAnchor}&X101&12 &$41.9$&$61.0$&$45.0$&$58.6$&$66.0$&$15.2$&$26.4$&$44.5$\\
FCOS \cite{FCOS}&X101&24&$42.5$&$62.1$&$45.7$&$58.2$&$64.4$&$14.9$&$25.4$&$41.9$\\
RPDet \cite{RepPoints}&X101&24&$44.2$&$65.5$&$47.8$&$58.7$&$63.3$&$15.4$&$23.4$&$39.5$\\
aLRP Loss \cite{aLRPLoss}&X101&100&$45.4$&$66.6$&$48.0$&$60.3$&$62.5$&$15.1$&$23.2$&$39.5$\\
\textit{Two Stage Methods:}& & & & & & & & &&\\
Faster R-CNN \cite{FasterRCNN}&R50&24&$37.9$&$59.3$&$41.1$&$51.0$&$68.8$&$17.4$&$25.7$&$45.4$\\
Faster R-CNN  \cite{FasterRCNN}&R101&12&$39.4$&$61.2$&$43.4$&$52.6$&$67.6$&$17.2$&$24.2$&$44.3$\\
Faster R-CNN \cite{FasterRCNN}&R101&24&$39.8$&$61.3$&$43.3$&$52.5$&$67.3$&$16.8$&$25.5$&$43.4$\\
Faster R-CNN \cite{FasterRCNN}&X101&12&$41.3$&$63.7$&$44.7$&$54.6$&$66.2$&$17.1$&$24.9$&$41.5$\\
Libra R-CNN \cite{LibraRCNN}&X101&12&$42.7$&$63.7$&$46.9$&$56.0$&$65.1$&$15.8$&$24.3$&$41.6$\\
Grid R-CNN \cite{GridRCNN}&X101&24&$43.0$&$61.6$&$46.7$&$56.7$&$64.2$&$14.4$&$24.7$&$42.3$\\
Guided Anchoring \cite{GuidedAnchoring}&X101&12&$43.9$&$63.7$&$48.3$&$59.9$&$64.4$&$14.8$&$25.6$&$41.8$\\
Cascade R-CNN  \cite{CascadeRCNN}&X101&20&$44.5$&$63.2$&$48.5$&$56.9$&$63.3$&$14.3$&$25.4$&$41.0$\\
Cascade R-CNN  \cite{CascadeRCNN}&X101&12&$44.7$&$63.6$&$48.9$&$57.4$&$63.2$&$14.4$&$23.9$&$40.9$\\
\hhline{===========}
\textbf{Keypoint Detection:}& &  & & & & & & &&\\
Keypoint R-CNN &R50&12&$64.0$&$86.4$&$69.3$&$71.0$&$44.8$&$12.8$&$10.8$&$18.6$\\
Keypoint R-CNN &R50&37&$65.5$&$87.2$&$71.1$&$72.4$&$43.0$&$12.3$&$10.4$&$17.3$\\
Keypoint R-CNN &R101&37&$66.1$&$87.4$&$72.0$&$73.1$&$42.0$&$11.9$&$9.0$&$17.8$\\
Keypoint R-CNN &X101&37&$66.0$&$87.3$&$72.2$&$73.2$&$41.9$&$11.7$&$8.8$&$18.1$\\
\hhline{===========}
\textbf{Instance Segmentation:}& &  & & & & & & &&\\
Mask R-CNN \cite{MaskRCNN}&R50&12&$34.7$&$55.7$&$37.2$&$47.8$&$71.2$&$18.9$&$29.1$&$48.1$\\
Carafe \cite{carafe}&R50&12&$35.8$&$57.4$&$38.2$&$49.0$&$70.3$&$18.6$&$28.8$&$45.9$\\
GRoIE \cite{groie}&R50&12&$36.0$&$57.0$&$38.5$&$49.1$&$70.3$&$18.2$&$28.5$&$46.7$\\
PointRend \cite{PointRend}&R50&12&$36.3$&$56.9$&$38.7$&$50.1$&$70.5$&$18.2$&$27.8$&$47.7$\\
Mask R-CNN  \cite{MaskRCNN}&R101&24&$36.6$&$57.9$&$39.1$&$48.8$&$69.4$&$18.2$&$25.9$&$46.3$\\
Mask R-CNN \cite{MaskRCNN}&X101&12&$38.4$&$60.6$&$41.3$&$50.3$&$67.8$&$18.3$&$24.9$&$43.5$\\
Cascade Mask R-CNN  \cite{CascadeRCNN}&X101&20&$39.5$&$61.3$&$42.5$&$50.5$&$66.8$&$18.0$&$24.3$&$42.1$\\
Mask Scoring R-CNN \cite{MaskScoringRCNN}&X101&12&$39.5$&$60.5$&$42.6$&$50.1$&$67.5$&$17.9$&$24.5$&$43.3$\\
DetectoRS \cite{detectors}&R50&12&$42.6$&$65.1$&$46.0$&$56.6$&$64.6$&$17.1$&$23.2$&$39.9$\\
Hybrid Task Cascade \cite{HTC}&X101&20&$43.8$&$66.8$&$47.1$&$57.4$&$63.6$&$17.0$&$23.4$&$37.9$\\\hline
\end{tabular}
\end{table*}

\renewcommand{\arraystretch}{0.5}
\begin{table*}
\setlength{\tabcolsep}{0.15em}
\caption{Performance comparison of several object detectors for two arbitrarily selected classes from COCO dataset essentially to provide insight on the components of oLRP Error, namely, ``person'' and ``broccoli'', on COCO 2017 val.  
The horizontal line splits one- and two-stage object detectors. 
\label{tbl:class_perf_comparison}} 
 \centering
\begin{tabular}{|c|l|c|c||c|c|c|c||c|c|c|c|}
\hline
&&& &\multicolumn{4}{|c||}{AP \& AR}&\multicolumn{4}{|c|}{oLRP Error \& Components}\\\cline{5-12}
Class&Method&Backbone&Epoch
&$\mathrm{AP^{C}} \uparrow$&$\mathrm{AP_{50}} \uparrow$&$\mathrm{AP_{75}} \uparrow$&$\mathrm{AR}^{C}_{100} \uparrow$&oLRP $\downarrow$&$ \mathrm{oLRP_{Loc}} \downarrow$&$ \mathrm{oLRP_{FP}} \downarrow$&$ \mathrm{oLRP_{FN}} \downarrow$\\
\hhline{============}
\multirow{6}{*}{\rotatebox[origin=c]{90}{Person}}&ATSS \cite{ATSS}&R50&12&$54.7$&$81.4$&$59.3$&$64.8$&$55.6$&$15.4$&$14.8$&$27.8$\\
&FCOS \cite{FCOS}&X101&24&$55.3$&$82.4$&$59.1$&$64.2$&$53.5$&$15.0$&$13.0$&$26.2$\\
&RetinaNet \cite{FocalLoss}&X101&12&$51.1$&$79.0$&$54.6$&$60.1$&$58.3$&$16.3$&$14.1$&$31.0$\\
&aLRP Loss \cite{aLRPLoss} &X101&100&$57.7$&$84.3$&$61.7$&$66.6$&$52.1$&$14.5$&$11.2$&$26.3$\\ \cline{2-12}
&Faster R-CNN \cite{FasterRCNN}&R101&24&$53.8$&$82.8$&$57.9$&$61.2$&$55.3$&$16.3$&$11.1$&$27.7$\\
&Cascade R-CNN \cite{CascadeRCNN}&X101&12&$57.8$&$83.4$&$62.7$&$65.2$&$52.1$&$14.6$&$13.5$&$24.5$\\ \hline
\multirow{6}{*}{\rotatebox[origin=c]{90}{Broccoli}} &ATSS \cite{ATSS}&R50&12&$24.1$&$43.3$&$24.0$&$55.0$&$81.6$&$20.1$&$53.2$&$52.9$\\
&FCOS \cite{FCOS}&X101&24&$21.9$&$41.5$&$21.6$&$48.4$&$82.4$&$21.4$&$45.1$&$58.7$\\
&RetinaNet \cite{FocalLoss}&X101&12&$22.1$&$44.4$&$19.7$&$49.3$&$81.3$&$21.8$&$41.3$&$56.7$\\
&aLRP Loss \cite{aLRPLoss}&X101&100&$24.3$&$45.5$&$23.5$&$51.7$&$80.2$&$21.2$&$45.7$&$51.6$\\ \cline{2-12}
&Faster R-CNN \cite{FasterRCNN}&R101&24&$22.0$&$43.6$&$19.3$&$43.8$&$81.2$&$22.7$&$49.2$&$48.4$\\
&Cascade R-CNN \cite{CascadeRCNN}&X101&12&$24.3$&$43.9$&$25.5$&$47.2$&$80.2$&$20.1$&$51.8$&$48.7$\\ \hline
\end{tabular}
\end{table*}

\begin{figure*}
        \captionsetup[subfigure]{}
        \centering
        \begin{subfigure}[b]{0.24\textwidth}
        \includegraphics[width=\textwidth]{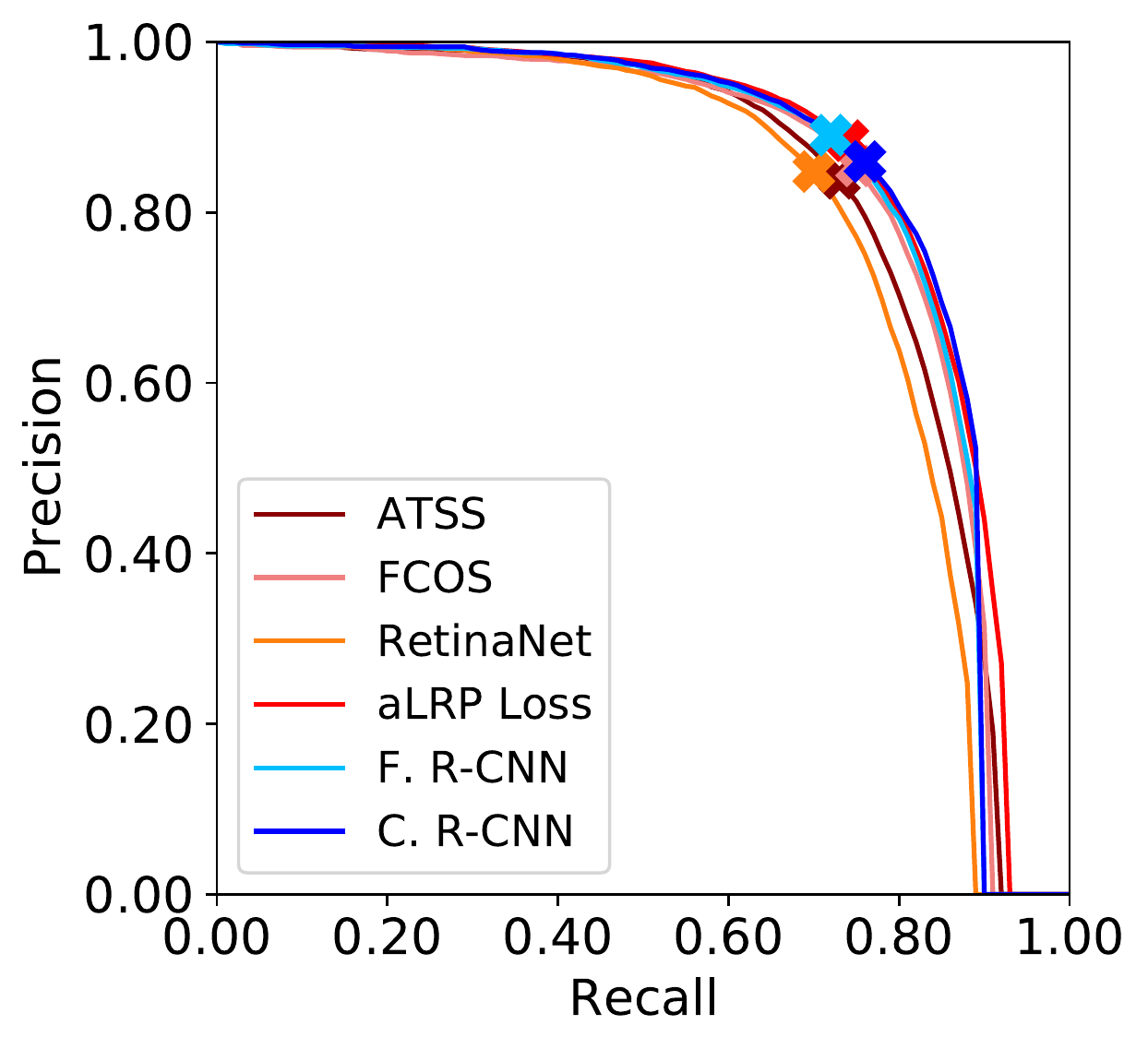}       
        \caption{Person}
        \end{subfigure}
        \begin{subfigure}[b]{0.24\textwidth}
        \includegraphics[width=\textwidth]{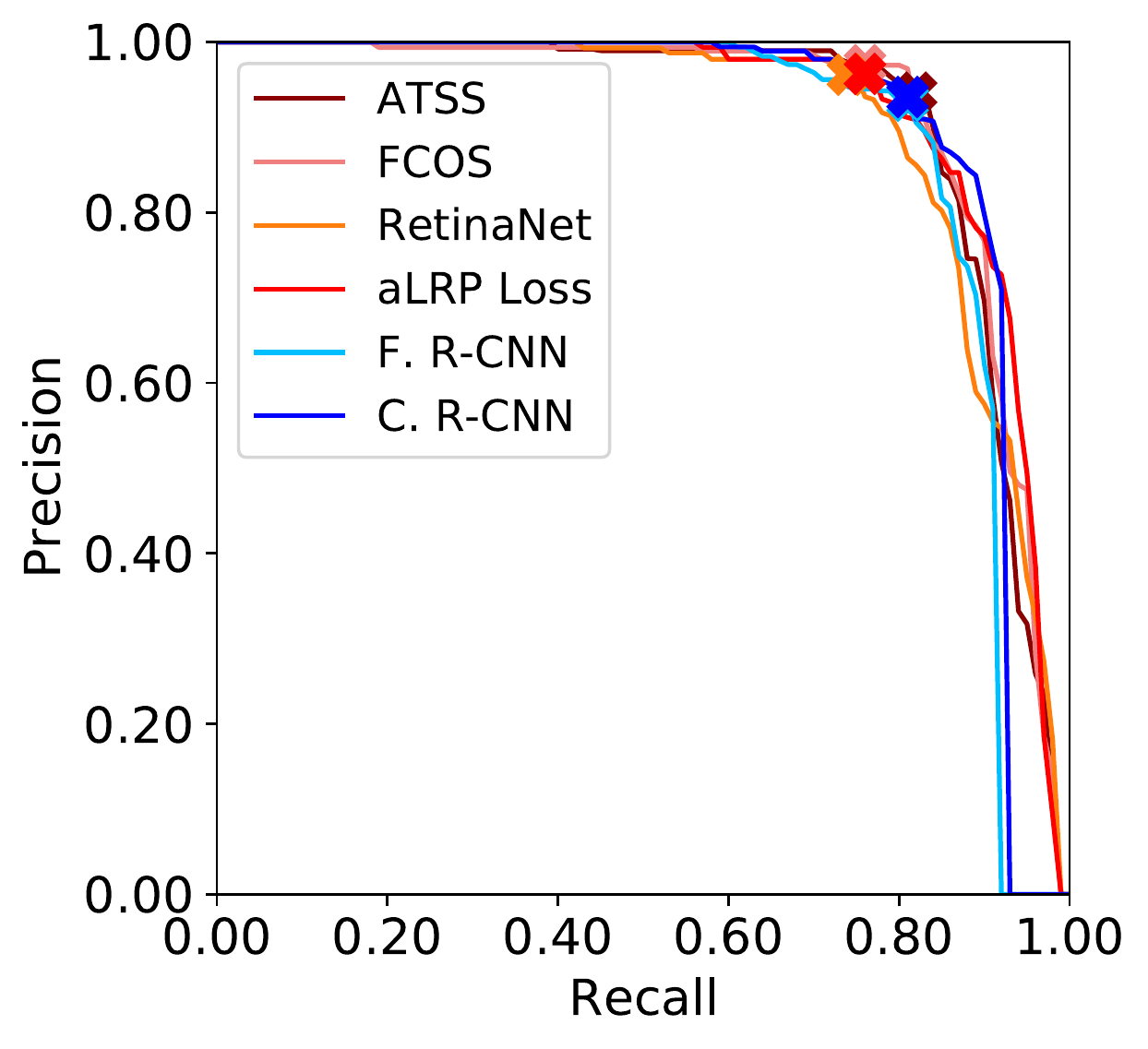}
        \caption{Zebra}
        \end{subfigure}
        \begin{subfigure}[b]{0.24\textwidth}
        \includegraphics[width=\textwidth]{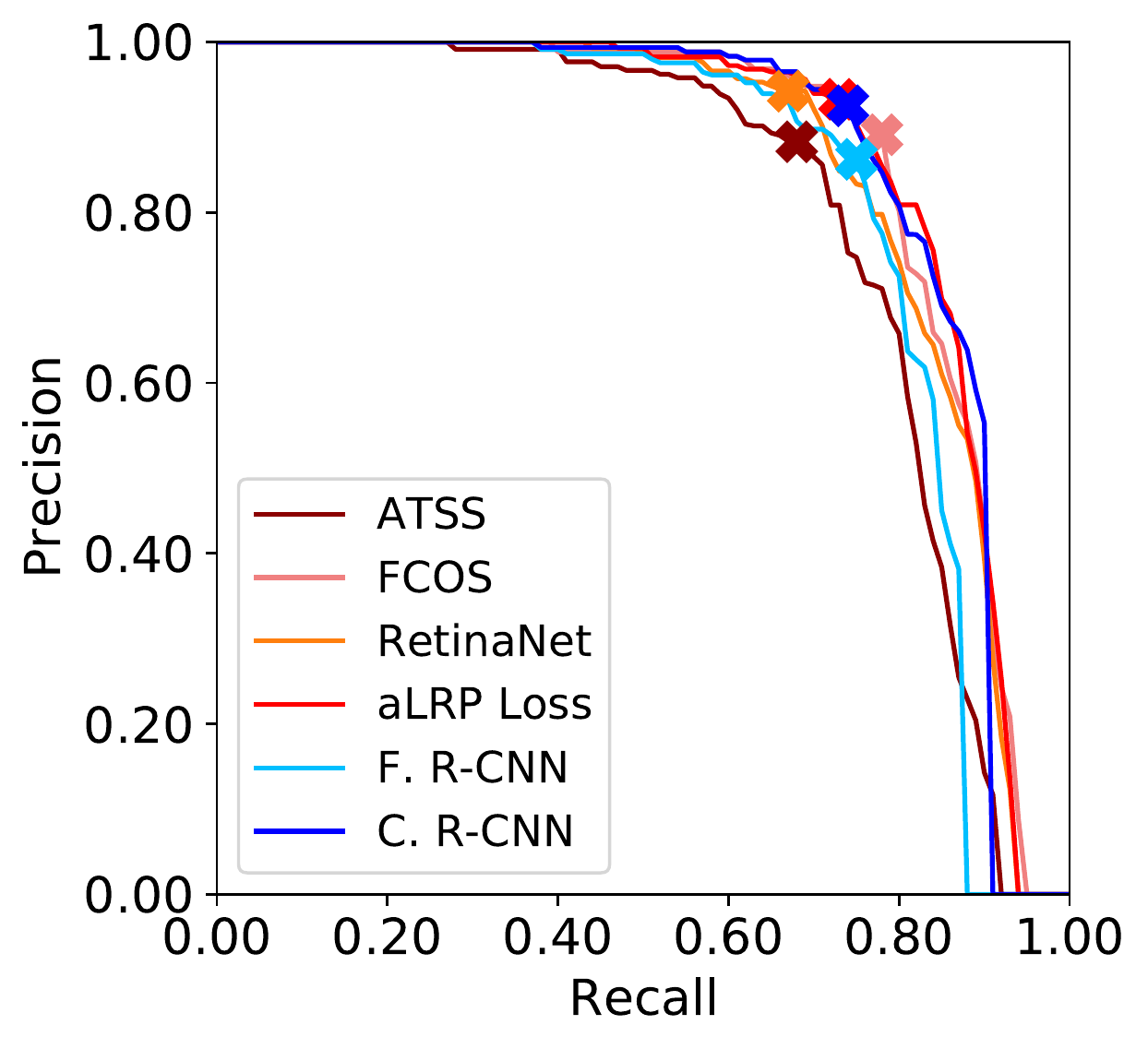}
        \caption{Bus}
        \end{subfigure}
        \begin{subfigure}[b]{0.24\textwidth}
        \includegraphics[width=\textwidth]{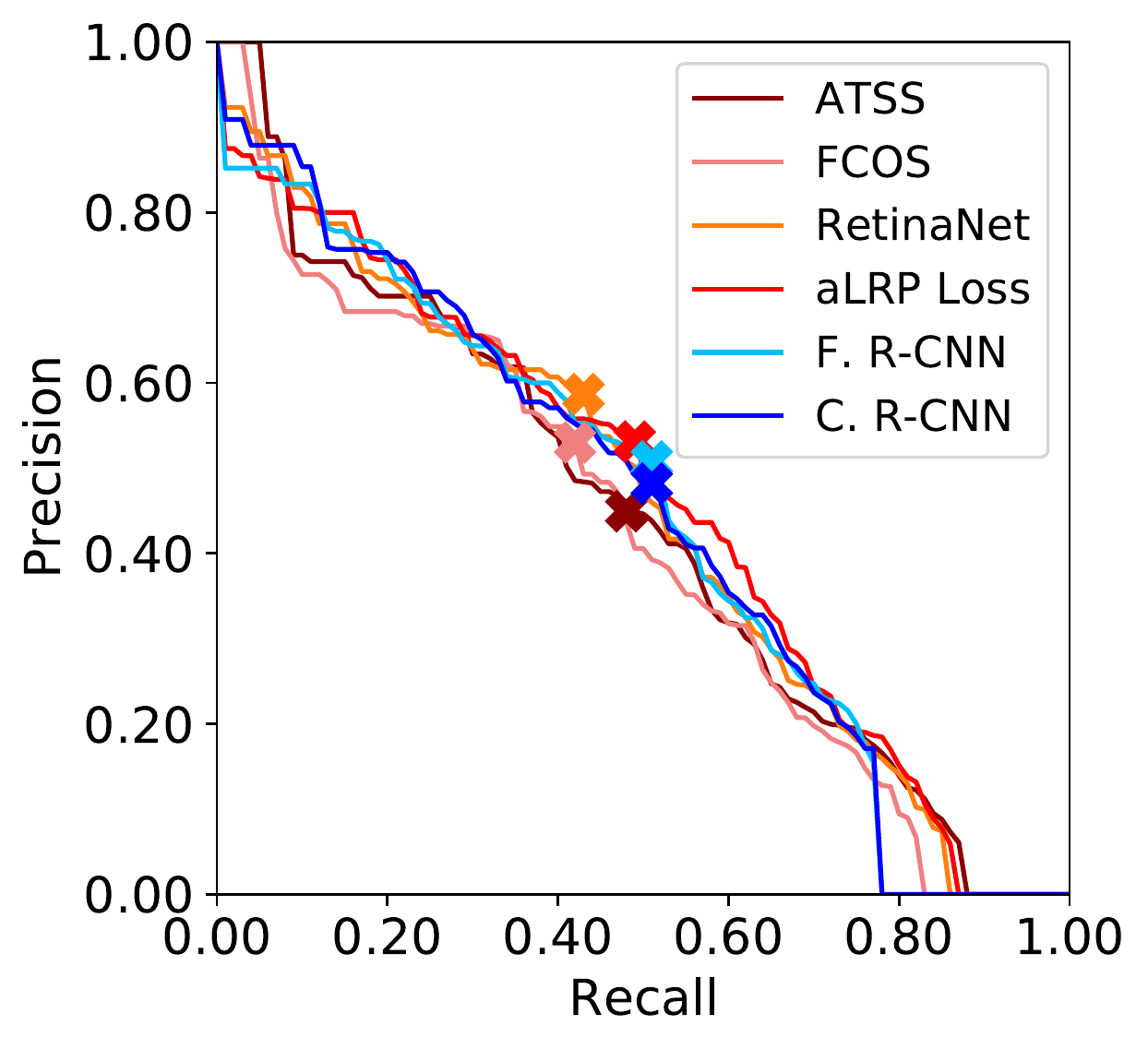}
        \caption{Broccoli}
        \end{subfigure}
        \caption{PR curves of four arbitrary  classes from COCO. The curves are drawn for $\tau=0.50$. The lines with different red/blue tones represent one/two-stage detectors. While $\mathrm{AP_{50}}$ considers  AUC of PR curve, LRP Error combines localisation, recall and precision errors, and hence oLRP Errors, marked with crosses, are found in the top right part of the curves. Thus, different from AP, oLRP Error is not affected by low-precision detections in the tail of the PR curves (e.g. one-stage detectors in (b), (d)). C.f. Table \ref{tbl:class_perf_comparison} for AP \& oLRP Error values, F. R-CNN: Faster R-CNN, C. R-CNN: Cascade R-CNN.}
        \label{fig:ClassPRCurves}
\end{figure*}

\subsection{Evaluating Soft Predictions on Object Detection, Keypoint Detection and Instance Segmentation Tasks}
\label{subsec:SoftPredictionExperiments}
This section compares oLRP Error with AP\&AR variants for soft predictions and shows that oLRP Error is more discriminative and interpretable. Unless explicitly specified, we use COCO dataset variants with corresponding annotations for each task.  The structure of this section is based on the limitations (Section \ref{subsec:APLimitations}) and analysis of AP (Section \ref{subsec:APAnalysis}) in terms of the important features (Section \ref{subsec:EvaluationMeasure}). While demonstrating these limitations and comparing with oLRP Error, we use both detector-level results (Table \ref{tbl:perf_comparison}) and class-level results (Table \ref{tbl:class_perf_comparison}) of the SOTA methods. For the detector-level performance comparison, we present the results of 36 SOTA visual detectors on three visual detection tasks. In order to provide insight on oLRP Error and its components and illustrate its usage at the class level, we select a subset of six object detectors by ensuring diversity (e.g. different backbones, one- and two-stage detectors, different assignment and sampling strategies etc.) and evaluate their performance on two arbitrarily selected classes, that are ``person'' and ``broccoli''.

\subsubsection{Analysis with respect to Completeness}
\label{subsec:Localisation}
AP loosely includes the localisation quality (Section \ref{subsec:APAnalysis}). Here we discuss the benefits of directly using the localisation quality as an input to the performance measure. 

Firstly, to see  how $\mathrm{AP_{50}}$ fails to include  localisation quality precisely, we consider the following three detectors with equal $\mathrm{AP_{50}}$ ($63.7 \%$) in Table \ref{tbl:perf_comparison}: Faster R-CNN (X101-12), Libra R-CNN and Guided Anchoring. oLRP Error and $\mathrm{AP^{C}}$, which  take into account the localisation quality,  rank these three detectors different from  $\mathrm{AP_{50}}$. Therefore, when a benchmark aims to promote methods yielding TPs with more accurate localisation, $\mathrm{AP_{50}}$ should not be selected as the single performance measure since it considers all TPs equally regardless of their localisation quality. A supporting example can be found in Section \ref{subsubsec:DifferentDatasets} where we compare $\mathrm{AP_{50}}$ with oLRP on Pascal dataset. 

To illustrate the drawback of $\mathrm{AP^{C}}$ or $\mathrm{AP_{75}}$ in terms of localisation, note that while neither $\mathrm{AP^{C}}$ nor $\mathrm{AP_{75}}$ assigns the largest performance value to NAS-FPN among one-stage object detectors, this detector has the least average localisation error ($\mathrm{oLRP_{Loc}}$ - see Table \ref{tbl:perf_comparison}): e.g. GHM outperforms NAS-FPN by $1.1 \%$ in terms of $\mathrm{AP_{75}}$, while its average localisation performance is $0.8 \%$ lower than NAS-FPN. Therefore, $\mathrm{AP^{C}}$ and $\mathrm{AP_{75}}$, too, may fail to   appropriately compare methods in terms of localisation quality.

Using oLRP Error is  easier and more intuitive than using the  AP variants mentioned above: (i) Unlike these AP variants, oLRP Error consistently and precisely (not loosely) combines localisation, FP and FN errors, and in this case, the performance gap between NAS-FPN and GHM reduces to $0.4 \%$ in terms of oLRP Error (i.e. while  GHM outperforms NAS-FPN by $1.1 \%$ with respect to $\mathrm{AP^{C}}$) thanks to the localisation performance of NAS-FPN. (ii) Different from all AP variants, $\mathrm{oLRP_{Loc}}$ quantifies the localisation error precisely and allows direct comparison among methods, classes, etc.: e.g. NAS-FPN outperforms GHM by $0.8 \%$ in terms of $\mathrm{oLRP_{Loc}}$. (iii) One can easily interpret  $\mathrm{oLRP_{Loc}}$  both at the class- and detector-level: for example, for ATSS, the mean $\mathrm{IoU}$ is $1-0.154=0.846$ and $1-0.201=0.799$  for the ``person” and ``broccoli” classes respectively (Table \ref{tbl:class_perf_comparison}).

Finally, being an interpretable localisation measure, $\mathrm{oLRP_{Loc}}$ can facilitate analysis of detectors. For example,  in Table \ref{tbl:perf_comparison}, we can easily notice that instance segmentation task has  room for  improvement in terms of localisation compared to other tasks. In particular, even the best performing instance segmentation method, Hybrid Task Cascade (HTC), yields $17.0\%$ $\mathrm{oLRP_{Loc}}$ error which corresponds to a mediocre localisation error relative to object detectors and keypoint detectors, typically achieving $14.3 \%$ and $11.7 \%$ $\mathrm{oLRP_{Loc}}$ errors, respectively. With this $17.0 \%$, HTC has a similar localisation performance with RetinaNet (R50-24) in terms of $\mathrm{oLRP_{Loc}}$. HTC outperforms RetinaNet (R50-24) by around $10\%$  $\mathrm{AP_{75}}$, suggesting that the same deduction cannot be obtained by $\mathrm{AP_{75}}$.


\subsubsection{Analysis with respect to Interpretability}
\label{subsec:Interpreability}
This section presents insights about the interpretability of oLRP Error compared to AP (see Section \ref{subsec:APAnalysis} for a discussion on the limited interpretability of AP).

While any AP variant does not provide insight on the performance of a visual detector, the components of oLRP Error does. To illustrate, we compare two object detectors with equal $\mathrm{AP^{C}}$, ATSS and Faster R-CNN (R101-12) (see in Table \ref{tbl:perf_comparison} that both have $39.4 \%$ $\mathrm{AP^{C}}$), using AP \& AR based measures and oLRP Error \& components as follows:
\begin{itemize}
    \item Faster R-CNN (R101-12) outperforms ATSS by $3.6 \%$ and $0.6 \%$ in terms of both $\mathrm{AP_{50}}$ and $\mathrm{AP_{75}}$ and ATSS outperforms Faster R-CNN (R101-12) by around $6 \%$ with respect to  $\mathrm{AR^C_{100}}$. Note that a clear conclusion (i.e. quantifying which detector is better on which performance aspect) is not possible with these AP \& AR based measures since $\mathrm{AP_{50}}$ and $\mathrm{AP_{75}}$ are combinations of precision \& recall and $\mathrm{AR^C_{100}}$ a combination of recall \& localisation quality.
    \item As for oLRP Error, Faster R-CNN (R101-12) outperforms ATSS by $6.1 \%$ and $2.3 \%$ in terms of FP and FN Errors respectively, and ATSS has $1.8 \%$ better localisation performance than Faster R-CNN (R101-12). Since each component corresponds to one performance aspect, one can easily deduce that while Faster R-CNN has better classification performance (wrt. both precision and recall), ATSS localises objects better. Overall, combining these components consistently, in this case, oLRP Error prefers Faster R-CNN (R101-12) over ATSS by $1.1 \%$ while they have the same $\mathrm{AP^{C}}$.
\end{itemize}

In addition, $\mathrm{oLRP_{FP}}$ and $\mathrm{oLRP_{FN}}$ provide insight on the structure of the PR curve by representing the point on the PR curve where the minimal LRP Error is achieved. To illustrate, for all methods, the ``person'' class has  lower FP \& FN error values than the ``broccoli'' class, implying the oLRP Error point of the ``person'' PR curve to be closer to the top-right corner. To see this, note that Faster R-CNN has $11.1 \%$ and $27.7 \%$ FP and FN error values, respectively  for the ``person'' class (Table \ref{tbl:class_perf_comparison}). Thus, without looking at the curve, one may conclude  that the oLRP Error point resides at $1-0.111=0.889$ precision and  $1-0.277=0.723$ recall. For the ``broccoli” curve, the oLRP Error point is achieved at $1-0.492=0.508$ and $1-0.484=0.516$ as precision and recall, respectively. Unlike the ``person’’ class, these values suggest that the optimal point of the ``broccoli’’ class is around the center of the PR range (cf. \figref{\ref{fig:ClassPRCurves}}(a) and (d)). Hence, $\mathrm{oLRP_{FP}}$ and $\mathrm{oLRP_{FN}}$ are also easily interpretable and in such a way, exhaustive examination of PR curves can be alleviated.

Similar to $\mathrm{oLRP_{Loc}}$, $\mathrm{oLRP_{FP}}$ and $\mathrm{oLRP_{FN}}$ facilitate analysis as well, which is not straightforward by using AP\&AR based measures. Suppose that we want to compare precision and recall performances of the visual object detectors. Comparing $\mathrm{oLRP_{FP}}$ and $\mathrm{oLRP_{FN}}$, it is obvious that current object detectors have significantly lower precision error than recall error (i.e. $\mathrm{oLRP_{FP}}-\mathrm{oLRP_{FN}}$ is around $15\%$ to $20\%$ for object detection and instance segmentation, $7\%$ to $8\%$ for keypoint detection - see Table \ref{tbl:perf_comparison}). Given AP\&AR based measures, one alternative can be to compare $\mathrm{AP_{50}}$ with $\mathrm{AR^C_{100}}$, which is again hampered by the loose and indirect combination of the performance aspects: Note that while $\mathrm{oLRP_{FP}}$ and $\mathrm{oLRP_{FN}}$, isolating errors with respect to performance aspects, assign significantly more recall error than precision error for ATSS ($\mathrm{oLRP_{FN}}-\mathrm{oLRP_{FP}}=16.5 \%$ - see Table \ref{tbl:perf_comparison}), AP- and AR- based performance measures favor recall performance over precision performance ($\mathrm{AR^C_{100}} > \mathrm{AP_{50}}$). Therefore, indirect contribution of the performance aspects makes the analysis more difficult for AP- and AR-based measures, while oLRP Error and components are easier to interpret and compare.






\begin{figure}
        \captionsetup[subfigure]{}
        \centering
        \begin{subfigure}[b]{0.24\textwidth}
        \includegraphics[width=\textwidth]{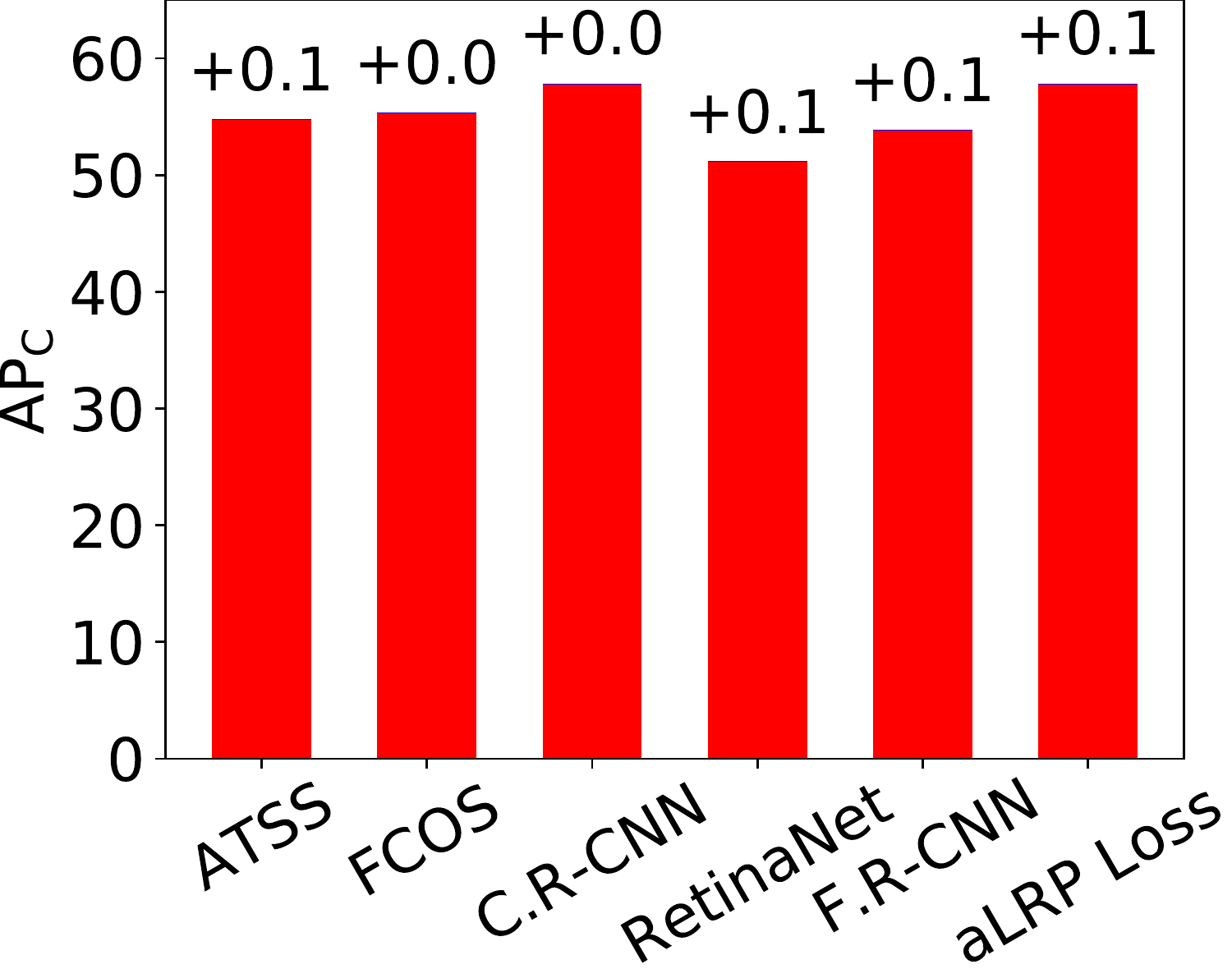}
        \caption{Person class}
        \end{subfigure}
        \begin{subfigure}[b]{0.24\textwidth}
        \includegraphics[width=\textwidth]{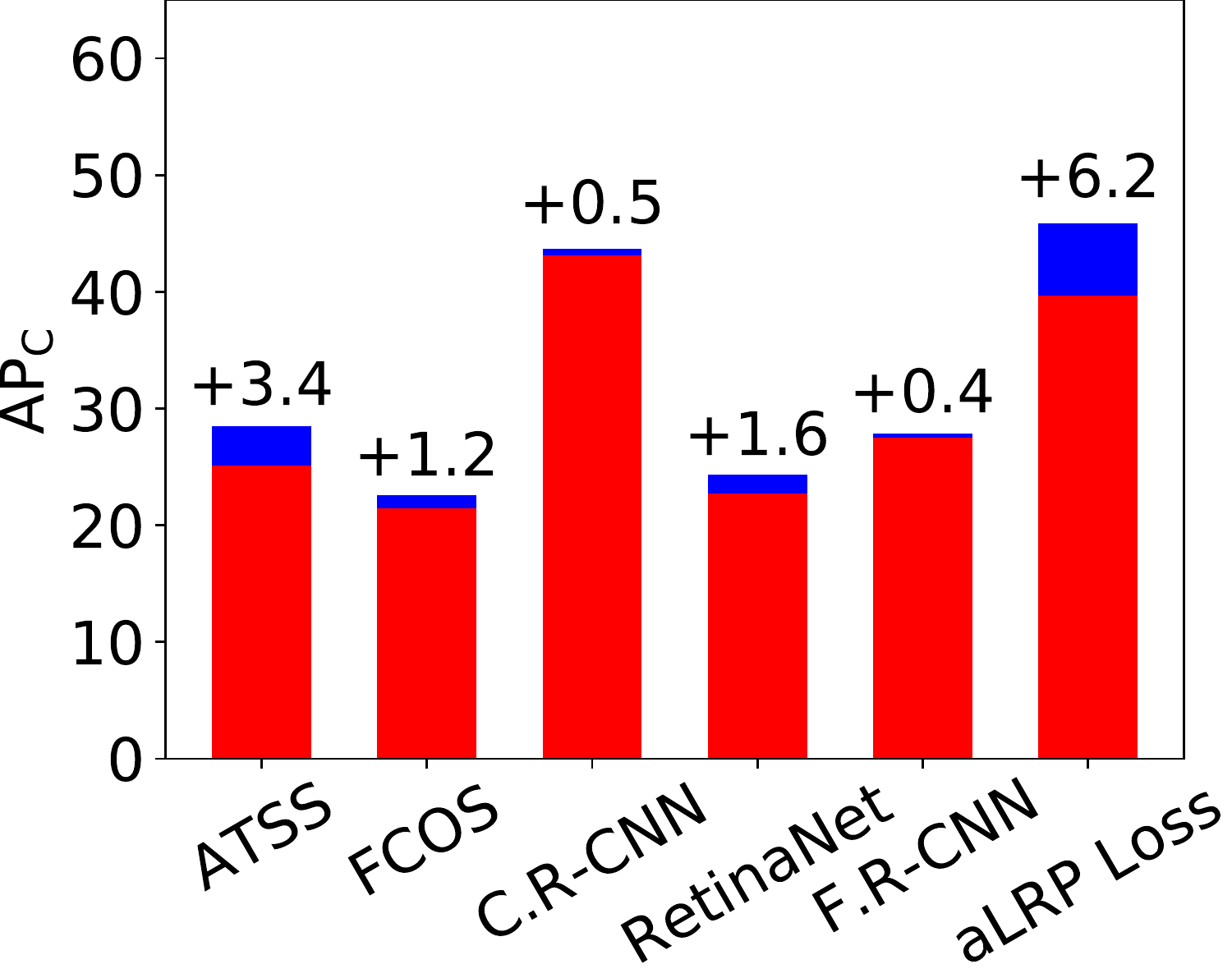}
        \caption{Toaster class}
        \end{subfigure}
        \caption{The effect of interpolating the PR curve on $\mathrm{AP^{C}}$ using COCO dataset on (a) the ``person'' class, the class with the most number of examples, (b) the ``toaster'' class, the class with the least number of examples. Red: AP without interpolation. Blue: Additional $\mathrm{AP^{C}}$ after interpolation. The numbers on the bars indicate this additional $\mathrm{AP^{C}}$ points due to interpolation. While the effect of interpolation is negligible for the  ``person'' class, there is a significant effect of interpolation (i.e. $2.2 \%$ $\mathrm{AP^{C}}$ on average, up to $6.2 \%$) on the performance of the toaster class (i.e. the class with the minimum number of examples) for all the detectors. C.R-CNN: Cascade R-CNN, F.R-CNN: Faster R-CNN.}
        \label{fig:interpolation}
\end{figure}

\subsubsection{Analysis with respect to Practicality}
\label{subsec:PracticalIssues}
This section presents how LRP Error can handle the practical limitations of AP (see Section \ref{subsec:APAnalysis} for a discussion why AP is limited in these practical issues).

\noindent \textbf{Evaluating Hard Predictions:} We discuss how LRP Error evaluates hard predictions in Section \ref{subsec:HardPredictionExperiments}.

\noindent \textbf{Thresholding Visual Object Detectors:} We discuss our class-specific thresholding approach using LRP-Optimal thresholds in Section \ref{subsec:ThresholdingExperiments}.

\noindent \textbf{Interpolating the PR Curve:} In order to present the effect of  interpolation on  classes with relatively fewer number of examples, we compute $\mathrm{AP^{C}}$ of the same six detectors from class-level comparison table (Table \ref{tbl:class_perf_comparison}) with and without interpolation on two classes (\figref{\ref{fig:interpolation}}): (i) the ``person'' class as the class with maximum number examples in COCO val 2017 (i.e. 21554 examples), and (ii) the ``toaster'' class , the class with the minimum number of examples (i.e. 17 examples). The $\mathrm{AP^{C}}$ differences in \figref{\ref{fig:interpolation}} show the significant effect of interpolation on the class with the less number of examples: (i) While the average $\mathrm{AP^{C}}$ difference over detectors between with and without interpolation is almost negligible for person class (i.e. less than $0.1 \%$ $\mathrm{AP^{C}}$), it is around $35 \times$ more for the toaster class ($2.2 \%$ $\mathrm{AP^{C}}$). (ii) There is even $6.2\%$ jump for the aLRP Loss after interpolation. Note that this corresponds to around  a superficial $20 \%$  relative performance improvement (from $33.4 \%$ to $39.6 \%$) in terms of $\mathrm{AP^{C}}$. Therefore, while the AP variants are sensitive  to interpolation especially for the rare classes in the dataset, oLRP Error does not employ interpolation. Note that, unlike AP, oLRP Error computation is exact (Section \ref{subsec:oLRP}).

\noindent \textbf{Approximating the Area Under PR Curve:} Table \ref{tbl:pascal} shows that the AP values from the same model evaluated by different APIs can significantly vary ($\sim 3 \mathrm{AP_{50}}$). Note that the main difference of these APIs is while Pascal API computes AUC exactly, COCO API approximates it (Section \ref{subsec:APAnalysis}, see practicality). On the other hand, oLRP Errors are equal. 

\noindent \textbf{Limiting the number of detections:} We show that LRP Error is insensitive to limiting the number of detections on the LVIS dataset  in Section \ref{subsubsec:DifferentDatasets}.

\blockcomment{
\begin{table}
\setlength{\tabcolsep}{0.1em}
\small
\caption{While evaluating the same model on two different evaluation APIs results in more than $2.5$ AP difference, due to the difference in approximating the area under PR curve, oLRP Error computation is standard and exact. All models are trained on Pascal 2007+2012 trainval set and tested on Pascal 2007 test set. F.R-CNN: Faster R-CNN \label{tbl:pascal}} 
 \centering
\begin{tabular}{|c|c||c||c|c|c|c|}
\hline
& & &\multicolumn{4}{|c|}{oLRP  $\downarrow$}\\\cline{4-7}
api &Model &$\mathrm{AP_{50}} \uparrow$&oLRP &$\mathrm{oLRP_{Loc}}$& $\mathrm{oLRP_{FP}}$&$\mathrm{oLRP_{FN}}$\\ \hline 
\multirow{3}{*}{\rotatebox{90}{Pascal}}&RetinaNet+R50&$77.3$&$56.8$&$16.2$&$17.1$&$28.1$ \\ \cline{2-7}
&F.R-CNN+R50&$79.5$&$56.7$&$18.2$&$14.5$&$24.6$ \\ \cline{2-7}
&F.R-CNN+R101&$81.3$&$54.1$&$17.3$&$13.1$&$22.8$ \\ \hline \hline 
\multirow{3}{*}{\rotatebox{90}{COCO}}&RetinaNet+R50&$80.0$&$56.8$&$16.2$&$17.1$&$28.1$ \\  \cline{2-7}
&F.R-CNN+R50&$82.3$&$56.7$&$18.2$&$14.5$&$24.6$ \\  \cline{2-7}
&F.R-CNN+R101&$84.0$&$54.1$&$17.3$&$13.1$&$22.8$ \\ \hline
\end{tabular}
\end{table}
}
\begin{table}
\small
\caption{Evaluation of the same model using  two different evaluation APIs (Pascal API and COCO API) results in $\sim 3 \mathrm{AP_{50}}$ difference, due to the difference in approximating the area under PR curve. On the other hand, oLRP and its components (not shown in the table),  are  exact, i.e. no interpolations or approximations are needed, which may introduce variability. All models are trained on Pascal 2007+2012 trainval set and tested on Pascal 2007 test set using mmdetection \cite{mmdetection}. F.R-CNN: Faster R-CNN \label{tbl:pascal}} 
 \centering
\begin{tabular}{|c||c|c|c|c|}
\hline
&\multicolumn{2}{|c|}{Pascal API}&\multicolumn{2}{|c|}{COCO API}\\\cline{2-5}
Model &$\mathrm{AP_{50}} \uparrow$&oLRP $\downarrow$&$\mathrm{AP_{50}} \uparrow$&oLRP $\downarrow$\\ \hline 
RetinaNet+R50&$77.3$&$56.8$&$80.0$&$56.8$ \\ \hline
F.R-CNN+R50&$79.5$&$56.7$&$82.3$&$56.7$ \\ \hline
F.R-CNN+R101&$81.3$&$54.1$&$84.0$&$54.1$ \\ \hline 
\end{tabular}
\end{table}
\renewcommand{\arraystretch}{0.6}
\begin{table*}
\setlength{\tabcolsep}{0.4em}
\caption{Detector-level performance results of panoptic segmentation methods as hard predictions on COCO dataset (with COCO-stuff). For each method, besides an overall average in All classes, we provide the performance of things and stuff as well. \label{tbl:SOTAPanoptic}} 
 \centering
\begin{tabular}{|l|c|c|c||c|c|c||c|c|c|c|}
\hline
& &&&\multicolumn{3}{|c||}{PQ \& Components}&\multicolumn{4}{|c|}{LRP Error \& Components}\\\cline{5-11}
Method&Backbone&Epoch&Type&PQ $\uparrow$&SQ $\uparrow$&RQ $\uparrow$&LRP $\downarrow$&$\mathrm{LRP_{Loc}} \downarrow$& $\mathrm{LRP_{FP}} \downarrow$& $\mathrm{LRP_{FN}} \downarrow$\\
\hhline{===========}
\multirow{3}{*}{Panoptic FPN \cite{PanopticFPN}}&\multirow{3}{*}{R50}&\multirow{3}{*}{12}&All&$39.4$&$77.8$&$48.3$&$77.5$&$21.0$&$39.3$&$57.2$ \\ \cline{4-11}
 &&&Things&$45.9$&$80.9$&$55.3$&$72.7$&$18.1$&$29.4$&$51.7$ \\ \cline{4-11}
 &&&Stuff&$29.6$&$73.3$&$37.7$&$84.6$&$25.3$&$54.3$&$65.5$ \\ \hline
\multirow{3}{*}{Panoptic FPN \cite{PanopticFPN}}&\multirow{3}{*}{R50}&\multirow{3}{*}{37}&All&$41.5$&$79.1$&$50.5$&$75.9$&$20.3$&$38.6$&$55.2$ \\ \cline{4-11}
 &&&Things&$48.3$&$82.2$&$57.9$&$70.8$&$17.8$&$29.3$&$49.1$ \\ \cline{4-11}
 &&&Stuff&$31.2$&$74.4$&$39.4$&$83.5$&$24.2$&$52.6$&$64.4$ \\ \hline
\multirow{3}{*}{Panoptic FPN \cite{PanopticFPN}}&\multirow{3}{*}{R101}&\multirow{3}{*}{37}&All&$43.0$&$80.0$&$52.1$&$74.6$&$19.4$&$37.0$&$53.6$ \\ \cline{4-11}
 &&&Things&$49.7$&$82.9$&$59.2$&$69.4$&$17.1$&$28.4$&$47.6$ \\ \cline{4-11}
 &&&Stuff&$32.9$&$75.6$&$41.3$&$82.3$&$22.9$&$50.2$&$62.7$ \\ \hline
\end{tabular}
\end{table*}

\subsection{Evaluating Hard Predictions on Panoptic Segmentation Task}
\label{subsec:HardPredictionExperiments}
In this section, we apply LRP Error to panoptic segmentation task on COCO dataset augmented by 53 classes from COCO-stuff \cite{COCOStuff} as background classes to present its ability to evaluate  hard predictions and also compare LRP Error with PQ.  In particular, we evaluate three different variants of Panoptic FPN \cite{PanopticFPN} using both LRP Error and PQ, and present the results in Table \ref{tbl:SOTAPanoptic} in three groups: (i) ``All'' includes all 133 classes, (ii) ``Things'' includes 80 object classes, and (iii) ``Stuff'' includes the remaining 53 classes, normally counted as background by other detection tasks. Similar to oLRP Error, we follow our analysis on PQ (Section \ref{subsec:PQAnalysis}) except the superiority of LRP Error on evaluating and thresholding soft predictions, which we discuss in Sections \ref{subsec:SoftPredictionExperiments} and \ref{subsec:ThresholdingExperiments} respectively.

\subsubsection{Analysis with respect to Interpretability}
\begin{figure}
        \centering
        \includegraphics[width=0.25\textwidth]{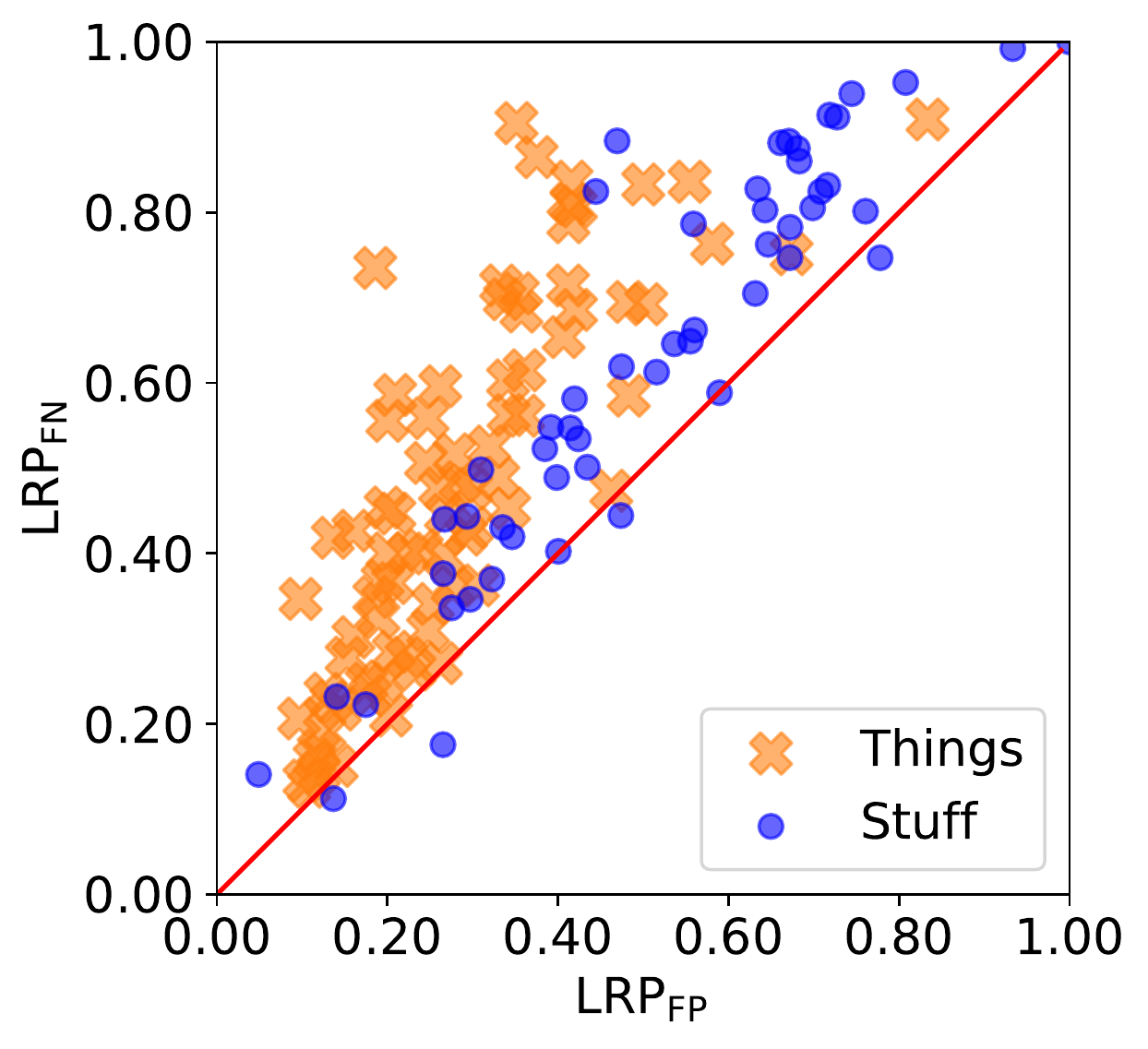}       
        \caption{Class-level $\mathrm{LRP_{FP}}$ vs. $\mathrm{LRP_{FN}}$ comparison on COCO dataset (with COCO Stuff). Overall, there is a tendency of larger $\mathrm{LRP_{FN}}$ error than $\mathrm{LRP_{FP}}$ error. This is more obvious for things classes. It is not possible to make the same observation using RQ.}
        \label{fig:PQvsLRP}
\end{figure}

The RQ component of PQ, the F-measure, does not provide discriminative information on precision and recall errors. On the other hand, LRP Error presents more insight on these errors with its FP and FN components. To illustrate, all Panoptic FPN variants suffer from the recall error more than the precision error, and this is more obvious for ``things" classes:  
(i) Table \ref{tbl:SOTAPanoptic} shows that $\mathrm{LRP_{FN}} > \mathrm{LRP_{FP}}$ for all  methods in class groups. 
(ii) 
While the gap between $\mathrm{LRP_{FP}}$ and $\mathrm{LRP_{FN}}$  for ``stuff'' classes is around $10 \%$, it is around $20 \%$ for ``things'' classes for all detectors. 
(iii) Finally, the same difference between ``things'' and ``stuff'' classes can easily be observed at the class-level in \figref{\ref{fig:PQvsLRP}} where the error is skewed towards $\mathrm{LRP_{FN}}$. Therefore, we argue that LRP FP and FN components present more insight than RQ.

\subsubsection{Analysis with respect to Practicality}
Since the definitions of LRP Error and PQ are similar (\equref{\ref{eq:PQvsLRP}}), LRP Error and PQ generally rank the detectors and classes similarly. However, we  observed certain differences owing to the over-promotion of TPs by PQ with its discontinuous nature: (i) We observed that 205 pair of classes for which the evaluation results of LRP Error and PQ conflict (i.e. $(\mathrm{PQ_i} < \mathrm{PQ_j})$  and $(\mathrm{LRP_j} > \mathrm{LRP_i})$ where the subscript represents the class label). As expected (see also \figref{\ref{fig:ErrorSpace}}(a,c) and \figref{\ref{fig:ErrorSpaceTwo}}(c)), PQ favors  classes with more TPs compared to LRP Error, and LRP Error favors the classes with better localisation performance. (ii) In some cases, the difference between the results of AP and PQ is significant. For example, while the ``bicycle'' and ``orange'' classes have $40.6 \%$ and $34.1 \%$ PQ respectively (i.e. ``bicycle'' outperforms by $6.5 \%$), their LRP Errors are $82.4 \%$ and $81.5 \%$ (i.e. ``orange'' outperforms by $0.9 \%$). 
The over-promotion of TPs by PQ can also be observed by examining its components: While the RQ of ``bicycle'' and ``orange'' are $55.9 \%$ and $38.4 \%$ respectively (i.e. ``bicycle'' outperforms by $17.5 \%$), SQ are $72.7 \%$ and $88.9 \%$ (i.e. ``orange'' outperforms by $16.2 \%$). These results suggest that while ``bicycle'' can be classified better than ``orange'', the localisation performance of ``bicycle'' is poorer. As a result, while LRP Errors are similar, PQ promotes the class with better classification (i.e. ``bicycle'') by $6.5 \%$ and assigns a lower priority to localisation. 

\begin{figure}
        \centering
        \begin{subfigure}[b]{0.24\textwidth}
        \includegraphics[width=\textwidth]{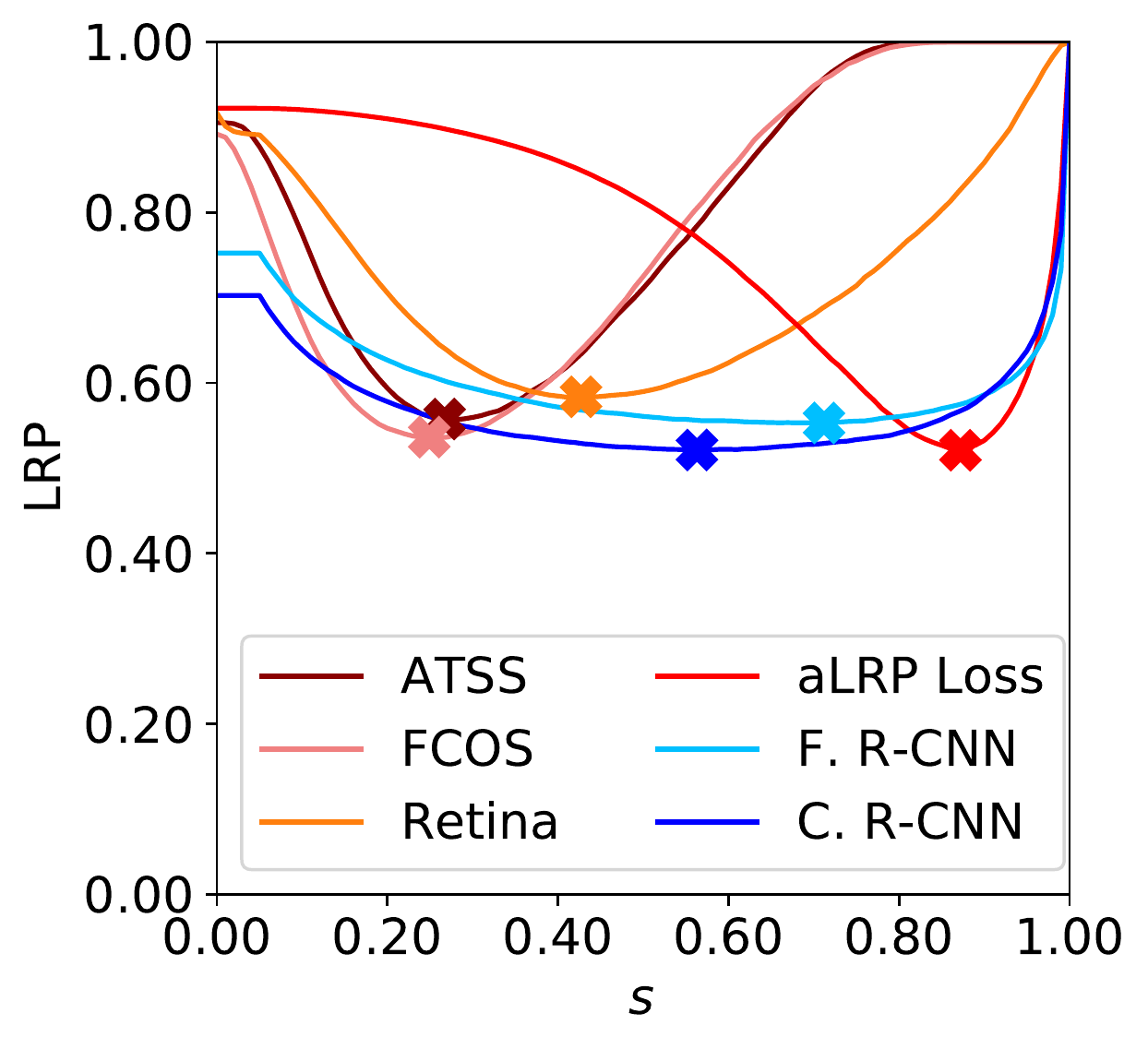}
        \caption{Person}
        \end{subfigure}
        \begin{subfigure}[b]{0.24\textwidth}
        \includegraphics[width=\textwidth]{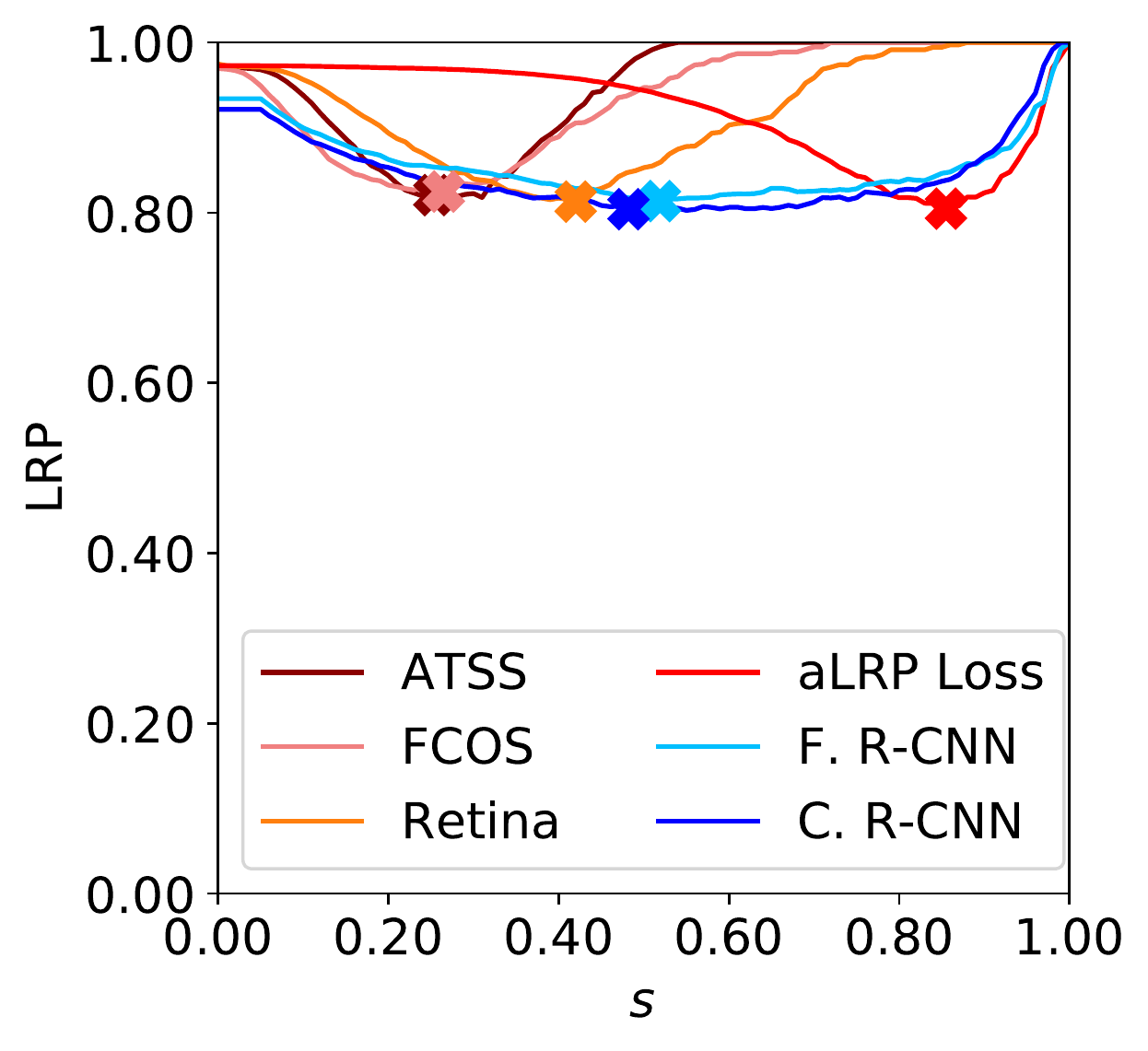}
        \caption{Broccoli}
        \end{subfigure}
        \caption{s-LRP curves of different object detectors for the arbitrarily chosen ``person'' and ``broccoli'' classes from COCO, whose PR curves are demonstrated in \figref{\ref{fig:ClassPRCurves}} (see Fig. S4 in Supp.Mat. for s-LRP curves of the ``zebra'', and ``bus''). The minimum-achievable LRP Error is oLRP Error (i.e. marked by ``x''). Note that for some detectors (e.g. ATSS), the performance with respect to $s$ changes abruptly implying sensitivity to thresholding. F. R-CNN: Faster R-CNN, C. R-CNN: Cascade R-CNN}
        \label{fig:LRPConfScoreCurves}
\end{figure}

\subsection{Evaluating Different Datasets and Tasks}
\label{subsec:DifferentDatasetsTasks}
This section shows that LRP Error can consistently evaluate other datasets with different characteristics and different visual detection tasks.

\subsubsection{Evaluating Datasets with Different Characteristics}
\label{subsubsec:DifferentDatasets}
\textbf{Evaluating rare classes on LVIS.} LVIS \cite{LVIS} is a long-tailed instance segmentation dataset in which a class is categorised as ``rare'', ``common'' and ``frequent'' if it has $\leq 10$, $11-100$ and $\geq 100$ instances respectively. Hence, $\mathrm{AP^C}$ on each of these partitions are also reported as $\mathrm{AP^C_{r}}$, $\mathrm{AP^C_{c}}$ and $\mathrm{AP^C_{f}}$ respectively besides the standard $\mathrm{AP^C}$. Accordingly, when we use oLRP Error on LVIS, we also report $\mathrm{oLRP_{r}}$, $\mathrm{oLRP_{c}}$ and $\mathrm{oLRP_{f}}$. We observe in Table \ref{tbl:LVIS} on Mask R-CNN that, with stronger backbones (e.g. X101) and more frequent classes (e.g. $\mathrm{oLRP_{f}}$), oLRP improves (i.e. decreases), implying consistent evaluation similar to AP.

\begin{table}
\setlength{\tabcolsep}{0.15em}
\small
\caption{Evaluating Mask R-CNN with different backbones on LVIS  v1.0. oLRP can also be calculated over different partitions of data, e.g. for LVIS, these are rare, common and frequent classes. We obtain models from mmdetection \cite{mmdetection}. \label{tbl:LVIS}} 
 \centering
\begin{tabular}{|c||c|c|c|c||c|c|c|c|}
\hline
&\multicolumn{4}{|c||}{AP $\uparrow$}&\multicolumn{4}{|c|}{oLRP Error  $\downarrow$}\\\cline{2-9}
Backbone&$\mathrm{AP^C}$ &$\mathrm{AP^C_{r}}$&$\mathrm{AP^C_{c}}$&$\mathrm{AP^C_{f}}$&oLRP&$\mathrm{oLRP_{r}}$& $\mathrm{oLRP_{c}}$& $\mathrm{oLRP_{f}}$\\
\hhline{=========}
R50&$21.7$&$9.6$&$21.0$&$27.8$&$80.7$&$91.0$&$80.7$&$76.2$ \\ \hline
R101&$23.6$&$13.2$&$22.7$&$29.3$&$79.0$&$87.5$&$79.3$&$74.9$ \\ \hline
X101&$25.5$&$16.0$&$24.8$&$30.5$&$77.5$&$85.1$&$77.7$&$73.8$ \\ \hline
\end{tabular}
\end{table}

Next, we provide a comparison with \textit{fixed AP} \cite{devil} (see practicality in Section \ref{sec:AveragePrecision}) using the model provided by detectron2 \cite{Detectron2}.
We observed in Table \ref{tbl:detectionnum} that (i) the differences between det\#=300 and det\#=1000 for dets/im. are $2.6 \%$ and $1.4 \%$ for AP and oLRP Error suggesting that oLRP Error is also sensitive to dets/im. but not as much as AP, (ii) when computed in the ``fixed'' style, oLRP Error is obviously more robust to dets/cl. in that AP and oLRP Error differences for det\#=1K and det\#=20K are $3.7 \%$ and $1.5 \%$ respectively and even oLRP Error saturates to $78.1 \%$ around 8K dets/cl. while AP does not saturate even with 20K dets/cl, and (iii) Even with 3K dets/cl. oLRP Error yields $78.4 \%$ which is close to $78.1 \%$, its saturated value, while AP has a significant gap ($24.5 \%$ vs. $25.8 \%$); hence it is possible to adopt oLRP Error with less \# of dets/cl. than AP.

\begin{table}
\setlength{\tabcolsep}{0.2em}
\small
\caption{Effect of number of detections (Det\#) on AP and oLRP Error on LVIS v1.0 when detections are limited per image (dets/im), and, as suggested by Dave et al. \cite{devil} as ``fixed'' version, per class (dets/cl). oLRP Error is more robust than AP wrt. these parameters. Underlined: default value for AP. \label{tbl:detectionnum}} 
 \centering
\begin{tabular}{|c|c||c|c|c|c||c|c|c|c|}
\hline
& &\multicolumn{4}{|c||}{AP $\uparrow$}&\multicolumn{4}{|c|}{oLRP Error $\downarrow$}\\\cline{3-10}
&Det\#&$\mathrm{AP^C}$ &$\mathrm{AP^C_{r}}$&$\mathrm{AP^C_{c}}$&$\mathrm{AP^C_{f}}$&oLRP&$\mathrm{oLRP_{r}}$& $\mathrm{oLRP_{c}}$& $\mathrm{oLRP_{f}}$\\
\hhline{==========}
\multirow{3}{*}{\rotatebox{90}{dets/im}}&\underline{300}&\underline{$22.7$}&\underline{$11.8$}&\underline{$21.7$}&\underline{$28.5$}&\underline{$79.8$}&\underline{$89.0$}&\underline{$80.1$}&\underline{$75.5$} \\ \cline{2-10}
&500&$24.0$&$13.5$&$23.5$&$29.1$&$79.1$&$87.3$&$79.1$&$75.4$ \\ \cline{2-10}
&1K&$25.3$&$16.9$&$24.7$&$29.6$&$78.4$&$84.4$&$78.7$&$75.4$ \\ \cline{2-10}
\hline \hline 
\multirow{7}{*}{\rotatebox{90}{dets/cl (i.e. \textit{fixed})}}
&1K&$22.1$&$17.3$&$22.9$&$23.4$&$79.6$&$83.9$&$79.3$&$78.2$ \\ \cline{2-10}
&2K&$23.7$&$17.8$&$24.1$&$25.9$&$78.7$&$83.6$&$78.8$&$76.6$ \\ \cline{2-10}
&3K&$24.5$&$18.4$&$24.6$&$27.0$&$78.4$&$83.2$&$78.6$&$76.1$ \\ \cline{2-10}
&5K&$25.1$&$18.6$&$25.0$&$28.1$&$78.2$&$82.9$&$78.5$&$75.7$ \\ \cline{2-10}
&8K&$25.5$&$18.8$&$25.3$&$28.7$&$78.1$&$82.8$&$78.5$&$75.6$ \\ \cline{2-10}
&\underline{10K}&\underline{$25.7$}&\underline{$18.9$}&\underline{$25.4$}&\underline{$29.0$}&\underline{$78.1$}&\underline{$82.7$}&\underline{$78.5$}&\underline{$75.5$} \\ \cline{2-10}
&20K&$25.8$&$18.9$&$25.4$&$29.2$&$78.1$&$82.7$&$78.5$&$75.5$ \\ \hline
\end{tabular}
\end{table}

\textbf{Evaluating partially annotated data on Open Images.} Open images \cite{OpenImages} is a significantly larger dataset than COCO, and differently has partial annotations, i.e. there are un-annotated objects in images. Table \ref{tbl:openimages} compares $\mathrm{AP_{50}}$, standard Open Images metric and oLRP Error on two different Faster R-CNN (R101 with atrous convolutions), both provided by the official tensorflow API \cite{tensorflow}, on the validation split. While one of the models uses 200 top-scoring proposals of RPN as the default setting for performance, the other one employs 30 proposals for efficiency. Note that (i) oLRP Error can consistently evaluate the performance of these models by assigning a lower (i.e. better) oLRP Error to the default model, (ii) when proposal\# decreases; while $\mathrm{oLRP_{FN}}$ degrades, $\mathrm{oLRP_{FP}}$ and $\mathrm{oLRP_{Loc}}$ improve, which is expected since the noisy proposals with lower scores are removed, (iii) between these models, the difference of oLRP Errors is not as large as $\mathrm{AP_{50}}$ ($0.7\%$ vs. $6.6\%$ gap) since unlike AP, oLRP Error does not favor detections with low precision for higher recall and considers the optimal combination of the performance aspects (see also ``s-LRP curves'' in Section \ref{subsec:oLRP} and \figref{\ref{fig:ClassPRCurves}}) and (iv) compared to object detection performance on COCO (Table \ref{tbl:perf_comparison}) with around 20 points difference between $\mathrm{oLRP_{FP}}$ and $\mathrm{oLRP_{FN}}$, the default model has only 8 point difference because some of the objects are not annotated implying more precision but less recall errors, thereby closing the gap.

\begin{table}
\setlength{\tabcolsep}{0.15em}
\small
\caption{Evaluating Faster R-CNN with R101 using 30 and 200 top-scoring proposals on Open Images val set. \label{tbl:openimages}} 
 \centering
\begin{tabular}{|c||c||c|c|c|c|}
\hline
proposal \# &$\mathrm{AP_{50}} \uparrow$&oLRP $\downarrow$&$\mathrm{oLRP_{Loc}} \downarrow$& $\mathrm{oLRP_{FP}} \downarrow$&$\mathrm{oLRP_{FN}} \downarrow$\\
\hhline{======}
200&$39.2$&$79.2$&$19.8$&$46.7$&$54.5$ \\ \hline
30&$32.6$&$79.9$&$19.1$&$41.0$&$60.2$ \\ \hline
\end{tabular}
\end{table}

\textbf{Evaluating more sparse objects on Pascal.} Pascal \cite{PASCAL} has less number of objects on average than COCO (2.4 vs. 7.3 objects/image) and generally the models perform better in Pascal compared to COCO. Table \ref{tbl:pascal} compares three different models wrt. $\mathrm{AP_{50}}$, as the standard measure of Pascal, and oLRP Error: Considering either pascal-api or coco-api results; (i) while Faster R-CNN (R50) outperforms RetinaNet by around $2\%$ wrt. $\mathrm{AP_{50}}$; they have similar oLRP Errors since $\mathrm{AP_{50}}$ loosely considers localisation quality and RetinaNet performs $2 \%$ better wrt. $\mathrm{oLRP_{Loc}}$, (ii) as expected, with a stronger backbone (R101), oLRP Error and components improve for Faster R-CNN, and (iii) similar to AP, oLRP Errors of these models on Pascal is better than those of COCO (Table \ref{tbl:perf_comparison}). These suggest that using oLRP Error on Pascal yields consistent evaluation and provides better insight than $\mathrm{AP_{50}}$.

\blockcomment{
\begin{table}
\setlength{\tabcolsep}{0.2em}
\small
\caption{Evaluating Pascal dataset. All models are trained on 2007+2012 trainval set and tested on 2007 test set. \label{tbl:pascal}} 
 \centering
\begin{tabular}{|c||c||c|c|c|c|}
\hline
& &\multicolumn{4}{|c|}{oLRP  $\downarrow$}\\\cline{3-6}
Model &$\mathrm{AP_{50}} \uparrow$&oLRP &$\mathrm{oLRP_{Loc}}$& $\mathrm{oLRP_{FP}}$&$\mathrm{oLRP_{FN}}$\\
\hhline{======}
RetinaNet+R50&$77.3$&$56.8$&$16.2$&$17.1$&$28.1$ \\ \hline
F.R-CNN+R50&$79.5$&$56.7$&$18.2$&$14.5$&$24.6$ \\ \hline
F. R-CNN+R101&$81.3$&$54.1$&$17.3$&$13.1$&$22.8$ \\ \hline
\end{tabular}
\end{table}
}
\textbf{Evaluating scale-based partitions on COCO.} Similar to scale-based APs, we compute oLRP Error over different scales, i.e. for small, medium and large objects. Table \ref{tbl:scaleeffect} compares three different scale imbalance methods, that are using deformable convolution in the last stage (DC5), feature pyramid network (FPN) \cite{FeaturePyramidNetwork} and recursive feature pyramid network (RFP) \cite{detectors}: (i) as expected, as the size of the objects increases, the performance wrt. oLRP Error improves similar to AP, (ii) Compared to DC5, FPN improves $\mathrm{oLRP_{S}}$ and $\mathrm{AP^C_{S}}$, while performing worse on $\mathrm{oLRP_{L}}$ and $\mathrm{AP^C_{L}}$ and (iii) RFP improves all scales wrt. AP and oLRP Error. Hence, oLRP Error can be used to evaluate performance over different scales.

\begin{table}
\setlength{\tabcolsep}{0.15em}
\small
\caption{Evaluating scale-based partitions of COCO \label{tbl:scaleeffect}} 
 \centering
\begin{tabular}{|c||c|c|c|c||c|c|c|c|}
\hline
&\multicolumn{4}{|c||}{AP $\uparrow$}&\multicolumn{4}{|c|}{oLRP Error $\downarrow$}\\\cline{2-9}
Method&$\mathrm{AP^C}$ &$\mathrm{AP^C_{S}}$&$\mathrm{AP^C_{M}}$&$\mathrm{AP^C_{L}}$&oLRP&$\mathrm{oLRP_{S}}$& $\mathrm{oLRP_{M}}$& $\mathrm{oLRP_{L}}$\\
\hhline{=========}
DC5 &$37.2$&$19.5$&$41.4$&$50.4$&$69.3$&$83.4$&$66.1$&$56.8$ \\ \hline
FPN&$37.8$&$21.6$&$41.5$&$49.3$&$69.1$&$82.2$&$66.1$&$58.2$ \\ \hline
RFP&$44.8$&$26.1$&$48.7$&$58.3$&$63.1$&$78.4$&$59.4$&$50.7$ \\ \hline
\end{tabular}
\end{table}

\blockcomment{
\begin{table}
\setlength{\tabcolsep}{0.2em}
\small
\caption{Effect of the absence of LRP Error components on oLRP measure. (Using Faster R-CNN on COCOval)} 
 \centering
\begin{tabular}{|c|c|c|c|c|c|c|}
\hline
\multicolumn{3}{|c|}{Performance Aspects}&\multirow{2}{*}{oLRP}&\multirow{2}{*}{$\mathrm{oLRP_{Loc}}$}&\multirow{2}{*}{$\mathrm{oLRP_{FP}}$}&\multirow{2}{*}{$\mathrm{oLRP_{FN}}$}\\\cline{1-3}
Loc. Error & FP rate & FN rate & & & & \\ \hline
\checkmark&\checkmark&\checkmark&$69.1$&$17.5$&$27.5$&$45.4$ \\ \hline
\xmark&\checkmark&\checkmark&$54.0$&$0.0$&$29.8$&$43.8$ \\ \hline
\checkmark&\xmark&\checkmark&$52.8$&$19.7$&$0.0$&$23.6$ \\ \hline
\checkmark&\checkmark&\xmark&$17.0$&$8.3$&$0.5$&$0.0$ \\ \hhline{=======}
\checkmark&\xmark&\xmark&$15.8$&$7.9$&$0.0$&$0.0$ \\ \hline
\xmark&\checkmark&\xmark&$0.4$&$0.0$&$0.4$&$0.0$ \\ \hline
\xmark&\xmark&\checkmark&$23.6$&$0.0$&$0.0$&$23.6$ \\ \hhline{=======}
\xmark&\xmark&\xmark&$0.0$&$0.0$&$0.0$&$0.0$ \\ \hline
\end{tabular}
\end{table}
}
\subsubsection{Evaluating Different Visual Detection Tasks}
\label{subsubsec:DifferentTasks}
\textbf{Visual relationship detection:} To show the usage of LRP Error for visual relationship, we use QPIC \cite{qpic} on V-COCO dataset \cite{v-coco}, which computes Role AP to determine the relationships among objects on two different styles, i.e. scenario 1 and scenario 2 (see V-COCO \cite{v-coco} for more details). Table \ref{tbl:VRD} presents that (i) as a simpler setting, the Role oLRP Error of scenario 1 is better for both backbones similar to Role AP, and (ii) Role oLRP Error and AP are similar for R50 and R101 for both scenarios. Therefore, oLRP Error can also be used to evaluate the visual relationship detection task. 

\begin{table}
\setlength{\tabcolsep}{0.2em}
\small
\caption{Evaluating visual relationship detection using QPIC on V-COCO. Sc.: Scenario (of V-COCO). \label{tbl:VRD}} 
 \centering
\begin{tabular}{|c|c||c||c|c|c|c|}
\hline
& &\multicolumn{1}{|c||}{Role AP $\uparrow$}&\multicolumn{4}{|c|}{Role oLRP Error $\downarrow$}\\\cline{3-7}
Type & Model &$\mathrm{AP_{50}}$&oLRP&$\mathrm{oLRP_{Loc}}$& $\mathrm{oLRP_{FP}}$& $\mathrm{oLRP_{FN}}$\\
\hhline{=======}
\multirow{2}{*}{\rotatebox{90}{Sc. 1}}&R50&$58.8$&$64.6$&$12.2$&$31.5$&$41.3$ \\ \cline{2-7}
&R101&$58.3$&$64.7$&$12.0$&$33.7$&$40.4$ \\ \hline
\multirow{2}{*}{\rotatebox{90}{Sc. 2}}&R50&$61.0$&$62.9$&$12.3$&$29.3$&$38.8$ \\ \cline{2-7}
&R101&$60.7$&$63.0$&$12.3$&$31.5$&$37.9$ \\ \hline
\end{tabular}
\end{table}

\textbf{Zero-shot detection (ZSD) and generalised zero-shot detection (GZSD)\footnote{While ZSD aims to detect unseen classes, GZSD includes detecting both seen and unseen classes.}:} Table \ref{tbl:ZSD} presents AP, AR and oLRP Error over different TP assignment thresholds ($\tau$) as conventionally done by ZSD and GZSD on background learnable cascade (BLC) \cite{BLC} with R50 using 48 classes of COCO as seen classes, and 17 classes as unseen classes following Zheng et al. \cite{BLC}. Following observations validate the usage of oLRP Error on ZSD and GZSD: (i) For both ZSD and GZSD, when $\tau$ increases and TPs are validated from higher localisation quality, oLRP Error degrades similar to AP and AR, (ii) for GZSD, the performance of seen classes is considerably better and (iii) similar to AP, the worst performance among all tasks is obtained for the unseen classes of GZSD (i.e. up to $96.2 \%$ oLRP Error). 

\begin{table}
\setlength{\tabcolsep}{0.35em}
\small
\caption{Evaluating ZSD and GZSD on COCO (i.e. 48/17 seen/unseen split) using BLC \cite{BLC}. $\mathcal{S}$: Seen classes, $\mathcal{U}$: Unseen classes for GZSD. \label{tbl:ZSD} } 
 \centering
\begin{tabular}{|c|c||c|c||c|c|c|c|}
\hline
& &\multicolumn{2}{|c||}{AP \& AR $\uparrow$}&\multicolumn{4}{|c|}{oLRP  Error $\downarrow$}\\\cline{3-8}
&$\tau$&$\mathrm{AP_\tau}$&$\mathrm{AR_{100}}$&oLRP&$\mathrm{oLRP_{Loc}}$& $\mathrm{oLRP_{FP}}$& $\mathrm{oLRP_{FN}}$\\
\hhline{========}
\multirow{3}{*}{\footnotesize{\rotatebox{90}{ZSD}}}&0.4&$11.8$&$51.3$&$91.1$&$20.9$&$78.7$&$73.7$ \\ \cline{2-8}
&0.5&$10.6$&$48.9$&$91.7$&$15.8$&$78.5$&$74.6$ \\ \cline{2-8}
&0.6&$9.7$&$45.0$&$92.5$&$13.8$&$79.8$&$75.9$ \\ \hline
\hhline{========}
\multirow{3}{*}{\rotatebox{90}{\footnotesize{GZSD - $\mathcal{S}$}}}&0.4&$44.0$&$59.7$&$73.4$&$17.5$&$37.4$&$55.7$ \\ \cline{2-8}
&0.5&$42.1$&$57.6$&$75.8$&$16.0$&$39.8$&$56.5$ \\ \cline{2-8}
&0.6&$39.3$&$54.0$&$78.6$&$14.3$&$42.1$&$59.2$ \\ \hline
\hhline{========}
\multirow{3}{*}{\rotatebox{90}{\footnotesize{GZSD - $\mathcal{U}$}}}&0.4&$4.9$&$49.6$&$95.4$&$22.7$&$89.3$&$76.7$ \\ \cline{2-8}
&0.5&$4.5$&$46.4$&$95.7$&$17.5$&$89.5$&$77.7$ \\ \cline{2-8}
&0.6&$4.1$&$41.9$&$96.2$&$15.0$&$90.4$&$79.0$ \\ \hline
\end{tabular}
\end{table}

\subsection{Thresholding Visual Detectors}
\label{subsec:ThresholdingExperiments}
In this section, we show that (i) the performances of visual detectors is sensitive to thresholding, (ii) the thresholds need to be set in a class-specific manner and (iii) LRP-Optimal thresholds can be used to alleviate this sensitivity.

Firstly, to see why visual object detectors can be sensitive to thresholding, \figref{\ref{fig:LRPConfScoreCurves}} shows on ``person'' and ``broccoli'' classes how  performance (in terms of LRP Error) evolves with different score thresholds ($s$) on different detectors. Note that the performances of some detectors (e.g ATSS, FCOS, aLRP Loss) improve and degrade rapidly around $s^*$, a situation which implies the sensitivity of these detectors with respect to the threshold choice (i.e. model selection). For example, for ATSS, choosing a threshold larger than $0.50$ has a significant impact on the performance, and even a threshold larger than $0.75$ results in a detector with no TPs (i.e. LRP Error $\approx 1$ in \figref{\ref{fig:LRPConfScoreCurves}}). Also, compare the detectors in \figref{\ref{fig:histogram}}(a) to see different one-stage detectors have very different LRP-Optimal threshold distributions. Thus, model selection is important for practical usage of visual detectors.

Secondly, for a given  detector, the variance of the LRP-Optimal thresholds over classes can be large  (\figref{\ref{fig:histogram}}- especially see RetinaNet in (a)). Thus, a general, fixed threshold for all classes can not provide  optimal performance for all classes. This is especially important for (i) rare classes in the dataset, which tend to have lower scores than frequent classes, and thus around $70 \%$ of such classes have $~0$ LRP-Optimal thresholds (\figref{\ref{fig:histogram}}(b)), and (ii) unseen classes in zero-shot detection and generalised zero-shot detection (\figref{\ref{fig:histogram}}(c)). Therefore,  class-specific thresholding is required for optimal performance of visual object detectors.


Based on these observations,  in Appendix \ref{subsec:ThresholdingExperiments2}, we present a use-case of LRP-Optimal thresholds on a video object detector which, first, collects  thresholded detections from a conventional object detector, and then associates  detections between frames. On this use-case, using class-specific LRP-Optimal thresholds significantly improves performance (up to around 9 points $\mathrm{AP}_{50}$ and 4 points oLRP Error) compared to using general, class-independent thresholding.


\blockcomment{
\begin{figure*}
        \captionsetup[subfigure]{}
        \centering
        \begin{subfigure}[b]{0.24\textwidth}
        \includegraphics[width=\textwidth]{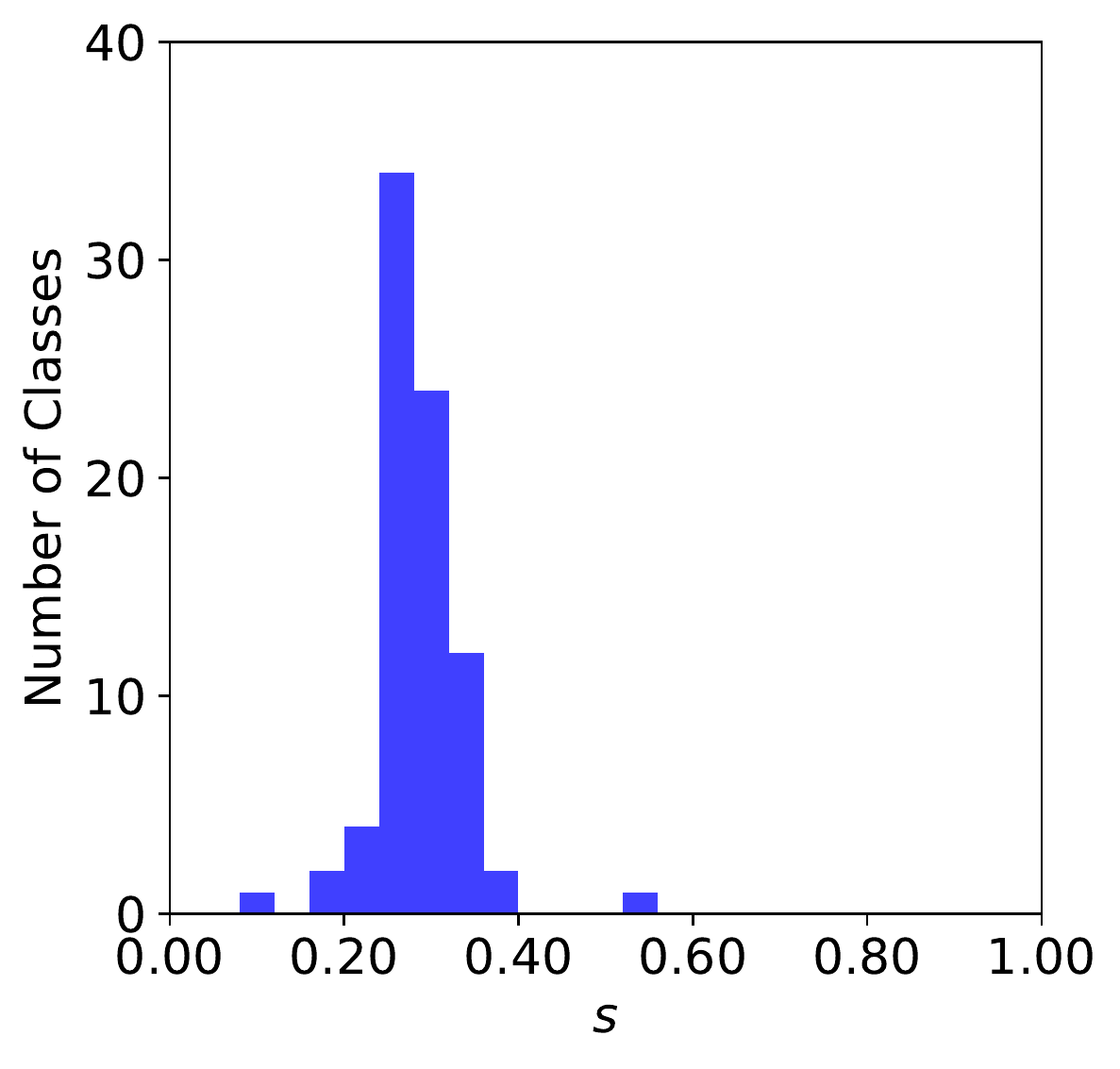}
        \caption{ATSS }
        \end{subfigure}
        \begin{subfigure}[b]{0.24\textwidth}
        \includegraphics[width=\textwidth]{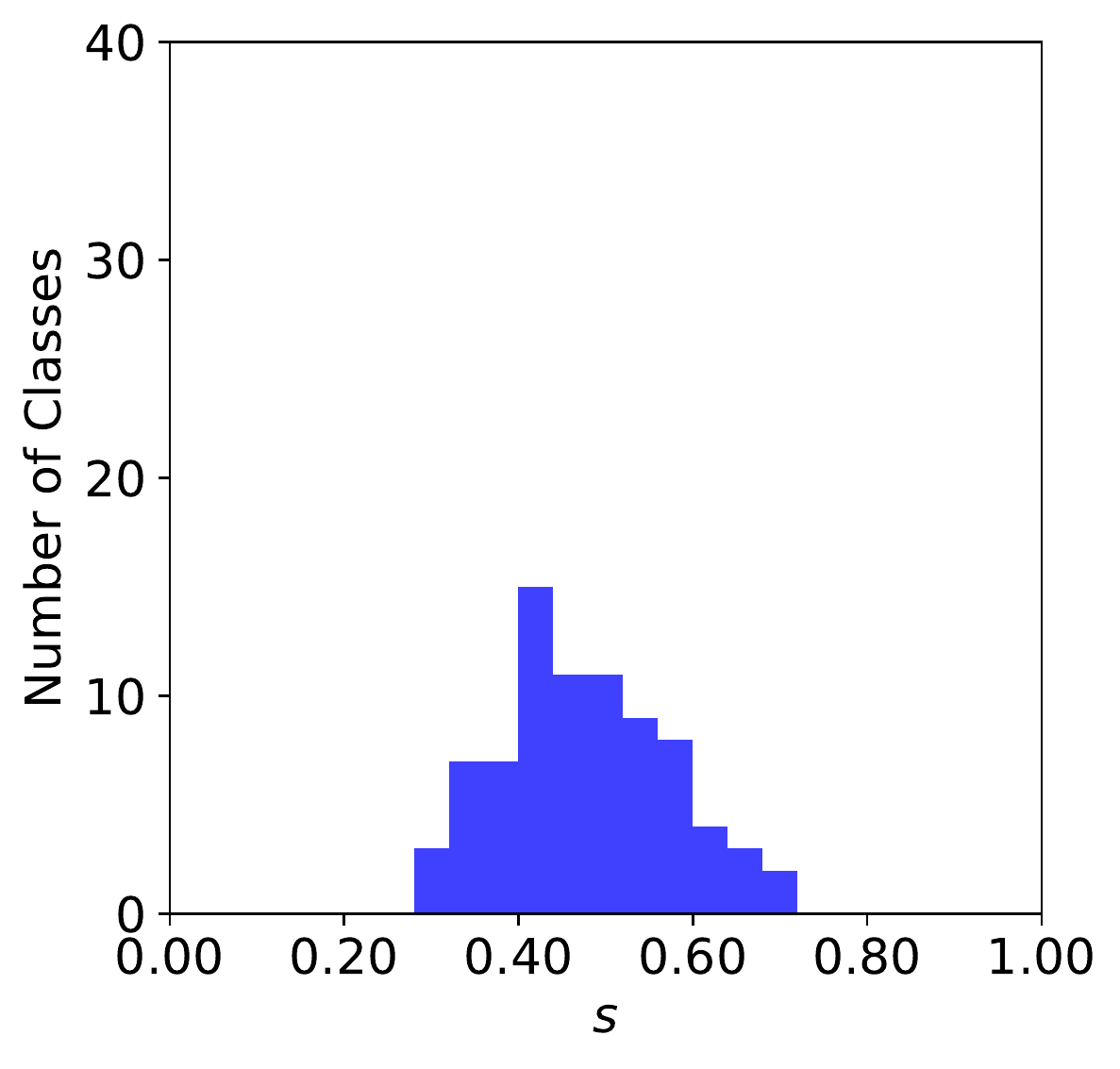} 
        \caption{RetinaNet}
        \end{subfigure}
        \begin{subfigure}[b]{0.24\textwidth}
        \includegraphics[width=\textwidth]{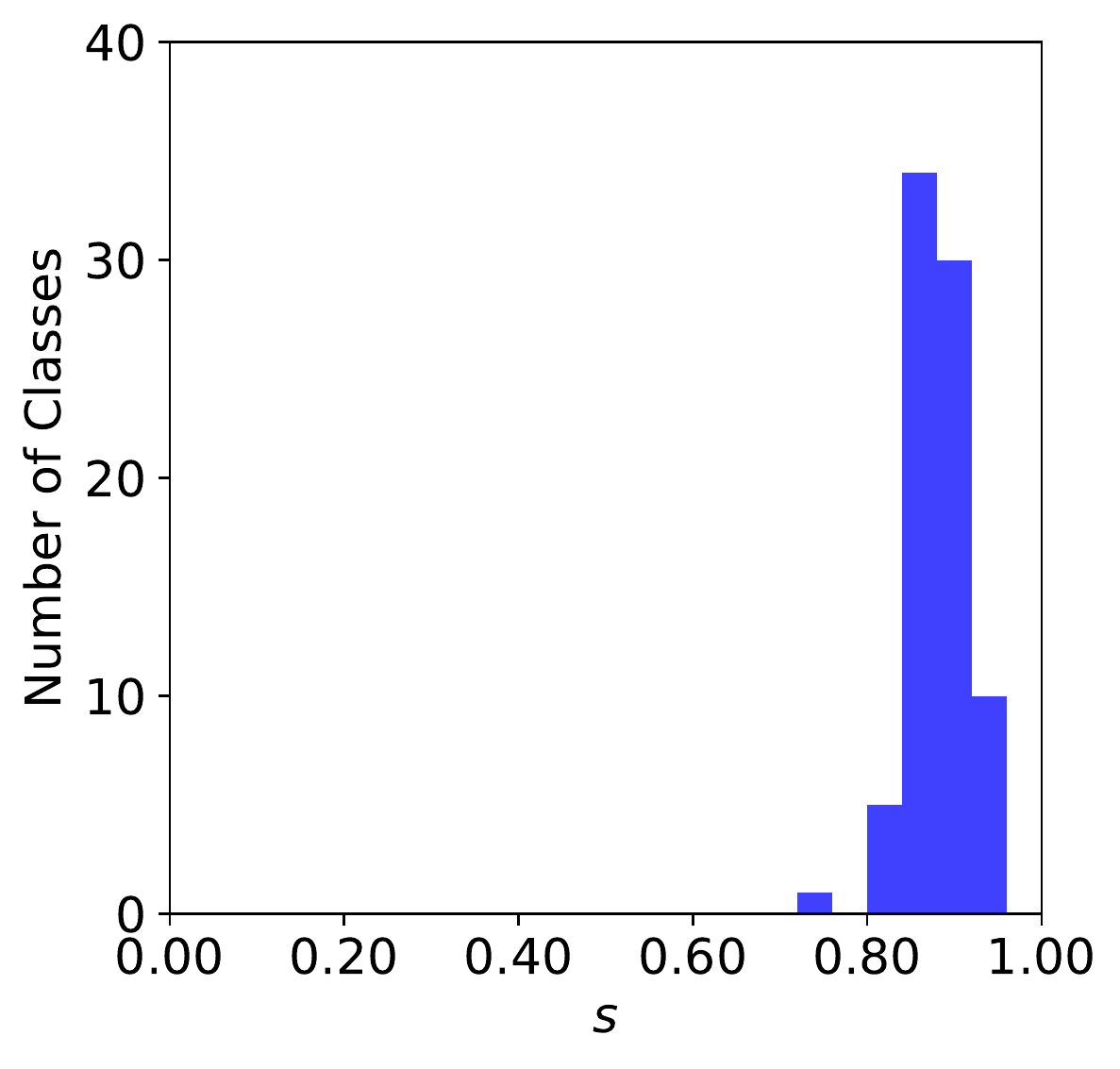}   
        \caption{aLRP Loss}
        \end{subfigure}
        \begin{subfigure}[b]{0.24\textwidth}
        \includegraphics[width=\textwidth]{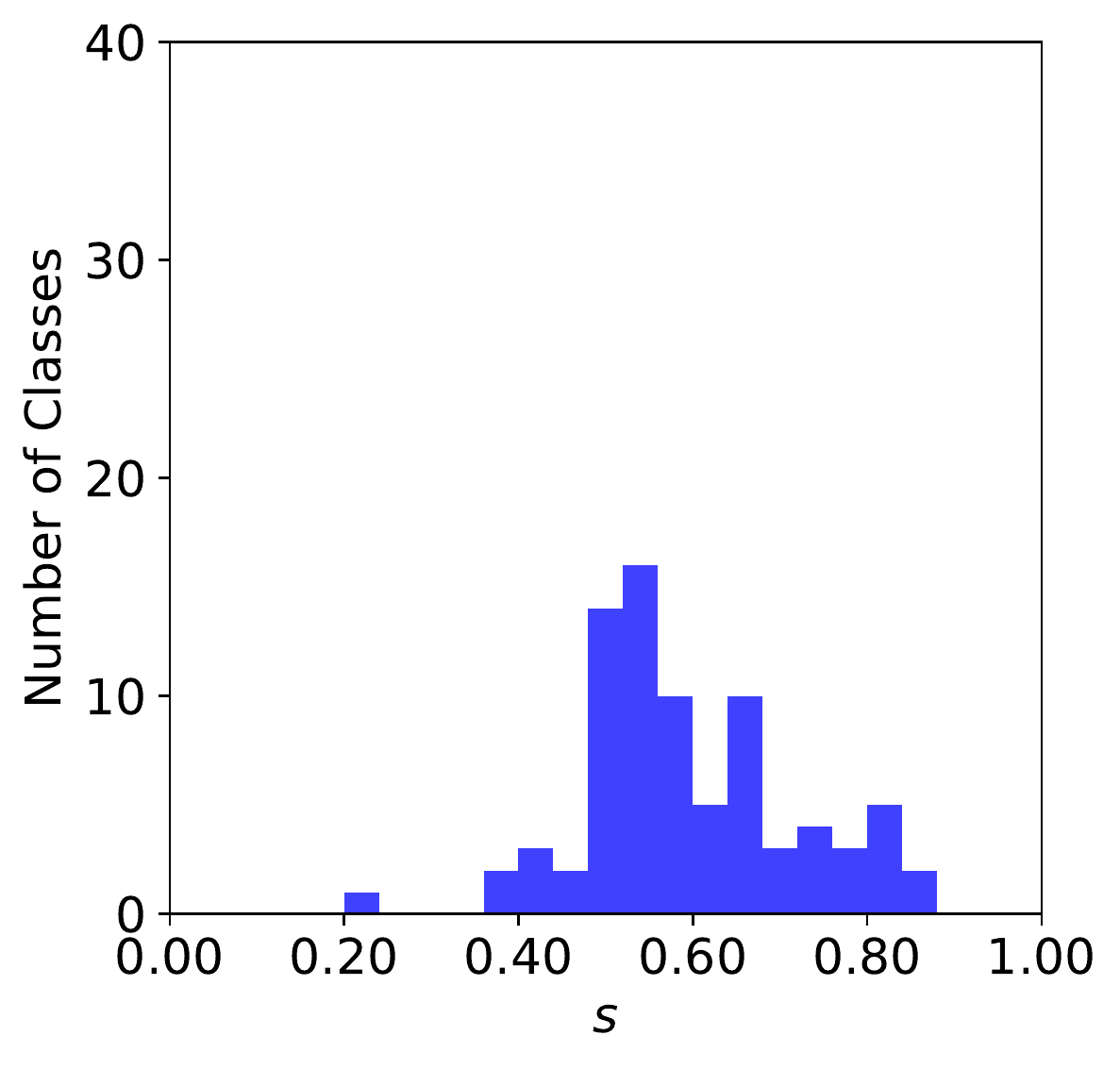}       
        \caption{Cascade R-CNN}
        \end{subfigure}
        \caption{The distributions of the class-specific LRP-Optimal Thresholds ($s^*$) for different methods. The variance of the LRP-Optimal thresholds can be large among classes. Thus, using a single general threshold for all classes will provide sub-optimal results.}
        \label{fig:histogram}
\end{figure*}
}
\begin{figure*}
        \captionsetup[subfigure]{}
        \centering
        \begin{subfigure}[b]{0.3\textwidth}
        \includegraphics[width=\textwidth]{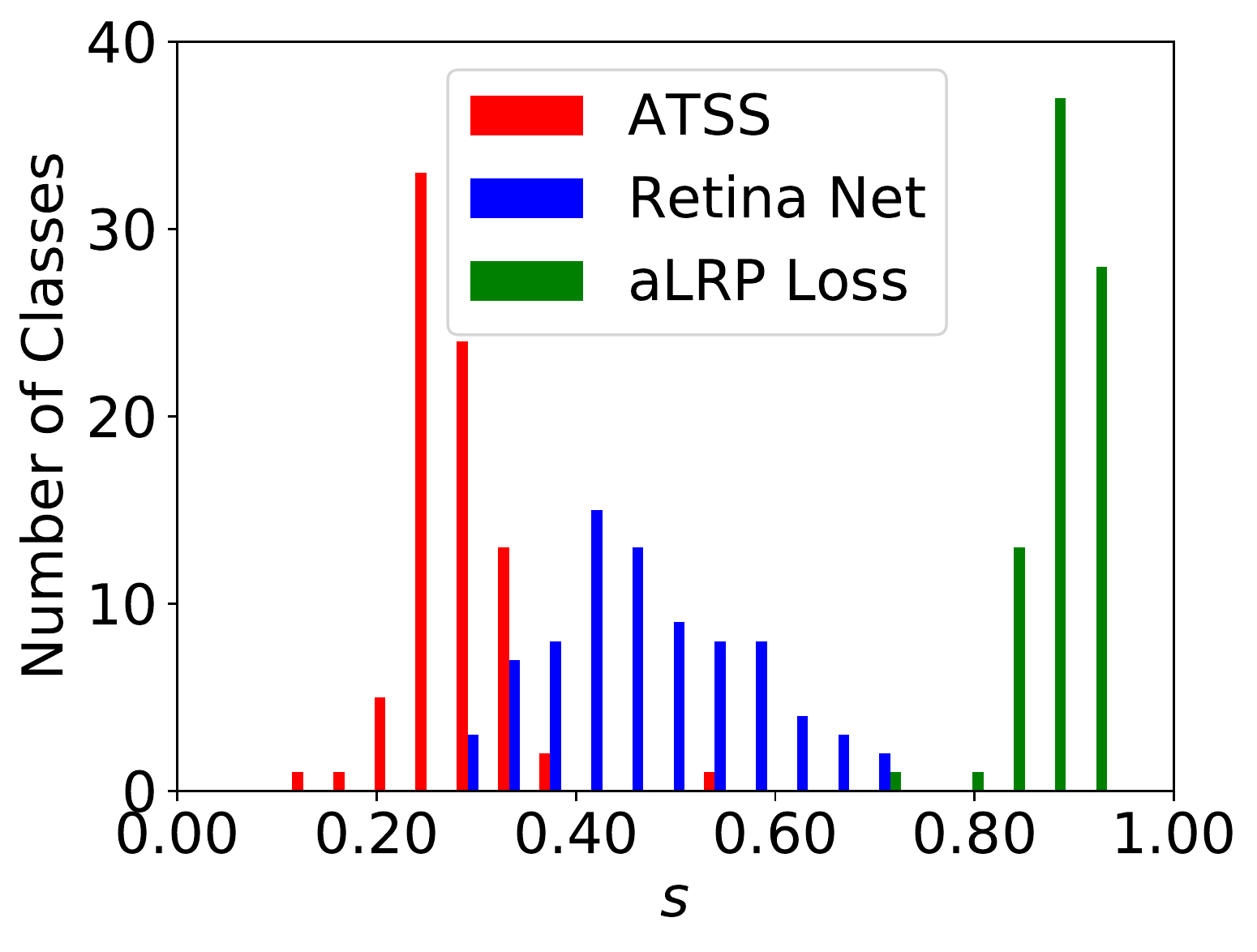}
        \caption{Different Methods on COCO}
        \end{subfigure}
        \begin{subfigure}[b]{0.3\textwidth}
        \includegraphics[width=\textwidth]{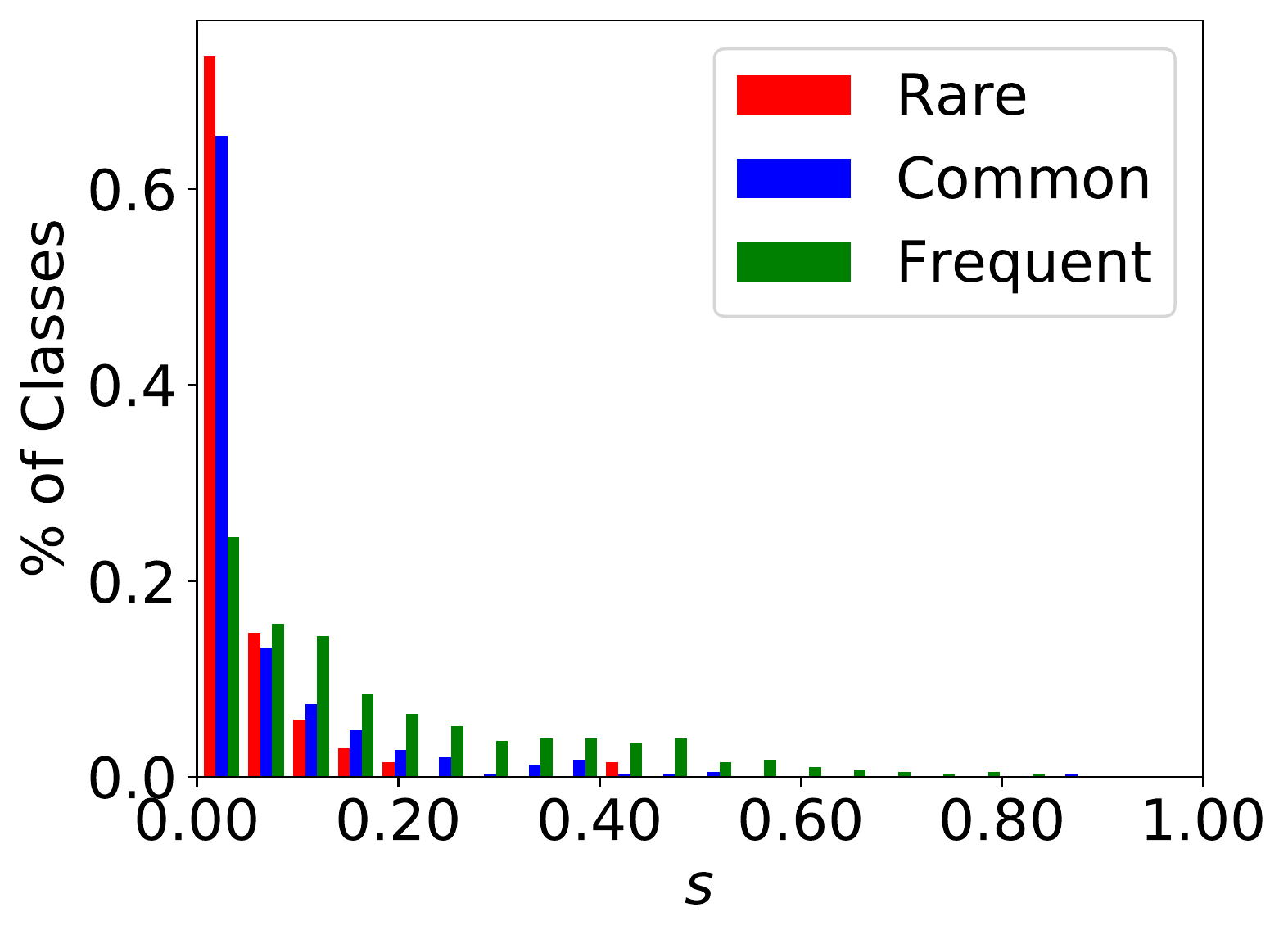} 
        \caption{Mask R-CNN+X101 on LVIS}
        \end{subfigure}
        \begin{subfigure}[b]{0.31\textwidth}
        \includegraphics[width=\textwidth]{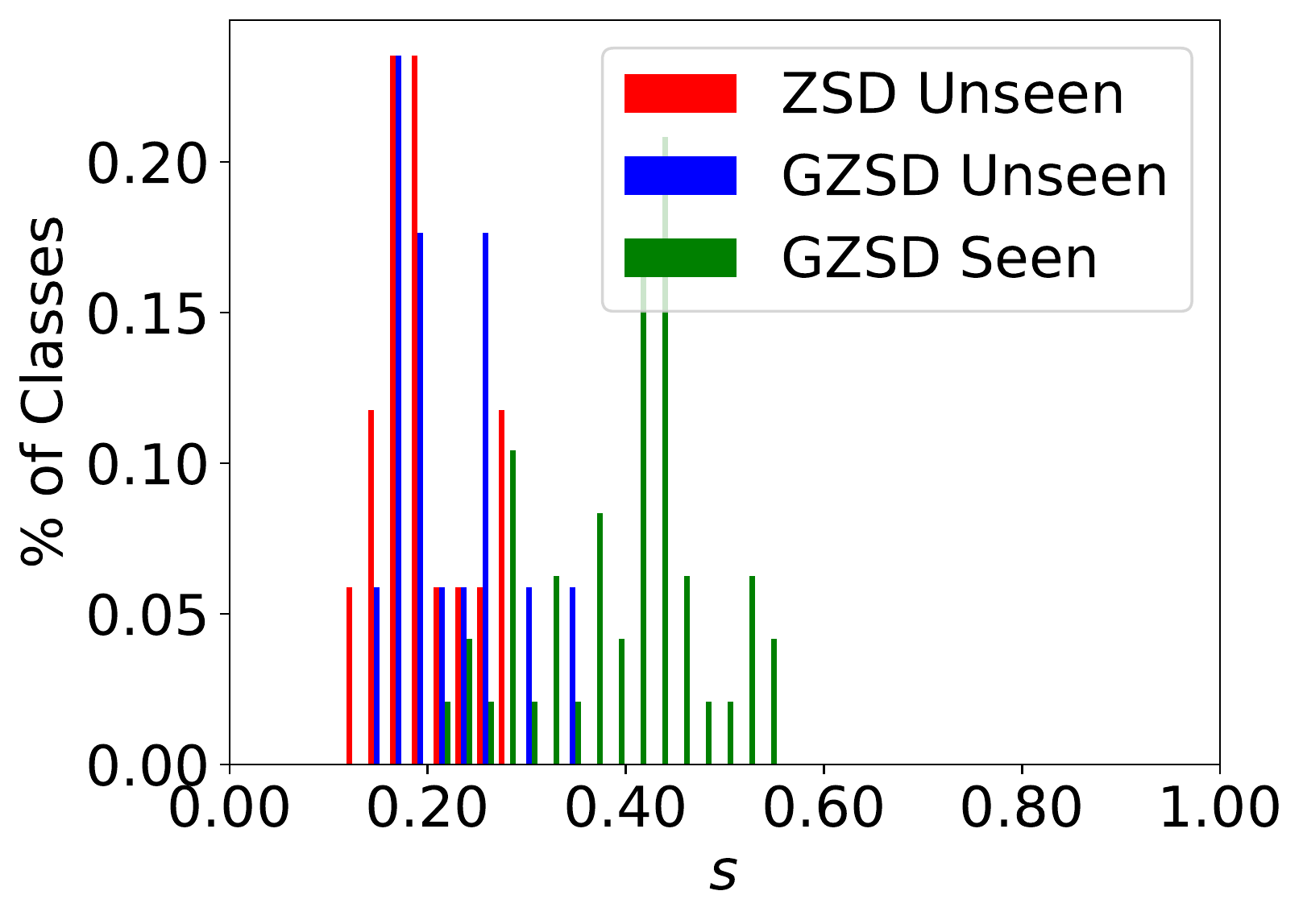}   
        \caption{BLC+R50 on ZSD and GZSD}
        \end{subfigure}
        \caption{The distributions of the class-specific LRP-Optimal thresholds ($s^*$) for different methods on different visual detection problems. (a) different methods on object detection, (b) rare, common and frequent classes on long-tailed instancce segmentation, and (c) unseen classes on zero-shot detection (ZSD), seen and unseen classes on generalised ZSD (GZSD). The variance of the LRP-Optimal thresholds can be large among classes, the distribution of LRP-Optimal thresholds can vary among methods and classes. Thus, using a single general threshold for all classes will provide sub-optimal results.}
        \label{fig:histogram}
\end{figure*}

\section{Conclusion}
\label{section:conclusions}
In this paper, we introduced a novel performance metric, LRP Error, as the average matching error of a visual detector, to evaluate \textit{all} visual detection tasks as an alternative to the widely-used measures AP and PQ. LRP Error has a number of advantages which we demonstrated in the paper: LRP Error (i) is ``complete'' in the sense that it precisely considers all important performance aspects (i.e. localization quality, recall, precision) of visual detectors, (ii) is easily interpretable through its components, and (iii) does not suffer from the practical drawbacks of AP and PQ. 

\textbf{Appendices:} This paper is accompanied with appendices, containing the definitions of the frequently used terms and notation; proofs showing that PQ is not a metric but LRP is; a discussion on weighting LRP Error due to practical needs; why average LRP is not suitable as a performance metric; the derivation of the similarity of PQ and LRP; the repositories of the models; and more experiments with LRP Error.

\section*{Acknowledgments}
This work was supported by the Scientific and Technological Research Council of Turkey (T\"UB\.{I}TAK) 
(under grants 117E054 and 120E494). We also gratefully acknowledge 
the computational resources kindly provided by T\"UB\.{I}TAK ULAKBIM High Performance and Grid Computing Center (TRUBA) and Roketsan Missiles Inc. used for this research. Dr. Oksuz is supported by the T\"UB\.{I}TAK 2211-A Scholarship and Dr. Kalkan by the BAGEP Award of the Science Academy, Turkey.

\bibliographystyle{IEEEtran}
\bibliography{proposalbibliography}
\section*{APPENDICES}

\renewcommand{\thefigure}{A.\arabic{figure}}
\renewcommand{\thetable}{A.\arabic{table}}
\renewcommand{\thetheorem}{A.\arabic{theorem}}
\renewcommand{\theequation}{A.\arabic{equation}}
\renewcommand{\thealgorithm}{A.\arabic{algorithm}}
\renewcommand{\thesection}{A}
\section{Frequently Used Terms and Notation}
\label{sec:DefNot}

Table \ref{tab:notation} presents the notation used throughout the paper, and below is a list of frequently used terms.
\begin{table}
\centering
\caption{Frequently used notations in the paper. \label{tab:notation}}
\renewcommand{\arraystretch}{0.6}\footnotesize
    \begin{tabular}{|>{\centering\arraybackslash}m{.15\linewidth}|>{\centering\arraybackslash}m{.7\linewidth}|}
         \hline
         \textbf{Symbol} &\textbf{Denotes} \\ \hline
         $\mathrm{AP^C}$&COCO-Style AP\\ \hline
         $\mathrm{AP}_\tau$&AP when the TPs are validated from the $\mathrm{lq}(\cdot,\cdot)$ threshold of $\tau$\\ \hline
         $d$&A detection such that $d \in \mathcal{D}$ \\ \hline
         $d_g$&A TP detection that matches ground truth $g$ and qualifies for performance evaluation\\ \hline
         $\mathcal{D}$&A set of detections\\ \hline
         FN&False Negative\\ \hline
         FP&False Positive\\ \hline
         $g$&A ground truth such that $g \in \mathcal{G}$ \\ \hline
         $\mathcal{G}$&A set of ground truths\\ \hline
         $\mathrm{lq}(\cdot, \cdot)$&A localisation quality function. (e.g. $\mathrm{IoU_B}(\cdot, \cdot)$) \\ \hline
         $\mathrm{N_{FN}}$&Number of FNs\\ \hline
         $\mathrm{N_{FP}}$&Number of FPs\\ \hline
         $\mathrm{N_{TP}}$&Number of TPs\\ \hline
         $s^*$&LRP-Optimal confidence score\\ \hline
         TP&True Positive\\ \hline
         $\tau$&TP validation threshold in terms of localisation quality\\ \hline
         
    \end{tabular}
\end{table}

\noindent \textbf{Hard Prediction:} A type of visual object detector output which identifies each object with (i) a set of identifiers to locate an object (e.g. bounding box, mask, keypoints), and (ii) its class label.

\noindent \textbf{Soft Prediction:} A type of visual object detector output which identifies each object with (i) a set of identifiers to locate an object (e.g. bounding box, mask, keypoints), (ii) its class label and (iii) the confidence score of the prediction.
 
\noindent \textbf{Bounding Box:} A rectangle on the image. Formally, a bounding box, denoted by $B$, is generally represented by $[x_1,y_1,x_2,y_2]$ with  $(x_1,y_1)$ denoting the top-left corner and  $(x_2,y_2)$ the bottom-right corner, with the constraints $x_2>x_1$ and $y_2>y_1$. 

\noindent \textbf{Keypoint Set:} A set of coordinates to represent an object on the image such that each element is a two tuple $(x_i, y_i)$ identifying a keypoint of an object. 

\noindent \textbf{Segmentation Mask:} A set of pixels presenting which pixels belong to a particular object. 

\noindent \textbf{Intersection Over Union (IoU):} For two polygons $P_g$ and $P_d$, Intersection over Union (IoU) \cite{PASCAL,ImageNet}, $\mathrm{IoU}(P_g,P_d)$ is defined as :
\begin{align}
\label{eq:IoU}
\mathrm{IoU}(P_g,P_d)&=\frac{\mathrm{A}(P_g \cap P_d)}{\mathrm{A}(P_g \cup P_d)},
\end{align}
where $\mathrm{A}(P)$ is the area of the polygon $P$ (i.e. number of pixels delimited by $P$). These polygons are represented by bounding boxes for object detection, and by masks for segmentation tasks (e.g. instance segmentation, panoptic segmentation). $\mathrm{IoU} \in [0,1]$ and it is used to evaluate the localisation quality of a detection bounding box/mask with respect to its ground truth box/mask.

\noindent \textbf{Object Keypoint Similarity (OKS):} Given target and estimated keypoint sets, $K_g$ and $K_d$ respectively, with $k_g^i \in K_g$ and $k_d^i \in K_d$ being the $i$th keypoint of the object represented by a 2D coordinate on the image, Object Keypoint Similarity (OKS) \cite{COCO} between $k_g^i$ and $k_d^i$, denoted by $\mathrm{OKS}(k_g^i,k_d^i)$, is
\begin{align}
\label{eq:OKSSingle}
\mathrm{OKS}(k_g^i,k_d^i) =  \frac{ \exp(-||k_g^i- k_d^i||^2)}{2 (S \kappa^i)^2},
\end{align}
where $|| \cdot - \cdot||$ is Euclidean distance, $\kappa^i$ is the constant corresponding to $i$th keypoint to control falloff, and $S$ is the object scale (e.g. the area of the ground truth bounding box divided by the total image area). Then, OKS between $K_g$ and $K_d$, $\mathrm{OKS}(K_g,K_d)$, is simply the average of $\mathrm{OKS}(k_g^i,k_d^i)$ over single keypoints annotated in the dataset:
\begin{align}
\label{eq:OKS}
\mathrm{OKS}(K_g,K_d) =  \frac{ 1}{|K_g|}\sum \limits_{i \in |K_g|} \mathrm{OKS}(k_g^i,k_d^i).
\end{align}
Similar to $\mathrm{IoU}(\cdot,\cdot)$; $\mathrm{OKS}(\cdot,\cdot) \in [0,1]$ and a larger $\mathrm{OKS}(\cdot,\cdot)$ implies better localisation quality.

\renewcommand{\thesection}{B}
\section{Proof that PQ Error is not a Metric}
\label{sec:PQMetricity}
\begin{figure}
    \centerline{
        \includegraphics[width=0.5\textwidth]{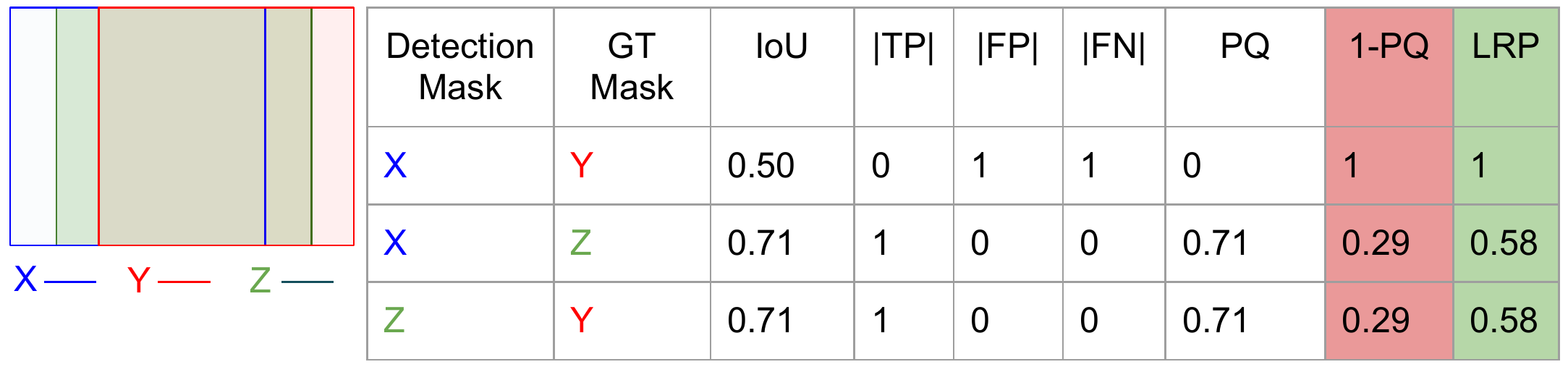}
    }
\caption{A counter-example which shows that PQ Error (i.e., 1-PQ) violates triangle inequality. Hence, PQ Error is not a metric. (a) Three different inputs (i.e. masks) $X$, $Y$ and $Z$. (b) $1-\mathrm{PQ}$ does not satisfy the triangle inequality (i.e. $1-\mathrm{PQ}(X,Y) > 1-\mathrm{PQ}(X,Z) + 1-\mathrm{PQ}(Z,Y)$), while $\mathrm{LRP}$ does (i.e. $\mathrm{LRP}(X,Y) \leq \mathrm{LRP}(X,Z) + \mathrm{LRP}(Z,Y)$). See Table \ref{tab:notation} for the notation.
}
\label{fig:PQMetricity}
\end{figure}

Here, we prove that  PQ Error (i.e. (1-PQ)) is not a metric due to the fact that PQ Error violates triangle inequality\footnote{PQ satisfies the other two metricity conditions, i.e. reflexivity and symmetry.}:

\begin{theorem}
PQ Error, defined by $1-\mathrm{PQ}(\mathcal{G},\mathcal{D})$, violates triangle inequality, and hence it is not a metric.
\end{theorem}
\begin{proof}
This is a proof by counter-example. If $1-\mathrm{PQ}(\mathcal{G},\mathcal{D})$ satisfied triangle inequality, then we would expect $\forall \mathcal{A} \forall \mathcal{B} \forall \mathcal{C}$ $1-\mathrm{PQ}(\mathcal{A},\mathcal{B}) \leq 1-\mathrm{PQ}(\mathcal{A},\mathcal{C}) + 1-\mathrm{PQ}(\mathcal{C},\mathcal{B})$. However, \figref{\ref{fig:PQMetricity}} presents a counterexample, therefore PQ Error ($1-\mathrm{PQ}(\mathcal{G},\mathcal{D})$) violates triangle inequality and it is not a metric.
\end{proof}

\renewcommand{\thesection}{C}
\section{Proof that LRP is a Metric}
\label{section:appendix}
In this section, we prove that, unlike PQ Error (Section \ref{sec:PQMetricity}), LRP Error is a metric if the localisation error is a metric (i.e. $1-\mathrm{lq}(\cdot,\cdot)$). In our proof, we obtain LRP Error using a reduction from Deficiency Aware Subpattern Assignment (DASA) performance metric \cite{DASA} from point multitarget tracking literature. Note that DASA is a proven metric. 
\begin{theorem}
LRP is a metric.
\end{theorem}
\begin{proof}
DASA metric is defined as:
\begin{align}
\label{eq:DASA1} 
\bar{e}^{(c)}_p(\mathcal{G},\mathcal{D})&:= \frac{l}{Z} \left( \frac{\mathrm{N_{TP}}}{l} \left(\frac{1}{\mathrm{N_{TP}}} \sum \limits_{i=1}^{|\mathcal{G}|} \mathbb{I}[\mathrm{d}(g_i, d_{g_i}) < c]  \mathrm{d}(g_i, d_{g_i})^p \right) \right. \\ 
& \left. +\left(\frac{c^p  \mathrm{N_{FP}}}{l} \right) +\left( \frac{c^p  \mathrm{N_{FN}}}{l} \right) \right)^{1/p}, \nonumber
\end{align}
where $l=\max(\mathcal{G},\mathcal{D})$, $Z = \mathrm{N_{TP}}+\mathrm{N_{FP}}+\mathrm{N_{FN}}$, $\mathbb{I}[\cdot]$ is the indicator function, $\mathrm{d}(g_i, d_{g_i})$ is an arbitrary metric, $c$ is the cut-off length to validate TPs (i.e. TP assignment threshold) based on $\mathrm{d}(g_i, d_{g_i})$, and finally $p$ is the lp-norm parameter (see Table \ref{tab:notation} for the rest of the notation)
\footnote{Unlike DASA \cite{DASA} using $\mathbb{I}[\mathrm{d}(g_i, d_{g_i}) \leq c]$, we use $\mathbb{I}[\mathrm{d}(g_i, d_{g_i}) < c]$ following OSPA \cite{OSPA}.}.

First, we set the lp-norm parameter, $p$, as $1$ and simplify the definition:
\begin{align}
\label{eq:DASA2} 
\frac{1}{Z} \left( \left(\sum \limits_{i=1}^{|\mathcal{G}|} \mathbb{I}[\mathrm{d}(g_i, d_{g_i}) < c]  \mathrm{d}(g_i, d_{g_i}) \right)+c  \mathrm{N_{FP}} + c  \mathrm{N_{FN}} \right). 
\end{align}
Second, we incorporate the TP validation criterion of visual object detectors in \equref{\ref{eq:DASA2}} as follows. A TP 
is identified if a ground truth, $g_i$, has a corresponding detection $d_{g_i}$ such that  $\mathrm{lq}(g_i, d_{g_i}) > \tau$. Note that  $\mathrm{lq}(g_i, d_{g_i}) \in [0,1]$. Then, to have a lower-better criterion which fits into \equref{\ref{eq:DASA2}}, we can rewrite this TP validation criterion as $1-\mathrm{lq}(g_i, d_{g_i}) < 1-\tau$. Having obtained the TP criterion, we set $\mathrm{d}(g_i, d_{g_i})=1-\mathrm{lq}(g_i, d_{g_i})$ and $c=1-\tau$ in \equref{\ref{eq:DASA2}}:
\begin{align}
\label{eq:DASA3} 
&\frac{1}{Z} \left(\left(\sum \limits_{i=1}^{|\mathcal{G}|}  \mathbb{I}[1-\mathrm{lq}(g_i, d_{g_i}) < 1-\tau]  (1 - \mathrm{lq}(g_i, d_{g_i})) \right)\right. \\ 
& \left. + (1-\tau) \mathrm{N_{FP}} +(1-\tau)\mathrm{N_{FN}} \right), \nonumber
\end{align}
which can be rewritten by simplifying the predicate of $\mathbb{I}[\cdot]$ as follows:
\begin{align}
\label{eq:DASA4} 
&\frac{1}{Z} \left(\left(\sum \limits_{i=1}^{|\mathcal{G}|}  \mathbb{I}[\mathrm{lq}(g_i, d_{g_i}) > \tau]  (1 - \mathrm{lq}(g_i, d_{g_i})) \right)\right. \\ 
& \left. + (1-\tau) \mathrm{N_{FP}} +(1-\tau)\mathrm{N_{FN}} \right). \nonumber
\end{align}

Next, we just simplify \equref{\ref{eq:DASA4}} in two steps: (i) We remove the Iverson Bracket  by replacing $|\mathcal{G}|$ by $\mathrm{N_{TP}}$ in the summation,
\begin{align}
\label{eq:DASA5} 
&\frac{1}{Z} \left(\left(\sum \limits_{i=1}^{\mathrm{N_{TP}}}  (1 - \mathrm{lq}(g_i, d_{g_i})) \right) + (1-\tau) \mathrm{N_{FP}} +(1-\tau)\mathrm{N_{FN}} \right),
\end{align}
and (ii) finally, noting that dividing by a constant does not violate metricity, in order to ensure the upper bound to be $1$ and facilitate the interpretation of LRP, we divide \equref{\ref{eq:DASA5}} by $1-\tau$:
\begin{align}
\label{eq:DASA6} 
&\frac{1}{Z} \left(\sum \limits_{i=1}^{\mathrm{N_{TP}}}  \frac{1 - \mathrm{lq}(g_i, d_{g_i})}{1-\tau} + \mathrm{N_{FP}} +\mathrm{N_{FN}} \right) = \mathrm{LRP}(\mathcal{G},\mathcal{D}).
\end{align}
To conclude, LRP can be reduced from DASA, a proven metric, and therefore LRP is a metric.
\end{proof}

\renewcommand{\thesection}{D}
\section{Weighting the components of the LRP Error for Practical Needs of Different Applications}
\label{app:weight}
LRP Error does not give priority to any of the performance aspects (FP rate, FN rate and localisation error) and weights each performance aspect by considering their maximum possible contribution to the total matching error (Section \ref{subsec:LRP}). On the other hand, depending on the requirements in a given application, one of the performance aspects can be given an emphasis. To illustrate a use-case, an online video object detector may want to gather as much detections as it can after discarding the ``noisy'' examples of the still image detector. Note that removing the noisy examples still requires some thresholding, however, the conventional LRP-Optimal threshold, balancing the contribution of FPs and FNs, may not be the best solution to fulfill this requirement, and a lower threshold can be more suitable. In a different use-case, different weights for the components can also be beneficial for evaluation as well. For example, a ballistic missile detector may not tolerate FNs but can accept more FPs errors. To this end, \equref{\ref{eq:LRPGeneralized1}} presents a weighted form of LRP Error:
\begin{align}
\label{eq:LRPGeneralized1}
\frac{1}{Z} \left( \sum \limits_{i=1}^{\mathrm{N_{TP}}} \mathrm{\alpha_{TP}} \frac{1-\mathrm{lq}(g_i, d_{g_i})}{1-\tau}+ \mathrm{\alpha_{FP}} \mathrm{N_{FP}} + \mathrm{\alpha_{FN}} \mathrm{N_{FN}} \right),
\end{align}
where $Z = \mathrm{\alpha_{TP}} \mathrm{N_{TP}}+ \mathrm{\alpha_{FP}}\mathrm{N_{FP}}+ \mathrm{\alpha_{FN}}\mathrm{N_{FN}}$, and $\mathrm{\alpha_{TP}}$, $\mathrm{\alpha_{FP}}$ and $\mathrm{\alpha_{FN}}$ correspond to the ``importance weights'' of each performance aspect. Following the interpretation of LRP (see Section \ref{section:LRP}), the importance weights imply duplicating each error by the value of this weight. Accordingly, they are included both in the ``total matching error'' (i.e. nominator) and the ``maximum possible value of the total matching error'' (i.e. normalisation constant). In order to increase the contribution of a component, the importance weight of the desired component is to be set larger than $1$ (e.g. to double the contribution of false negatives, then $\mathrm{\alpha_{FN}}=2$ and $\mathrm{\alpha_{TP}}=\mathrm{\alpha_{FP}}=1$). Note that when $\mathrm{\alpha_{TP}}=\mathrm{\alpha_{FP}}=\mathrm{\alpha_{FN}}=1$, \equref{\ref{eq:LRPGeneralized1}} reduces to the conventional definition of LRP (\equref{\ref{eq:LRPdefcompact}}). Finally, we note that this modification naturally violates the symmetry of the metric properties when $\mathrm{\alpha_{FP}} \neq \mathrm{\alpha_{FN}}$.
\renewcommand{\thesection}{E}
\section{Why Average LRP (aLRP) is not an Ideal Performance Measure?}
\label{app:aLRP}
This section discusses why the recently proposed loss function aLRP Loss \cite{aLRPLoss} is not an ideal performance measure. 

Having a similar intuition to oLRP, we   define aLRP Error by averaging the LRP Errors over the confidence scores (see Table \ref{tab:notation} for the notation):
\begin{align}
\label{eq:AverageLRP}
\mathrm{aLRP}:= \frac{1}{|\mathcal{S}|}\sum \limits_{s \in \mathcal{S}} \mathrm{LRP}(\mathcal{G},\mathcal{D}_s).
\end{align}
where $\mathcal{D}_s$ is the set of detections thresholded at confidence score $s$ (i.e. those detections with larger confidence scores than $s$ are kept, and others are discarded). However, without any improvement in the detection performance, aLRP Error can be reduced to oLRP Error with the following two steps:
\begin{enumerate}
    \item Delete all the detections with $s < s^*$ from the detection output,
    \item Set the confidence score of the remaining detections to $1.00$.
\end{enumerate}
This two-step simple algorithm will make the s-LRP curve to be a line determined by $s=s^*$ (see \figref{\ref{fig:LRPVariants}}), and averaging over the confidence scores will yield oLRP. As a result, considering the fact that the performance with respect to aLRP is affected without any improvement in the detection performance, we do not prefer aLRP Error as a performance measure.
\renewcommand{\thesection}{F}
\section{The Similarity between PQ and LRP Errors}
\label{app:similarity}
In this section, we derive \equref{\ref{eq:PQvsLRP}}: 
\begin{align}
        \label{eq:PQvsLRP2}
        1 -\mathrm{PQ} =  \frac{1}{\hat{Z}} \left( \sum \limits_{i=1}^{\mathrm{N_{TP}}}  \frac{1-\mathrm{lq}(g_i, d_{g_i})}{1-0.50}+\mathrm{N_{FP}} +\mathrm{N_{FN}} \right) ,
\end{align}
where $\hat{Z} = {\textcolor{red}{2} \mathrm{N_{TP}}+\mathrm{N_{FP}} +\mathrm{N_{FN}}}$. LRP and PQ Errors are very similar:  Removing $2$ (in red) from $\hat{Z}$ yields $1-\mathrm{PQ} = \mathrm{LRP}$. 

Recall from \equref{\ref{eq:PQ}} that PQ is defined as (see Table \ref{tab:notation} for the notation):
\begin{align}
\label{eq:PQ_}
\mathrm{PQ}(\mathcal{G},\mathcal{D}) = \frac{1 }{{\mathrm{N_{TP}}+ \frac{1}{2} \mathrm{N_{FP}} + \mathrm{\frac{1}{2} N_{FN}}}} \left( \sum \limits_{i=1}^{\mathrm{N_{TP}}} \mathrm{IoU}(g_i, d_{g_i}) \right).
\end{align}
First, we replace $\mathrm{IoU}(\cdot, \cdot)$ in \equref{\ref{eq:PQ_}} by $\mathrm{lq}(\cdot, \cdot)$ to align the definitions of LRP and PQ:
\begin{align}
\label{eq:PQ2}
\mathrm{PQ}(\mathcal{G},\mathcal{D}) = \frac{1 }{{\mathrm{N_{TP}}+ \frac{1}{2} \mathrm{N_{FP}} + \mathrm{\frac{1}{2} N_{FN}}}} \left( \sum \limits_{i=1}^{\mathrm{N_{TP}}} \mathrm{lq}(g_i, d_{g_i}) \right).
\end{align}
Then, just by simple algebraic operations, we manipulate \equref{\ref{eq:PQ2}}:
\begin{align}
    \mathrm{PQ} &= 1 - 1 + \frac{\sum \limits_{i=1}^{\mathrm{N_{TP}}} \mathrm{lq}(g_i, d_{g_i}) }{\mathrm{N_{TP}} + \frac{1}{2} \mathrm{N_{FP}}+ \frac{1}{2} \mathrm{N_{FN}}} \\
    &= 1 - \left( 1 - \frac{\sum \limits_{i=1}^{\mathrm{N_{TP}}} \mathrm{lq}(g_i, d_{g_i}) }{\mathrm{N_{TP}} + \frac{1}{2} \mathrm{N_{FP}}+ \frac{1}{2} \mathrm{N_{FN}}} \right) \\
    &= 1 -  \frac{ \mathrm{N_{TP}} + \frac{1}{2} \mathrm{N_{FP}}+ \frac{1}{2} \mathrm{N_{FN}} - \sum \limits_{i=1}^{\mathrm{N_{TP}}} \mathrm{lq}(g_i, d_{g_i}) }{\mathrm{N_{TP}} + \frac{1}{2} \mathrm{N_{FP}}+ \frac{1}{2} \mathrm{N_{FN}}} \\
    &= 1 -  \frac{ \mathrm{N_{TP}}  - \sum \limits_{i=1}^{\mathrm{N_{TP}}} \mathrm{lq}(g_i, d_{g_i})+ \frac{1}{2} \mathrm{N_{FP}}+ \frac{1}{2} \mathrm{N_{FN}} }{\mathrm{N_{TP}} + \frac{1}{2} \mathrm{N_{FP}}+ \frac{1}{2} \mathrm{N_{FN}}}\\
    &= 1 -  \frac{ \sum \limits_{i=1}^{\mathrm{N_{TP}}} 1  - \sum \limits_{i=1}^{\mathrm{N_{TP}}} \mathrm{lq}(g_i, d_{g_i})+ \frac{1}{2} \mathrm{N_{FP}}+ \frac{1}{2} \mathrm{N_{FN}} }{\mathrm{N_{TP}} + \frac{1}{2} \mathrm{N_{FP}}+ \frac{1}{2} \mathrm{N_{FN}}} 
\end{align}
\begin{align}
    &= 1 -  \frac{ \sum \limits_{i=1}^{\mathrm{N_{TP}}} \left(1  - \mathrm{lq}(g_i, d_{g_i}) \right) + \frac{1}{2} \mathrm{N_{FP}}+ \frac{1}{2} \mathrm{N_{FN}} }{\mathrm{N_{TP}} + \frac{1}{2} \mathrm{N_{FP}}+ \frac{1}{2} \mathrm{N_{FN}}} \\
    \label{eq:PQfinal}
    &=1 -  \frac{1}{\hat{Z}} \left( \sum \limits_{i=1}^{\mathrm{N_{TP}}}  \frac{1-\mathrm{lq}(g_i, d_{g_i})}{1-0.50}+\mathrm{N_{FP}} +\mathrm{N_{FN}} \right),
\end{align}
where $\hat{Z} = {\textcolor{red}{2} \mathrm{N_{TP}}+\mathrm{N_{FP}} +\mathrm{N_{FN}}}$. As a result, we can rewrite \equref{\ref{eq:PQfinal}} to express the PQ Error (i.e. 1-PQ) as follows:
\begin{align}
    1-\mathrm{PQ}= \frac{1}{\hat{Z}} \left( \sum \limits_{i=1}^{\mathrm{N_{TP}}}  \frac{1-\mathrm{lq}(g_i, d_{g_i})}{1-0.50}+\mathrm{N_{FP}} +\mathrm{N_{FN}} \right)
\end{align}
\renewcommand{\thesection}{G}
\section{More Experiments}
\label{subsec:ThresholdingExperiments2}
In addition to the experiments in our paper, here we demonstrate tuning hyper-parameters with oLRP Error, discuss how manually manipulating sources of errors (e.g. by setting $\mathrm{N_{FP}}=0$) affects LRP Error on an example visual detector, provide a use-case of LRP-Optimal thresholds, analyse the latency of LRP computation and the effect of TP validation threshold, $\tau$, on LRP Error, and present more s-LRP curves.

\subsection{Tuning Hyperparameters using LRP/oLRP Errors}
To demonstrate that LRP/oLRP Errors can be used to tune the hyperparameters, we tune the balancing parameter ($t$ below) of Repeat Factor Sampling (RFS) on LVIS and IoU threshold of Non Maximum Suppression (NMS) on COCO.

\textbf{Tuning RFS.} RFS repeats an image $i$ for $r_i$ times such that $r_i$ is the maximum of the repeat factors over classes, computed as $r_c = \max(1, \sqrt{t/f_c})$ where $f_c$ is the ratio of images in the training split including an example from class $c$ and $t$ is a hyperparameter. With such a sampling, RFS aims to promote the rare classes. Table \ref{tbl:repeatfactor} demonstrates that oLRP Error can also be used to tune the hyperparameter $t$ of RFS: (i) Without RFS ($t=0$), $\mathrm{AP_{r}}=0$ $\mathrm{oLRP_{r}}=100 \%$ (or just $1$), and (ii) for both AP and oLRP Error, $t = 0.01 $ has the best performance (underlined).

\renewcommand{\arraystretch}{0.6}
\begin{table}
\setlength{\tabcolsep}{0.2em}
\small
\caption{Tuning $t$ of RFS by AP and oLRP Error on Mask R-CNN+R50 using LVIS v1.0 dataset. \label{tbl:repeatfactor}} 
 \centering
\begin{tabular}{|c||c|c|c|c||c|c|c|c|}
\hline
&\multicolumn{4}{|c||}{AP $\uparrow$}&\multicolumn{4}{|c|}{oLRP Error $\downarrow$}\\\cline{2-9}
$t$&AP&$\mathrm{AP_{r}}$&$\mathrm{AP_{c}}$&$\mathrm{AP_{f}}$&oLRP&$\mathrm{oLRP_{r}}$& $\mathrm{oLRP_{c}}$& $\mathrm{oLRP_{f}}$\\
\hhline{=========}
$0$&$15.8$&$0.0$&$11.7$&$27.2$&$86.2$&$100.0$&$89.2$&$76.7$ \\ \hline
$0.0001$&$16.5$&$2.3$&$12.4$&$27.2$&$85.6$&$97.8$&$88.6$&$76.8$ \\ \hline
$0.0010$&$21.7$&$9.6$&$21.0$&$27.8$&$80.7$&$91.0$&$80.7$&$76.2$ \\
\hline
\underline{$0.0100$}&\underline{$23.9$}&\underline{$14.8$}&\underline{$23.2$}&\underline{$28.6$}&\underline{$78.8$}&\underline{$86.2$}&\underline{$78.9$}&\underline{$75.4$} \\
\hline
$0.1000$&$20.5$&$8.4$&$19.3$&$27.1$&$81.5$&$92.2$&$82.0$&$76.4$ \\
\hline
\end{tabular}
\end{table}

\textbf{Tuning NMS.} State-of-the-art methods commonly employ NMS as a postprocessing method in which if two detection boxes overlap more than a tuned IoU threshold, the one with the lower score is eliminated. Accordingly, in the extreme cases, $\mathrm{IoU} = 0.00$ implies the smallest detection set with no overlapping detections and $\mathrm{IoU} = 1.00$ corresponds to the largest detection set, practically without NMS. Similar to AP, oLRP Error can be used to tune IoU threshold of NMS, set to $0.50$ as default (Table \ref{tbl:NMSeffect}). 

\textbf{NMS-free methods.} As a different set of methods, recently proposed attention-based visual detectors  do not require NMS. Table \ref{tbl:transformer} presents oLRP Error can also properly rank these NMS-free visual detectors, i.e. DETR \cite{detr}, Sparse R-CNN \cite{sparsercnn} and Deformable DETR \cite{ddetr} with the same R50 backbone. 

\begin{table}
\setlength{\tabcolsep}{0.25em}
\small
\caption{Tuning IoU threshold of NMS on Faster R-CNN+R50. For both AP and oLRP Error, IoU=0.50 is chosen as the best IoU threshold (underlined). \label{tbl:NMSeffect}} 
 \centering
\begin{tabular}{|c||c|c|c||c|c|c|c|}
\hline
&\multicolumn{3}{|c||}{AP $\uparrow$}&\multicolumn{4}{|c|}{oLRP Error $\downarrow$}\\\cline{2-8}
IoU&AP&$\mathrm{AP_{50}}$&$\mathrm{AP_{75}}$&oLRP&$\mathrm{oLRP_{Loc}}$& $\mathrm{oLRP_{FP}}$& $\mathrm{oLRP_{FN}}$\\
\hhline{========}
0.00&$33.3$&$50.8$&$36.6$&$71.5$&$17.2$&$26.0$&$51.4$ \\ \hline
0.25&$36.5$&$56.9$&$39.8$&$69.4$&$17.5$&$27.5$&$46.3$ \\ \hline
\underline{0.50}&\underline{$37.8$}&\underline{$58.6$}&\underline{$41.0$}&\underline{$69.1$}&\underline{$17.5$}&\underline{$27.5$}&\underline{$45.4$} \\ \hline
0.75&$36.0$&$53.3$&$40.7$&$72.5$&$17.3$&$31.7$&$51.6$ \\ \hline
0.90&$27.6$&$38.2$&$31.7$&$79.8$&$16.1$&$47.8$&$61.7$ \\ \hline
1.00&$11.2$&$14.8$&$12.7$&$90.5$&$14.3$&$76.2$&$75.9$ \\ \hline
\end{tabular}
\end{table}

\begin{table}
\setlength{\tabcolsep}{0.05em}
\small
\caption{Evaluating NMS-free methods. \label{tbl:transformer}} 
 \centering
\begin{tabular}{|c||c|c|c||c|c|c|c|}
\hline
&\multicolumn{3}{|c||}{AP $\uparrow$}&\multicolumn{4}{|c|}{oLRP  $\downarrow$}\\\cline{2-8}
Method&AP&$\mathrm{AP_{50}}$&$\mathrm{AP_{75}}$&oLRP&$\mathrm{oLRP_{Loc}}$& $\mathrm{oLRP_{FP}}$& $\mathrm{oLRP_{FN}}$\\
\hhline{========}
DETR &$40.1$&$60.6$&$42.0$&$66.8$&$17.1$&$23.5$&$43.9$ \\ \hline
Sp. R-CNN &$45.0$&$64.1$&$48.9$&$63.6$&$14.6$&$24.1$&$41.6$ \\ \hline
DDETR &$46.8$&$66.3$&$50.7$&$62.1$&$14.3$&$24.5$&$39.0$ \\ \hline
\end{tabular}
\end{table}

\subsection{Effect of performance aspects on LRP Error} 
In order to provide more insight on LRP Error, this section investigates the effect of the performance aspects (i.e. localisation error, false positive rate and false negative rate) on LRP Error using Panoptic FPN. In Table \ref{tbl:performanceaspect}, we gradually remove the errors originating from each performance aspect and then present the resulting error. Note that (i) the localisation errors contribute to LRP Error after a normalisation by $1-\tau$ (\equref{3}) and $\tau=0.50$ in this case, and (ii) Number of TPs, FPs and FNs will also have an effect on the error besides the component values (Eq. 3). Still, we can deduce that (i) removing the errors originating from a performance measure decreases LRP Error and it decreases the most when the errors of FN rate is removed since $\mathrm{LRP_{FN}}$ is the largest error, (ii) Removing two components decreases LRP Error more than removing one component, and (iii) Once all errors are removed, as expected, LRP Error decreases to $0$, i.e. the detection and the ground truth sets are equal.

\begin{table}
\setlength{\tabcolsep}{0.25em}
\small
\caption{Effect of performance aspects on LRP Error on COCO dataset (with COCO-stuff). We use Panoptic FPN+R50 trained for 12 epochs. \checkmark: We keep the performance aspect as it is, \xmark: We remove the errors originating from the performance aspect. \label{tbl:performanceaspect}} 
 \centering
\begin{tabular}{|c|c|c|c|c|c|c|}
\hline
\multicolumn{3}{|c|}{Performance Aspects}&\multirow{2}{*}{LRP}&\multirow{2}{*}{$\mathrm{LRP_{Loc}}$}&\multirow{2}{*}{$\mathrm{LRP_{FP}}$}&\multirow{2}{*}{$\mathrm{LRP_{FN}}$}\\\cline{1-3}
Loc. Error & FP rate & FN rate & & & & \\ \hline
\checkmark&\checkmark&\checkmark&$77.5$&$21.0$&$39.3$&$57.2$ \\ \hline
\xmark&\checkmark&\checkmark&$64.9$&$0.0$&$39.3$&$57.2$ \\ \hline
\checkmark&\xmark&\checkmark&$72.8$&$21.0$&$0.0$&$57.2$ \\ \hline
\checkmark&\checkmark&\xmark&$62.4$&$21.0$&$39.3$&$0.0$ \\ \hhline{=======}
\checkmark&\xmark&\xmark&$42.0$&$21.0$&$0.0$&$0.0$ \\ \hline
\xmark&\checkmark&\xmark&$39.3$&$0.0$&$39.3$&$0.0$ \\ \hline
\xmark&\xmark&\checkmark&$57.2$&$0.0$&$0.0$&$57.2$ \\ \hhline{=======}
\xmark&\xmark&\xmark&$0.0$&$0.0$&$0.0$&$0.0$ \\ \hline
\end{tabular}
\end{table}

\subsection{A Use-Case of LRP-Optimal Thresholds in Video Object Detection}
\label{subsubsec:thresholdexperiment}

\textbf{Related Work on Setting the thresholds of the classifiers.} Employing visual detectors for a practical application requires a confidence threshold that balances precision, recall and localisation performance since, otherwise, the resulting output would be dominated by several false positives with low confidence scores. However, this topic has not received much attention from the research community and is usually handled by practitioners in a problem or deployment-specific manner. Conventionally, the thresholds of classifiers are set by finding the optimal F-measure on the PR curve or G-mean on the receiver operating characteristics curve, which do not consider localisation quality. Prior work on setting the classifier thresholds focused on the probabilistic models \cite{f1optimizer,maximizeF1Lipton}. Parambath et al. \cite{MaximizeF1NIPS} present a theoretical analysis of the F-measure, and propose a practical algorithm discretizing the confidence scores  in order to search for the optimal F-measure. Currently, for object detection, a general single threshold is used for all classes instead of class-specific thresholds \cite{AssociationLSTM}.

\textbf{The Experimental Setup.} Here, we demonstrate a use-case where oLRP Error helps us to set class-specific optimal thresholds as an alternative to the naive approach of using a general, class-independent threshold. To this end, we develop a simple, online video object detection framework where we use an off-the-shelf still-image object detector (RetinaNet-50 \cite{FocalLoss} trained on COCO \cite{COCO}) and built three different versions of the video object detector. The first version, denoted with $B$, uses the still-image object detector to process each frame of the video independently. The second and third versions, denoted with $G$ and $S$, respectively, again use the still-image object detector to process each frame and in addition, they link bounding boxes across subsequent frames using the Hungarian matching algorithm \cite{Hungarian} and update the scores of these linked boxes using a simple Bayesian rule (details of this simple online video object detector is given below). For the sake of efficiency in linking\footnote{Note that Hungarian matching has a time complexity of $\mathcal{O}(N^3)$ and RetinaNet can output thousands of detections without any thresholding.}, both of these detectors discard the low-precision detections. Accordingly, the only difference between $G$ and $S$ is that while $G$ uses a validated threshold of $0.50$ (see Fig. 11(a) in the paper to notice that the LRP-Optimal threshold distribution of RetinaNet has a mean around $0.50$) as the confidence score threshold for all classes, $S$ uses LRP-Optimal threshold per class. We test these three detectors on 346 videos of ImageNet VID validation set \cite{ILSVRC} for 15 object classes which also happen to be included in COCO.

\textbf{Details of the Online Video Object Detectors.} There are two online video object detectors: $G$ and $S$ which respectively use the general, class-independent thresholding approach with $0.50$ as threshold and the class-specific thresholds. For each of the online detectors, at each time interval, the detections from the previous and current frames are associated using the Hungarian algorithm \cite{Hungarian} considering a box linking function and the confidence scores of associated BBs of the current frame are updated using the score distributions from both frames. Since an online tracker, specifically \cite{CFCF}, is also used in our method, we use the L1 norm of the difference of confidence score distributions of neighbouring frames and the IoU overlap of the tracker prediction and the detection at current frame. While choosing this box linking function, we inspired from the tube linking score of \cite{DetectToTrack}. The updated score is estimated using the Bayes Theorem such that the prior is the updated tubelet score in the previous frame and likelihood is the currently associated high confidence detection with that tubelet. In such an update method, even though the updated scores converge to $1$ quickly, which is harmful for lower recall, precision improves in larger recall portions. Also, we call a BB as ``dominant object’’ if its updated score increases by $0.20$. In order to increase the recall, the disappearance of a ``dominant object’’ is closely inspected by using the tracker again to predict the possible location, then the cropped region is classified by class-wise binary classifiers (object vs. background).

\renewcommand{\arraystretch}{0.6}
\begin{table*}
\setlength{\tabcolsep}{0.4em}
\caption{Comparison among $B$, $G$, $S$ with respect to AP \& oLRP Error and their best class-specific configurations. The mean of class thresholds are assigned as N/A since the thresholds are set class-specific and the mean is not used. $s^{*}$ denotes the LRP-Optimal thresholds. Note that unlike AP, lower scores are better for LRP Error.} 
 \centering
 \footnotesize 
 \begin{tabular}{|c|l|r|r|r|r|r|r|r|r|r|r|r|r|r|r|r|r|}
 \hline
 & \multicolumn{1}{c|}{\rotatebox[origin=c]{90}{Method}} & \multicolumn{1}{c|}{\rotatebox[origin=c]{90}{airplane}}  & \multicolumn{1}{c|}{\rotatebox[origin=c]{90}{bicycle}}  & \multicolumn{1}{c|}{\rotatebox[origin=c]{90}{bird}}  & \multicolumn{1}{c|}{\rotatebox[origin=c]{90}{bus}} & \multicolumn{1}{c|}{\rotatebox[origin=c]{90}{car}}  & \multicolumn{1}{c|}{\rotatebox[origin=c]{90}{cow}}  & \multicolumn{1}{c|}{\rotatebox[origin=c]{90}{dog}}  & \multicolumn{1}{c|}{\rotatebox[origin=c]{90}{cat}} & \multicolumn{1}{c|}{\rotatebox[origin=c]{90}{elephant}}  & \multicolumn{1}{c|}{\rotatebox[origin=c]{90}{horse}}  & \multicolumn{1}{c|}{\rotatebox[origin=c]{90}{motorcycle}} & \multicolumn{1}{c|}{\rotatebox[origin=c]{90}{sheep}} & \multicolumn{1}{c|}{\rotatebox[origin=c]{90}{train}} & \multicolumn{1}{c|}{\rotatebox[origin=c]{90}{boat}} & \multicolumn{1}{c|}{\rotatebox[origin=c]{90}{zebra}} & \multicolumn{1}{c|}{\rotatebox[origin=c]{90}{mean}}\\
 \hline
 \parbox[t]{2mm}{\multirow{3}{*}{\rotatebox[origin=c]{90}{$\mathrm{AP_{50}}$}}}
 &B&$\mathbf{68.1}$&$\mathbf{63.0}$&$\mathbf{54.7}$&$\mathbf{56.5}$&$\mathbf{55.5}$&$\mathbf{58.7}$&$\mathbf{46.3}$&$\mathbf{60.1}$&$\mathbf{66.1}$&$\mathbf{47.3}$&$\mathbf{60.2}$&$\mathbf{56.1}$&$\mathbf{71.3}$&$\mathbf{82.9}$&$\mathbf{81.6}$&$\mathbf{61.9}$\\
 &G&$62.1$&$44.5$&$49.2$&$39.8$&$41.7$&$51.0$&$41.6$&$56.8$&$58.8$&$44.1$&$57.1$&$54.7$&$60.0$&$76.9$&$76.5$&$54.4$\\
 &S&$64.5$&$53.5$&$50.0$&$48.5$&$41.9$&$49.2$&$43.4$&$56.9$&$58.9$&$44.4$&$57.3$&$54.5$&$60.9$&$79.2$&$78.2$&$56.1$\\
  \hline  
  \parbox[t]{2mm}{\multirow{3}{*}{\rotatebox[origin=c]{90}
 {oLRP}}}&B&$62.7$&$77.6$&$71.8$&$70.2$&$75.9$&$69.2$&$72.8$&$70.0$&$62.5$&$72.3$&$69.2$&$67.7$&$\mathbf{58.3}$&$59.4$&$43.6$&$66.9$\\
 &G&$60.6$&$78.3$&$69.1$&$72.7$&$\mathbf{75.8}$&$67.9$&$71.4$&$\mathbf{69.7}$&$61.4$&$\mathbf{69.9}$&$\mathbf{65.4}$&$\mathbf{64.8}$&$58.6$&$55.3$&$43.2$&$65.6$\\
 &S&$\mathbf{60.3}$&$\mathbf{76.2}$&$\mathbf{68.7}$&$\mathbf{68.8}$&$75.9$&$\mathbf{67.8}$&$\mathbf{71.2}$&$\mathbf{69.7}$&$\mathbf{61.3}$&$70.1$&$65.5$&$64.9$&$\mathbf{58.3}$&$\mathbf{55.1}$&$\mathbf{42.5}$&$\mathbf{65.1}$\\
 \hline
 \parbox[t]{2mm}{\multirow{3}{*}{\rotatebox[origin=c]{90}{$\mathrm{oLRP_{Loc}}$}}}&B&$18.2$&$27.1$&$16.9$&$17.7$&$20.7$&$14.5$&$16.6$&$20.3$&$17.0$&$15.5$&$19.2$&$15.4$&$15.9$&$19.9$&$12.8$&$17.9$\\
 &G&$18.1$&$25.8$&$17.0$&$16.0$&$20.7$&$15.1$&$16.5$&$20.0$&$17.0$&$16.0$&$19.5$&$15.5$&$15.6$&$19.5$&$12.8$&$17.7$\\
 &S&$18.6$&$27.0$&$17.0$&$17.3$&$20.7$&$14.8$&$17.0$&$20.0$&$17.0$&$16.0$&$19.4$&$15.5$&$15.9$&$19.7$&$13.1$&$17.9$\\
 \hline
  \parbox[t]{2mm}{\multirow{3}{*}{\rotatebox[origin=c]{90}{$\mathrm{oLRP_{FP}}$}}}&B&$8.0$&$22.8$&$30.0$&$20.3$&$30.3$&$22.4$&$24.2$&$24.8$&$9.5$&$24.6$&$15.8$&$14.1$&$9.9$&$16.3$&$3.4$&$18.4$\\
 &G&$8.6$&$11.6$&$17.4$&$13.7$&$31.1$&$21.8$&$22.9$&$27.9$&$7.1$&$22.1$&$4.9$&$7.8$&$9.1$&$7.7$&$1.6$&$14.2$\\
 &S&$8.7$&$22.6$&$18.4$&$19.3$&$32.0$&$18.2$&$26.9$&$28.3$&$7.5$&$23.1$&$8.4$&$7.8$&$11.0$&$8.9$&$3.0$&$16.3$\\
 \hline
  \parbox[t]{2mm}{\multirow{3}{*}{\rotatebox[origin=c]{90}{$\mathrm{oLRP_{FN}}$}}}&B&$38.3$&$42.7$&$47.8$&$47.7$&$49.9$&$50.4$&$53.3$&$39.4$&$39.5$&$54.0$&$44.8$&$49.4$&$34.4$&$22.4$&$22.0$&$42.4$\\
 &G&$35.9$&$52.3$&$48.0$&$57.1$&$49.3$&$47.3$&$51.2$&$37.2$&$38.8$&$49.4$&$41.5$&$46.7$&$36.0$&$22.1$&$22.7$&$42.4$\\
 &S&$32.6$&$38.9$&$48.9$&$46.1$&$48.8$&$49.0$&$48.0$&$36.9$&$38.5$&$49.3$&$40.6$&$46.8$&$33.9$&$20.3$&$20.2$&$39.8$\\
 \hline
   \parbox[t]{2mm}{\multirow{3}{*}{\rotatebox[origin=c]{90}{$s^{*}$}}}&B&$0.38$&$0.31$&$0.44$&$0.27$&$0.49$&$0.61$&$0.42$&$0.49$&$0.49$&$0.52$&$0.45$&$0.51$&$0.41$&$0.45$&$0.31$&N/A\\
 &G&$0.00$&$0.69$&$0.97$&$0.68$&$0.00$&$0.96$&$0.48$&$0.70$&$0.33$&$0.64$&$0.60$&$0.84$&$0.59$&$0.90$&$0.00$&N/A\\
 &S&$0.00$&$0.54$&$0.98$&$0.45$&$0.00$&$0.91$&$0.49$&$0.64$&$0.39$&$0.58$&$0.63$&$0.85$&$0.55$&$0.89$&$0.54$&N/A\\
 \hline 
 \end{tabular}
\label{table2} 
 \end{table*}

\begin{figure*}
\centering
\includegraphics[width=0.75\textwidth]{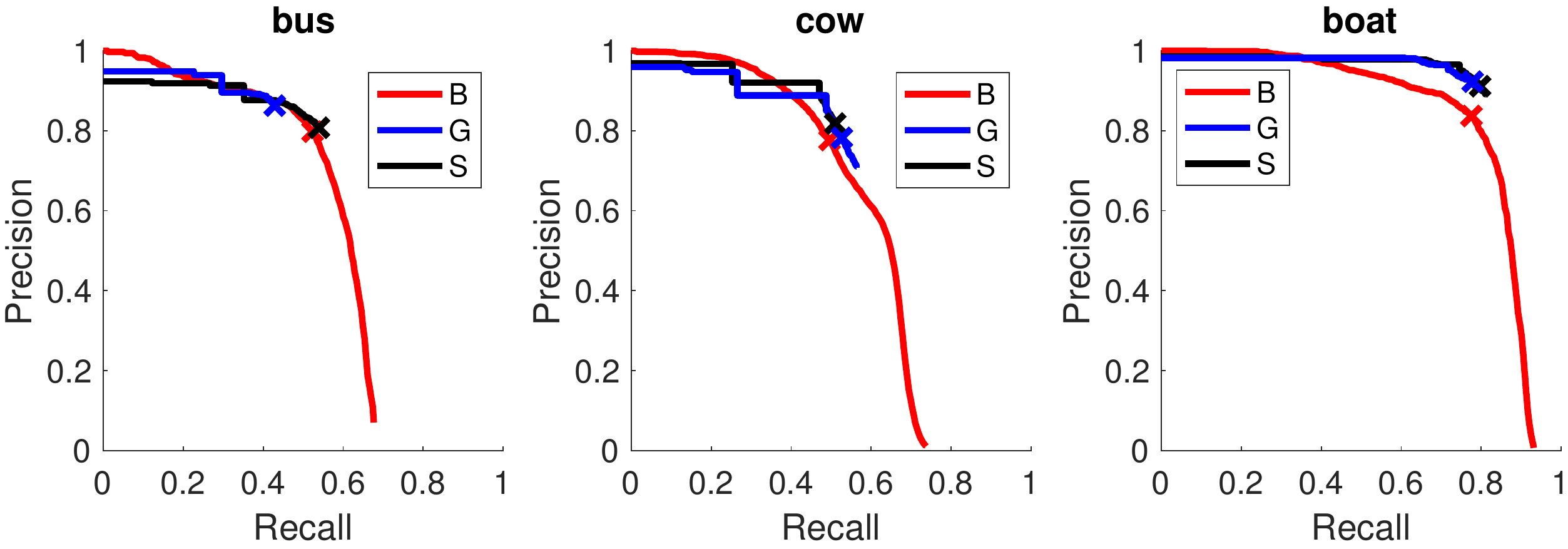}
\caption{Example PR curves of the methods on three example classes. Optimal PR pairs are marked with crosses. Using a LRP-Optimal threshold balances FP and FN errors resulting in a stretched PR curve either in recall (e.g.(a)) or precision (e.g.(b)) axis depending on the chosen general threshold. Furthermore, using AP for these pruned PR curves does not provide consistent performance evaluation.}
\label{fig:PRCurves}
\end{figure*}

\noindent \textbf{AP vs. oLRP Error for Video Object Detection:} We compare $G$ with $B$ in order to represent the evaluation perspectives of AP and oLRP Error -- see \figref{\ref{fig:PRCurves}} and Table \ref{table2}. Since $B$ is a conventional object detector, with conventional PR curves as illustrated in \figref{\ref{fig:PRCurves}}. On the other hand, in order to be faster, $G$ ignores some of the detections causing its maximum recall to be less than that of $B$. Thus, these shorter ranges in the recall set a big problem in the evaluation with respect to AP. Quantitatively, $B$ surpasses $G$ by $7.5\%$ AP. On the other hand, despite limited recall coverage, $G$ obtains higher precision than $B$ especially through the end of its PR curve. To illustrate, for the ``boat” class in \figref{\ref{fig:PRCurves}}, $G$ has significantly better precision after approximately between $0.5$ and $0.9$ recall even though its AP is lower by $6\%$. Since oLRP Error compares methods concerning their best configurations, this difference is clearly addressed comparing their oLRP Error in which $G$ surpasses $S$ by $4.1\%$. Furthermore, the superiority of $G$ is shown to be its higher precision since FN components of $G$ and $S$ are very close while FP component of $G$ is $8.6\%$ better, which is also the exact difference of precisions in their peaks of PR curves.

Therefore, while $G$ seems to have very low performance in terms of AP, for 12 classes $G$ reaches better peaks than $B$ as illustrated by the oLRP Errors in Table \ref{table2}. This suggests that oLRP Error is better than AP in capturing the performance details of this kind methods that uses thresholding.

\noindent \textbf{Effect of the Class-specific LRP-Optimal Thresholds:} Compared to $G$, owing to the class-specific thresholds, $S$ has $1.7\%$ better AP and $0.5\%$ better oLRP Error as shown in Table \ref{table2}. However, since the mean is dominated by $s^*$ around $0.50$, it is better to focus on classes with low or high $s^*$ values in order to grasp the effect of the approach. The ``bus” class has the lowest $s^*$ with $27\%$. For this class, $S$ surpasses $G$ by $8.7\%$ in AP and $3.9\%$ in oLRP Error. This performance increase is also observed for other classes with very low thresholds, such as ``airplane”, ``bicycle” and ``zebra”. For these classes with lower thresholds, the effect of LRP-Optimal threshold on the PR curve is to stretch the curve in the recall domain (maybe by accepting some loss in precision) as shown in the ``bus” example in \figref{\ref{fig:PRCurves}}. Not surprisingly, ``cow” is one of the two classes for which AP of $S$ is lower since its threshold is the highest and thereby causing recall to be more limited. On the other hand, regarding oLRP Error, the result is not worse since this time the PR curve is stretched through the positive precision, as shown in \figref{\ref{fig:PRCurves}}, allowing better FP errors. Thus, in any case, lower or higher, the LRP-Optimal threshold aims to discover the best PR curve. There are four classes in total for which $G$ is better than $S$ in terms of oLRP Error. However, note that the maximum difference is $0.2\%$ in oLRP Error and these are the classes with thresholds around $0.5$. These suggest that choosing class-specific thresholds rather than the general, class-independent thresholding approach increases the performance of the detector especially for classes with low or high class-specific thresholds.

\subsection{Analysing Latency of LRP Error Computation}
\label{subsec:Latency}
Since the PQ and LRP Error are very similar in formulation (Section 6 in the paper), it is obvious that they require very similar time to compute. Hence, in this section we focus on how much time computing LRP Error adds to the AP computation. Using COCO api \cite{COCO}, COCO evaluation follows a five-step algorithm to compute AP: (i) loading annotations into memory, (ii) loading and preparing results, (iii) per image evaluation, (iv) accumulating evaluation results, and (v) summarizing (i.e. printing) the results. Since step (i) and (ii) are independent of the performance measure, and (v) is a simple printing operation (i.e. it takes less than $3$ seconds compute (i), (ii) and (v) in total) and we do not change (iii) per image evaluation except returning the computed IoUs of TPs, we analysed the additional latency of computing oLRP Error for step (iv) accumulation, in which the per-image evaluation results are combined into performance values. In order to do that using a standard CPU, we computed and averaged the runtime of this step using all 36 SOTA models in Table 3 in the paper on COCO 2017 val with 5000 images. We observed that LRP Error computation (including LRP, oLRP, their components, class-specific LRP-Optimal thresholds for different ``size'' and ``maximum detection number'' criteria as done by COCO api - see \cite{COCO} for details) introduces a negligible overhead with around one second both for (iv) accumulate step (from $5.8$ seconds to $6.6$ seconds) and for entire computation (from $38.7$ to $39.6$ seconds).

\begin{figure*}
\centering
\includegraphics[width=1.0\textwidth]{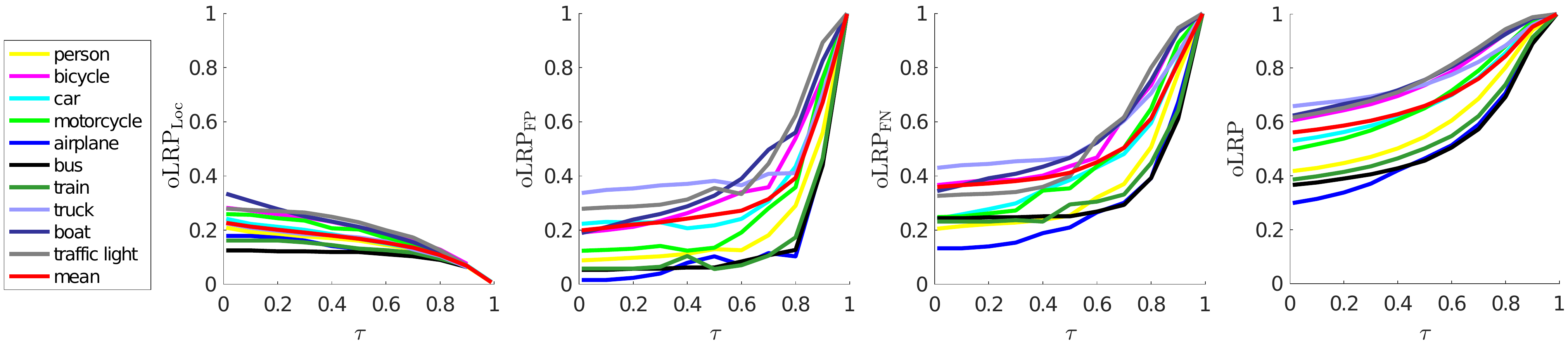}
\caption{For each class from COCO, oLRP Error and its components for Faster R-CNN (X101-12) are plotted against $\tau$. The mean represents the mean of 80 classes. }
\label{fig:tau}
\end{figure*}

\subsection{Analyzing the Effect of TP Validation Threshold} 
\label{subsec:tauParameter}
Finally, we analyse how LRP Error is affected from the TP validation threshold  parameter, $\tau$. We use Faster R-CNN (X101-12) (Table 3 in the paper) results of the first 10 classes and mean-error for clarity, the effect of the $\tau$ parameter is analysed in \figref{\ref{fig:tau}} 
%
%
on oLRP Error. As expected, larger $\tau$ values imply lower localisation error ($\mathrm{oLRP_{Loc}}$). On the other hand, a larger $\tau$ causes FP and FN components to increase rapidly, leading to higher total error (oLRP Error). This is intuitive since at the extreme case, i.e., when $\tau=1$, there are hardly any TP (i.e. all the detections are FPs), which makes oLRP Error to be $1$. Therefore, LRP Error allows measuring the performance of a detector designed for an application that requires a different $\tau$ by also providing additional information. In addition, investigating oLRP Error for different $\tau$ values represents a good extension for ablation studies. 

\subsection{More Examples for s-LRP Curves}
\textbf{COCO Dataset:} \figref{\ref{fig:LRPConfScoreCurves}} provides s-LRP curves of different object detectors for the ``zebra'' and ``bus'' classes whose PR curve is demonstrated in Fig. 7 of the paper but not included in Fig. 10 due to space limitations.

\begin{figure}
        \captionsetup[subfigure]{}
        \centering
        \begin{subfigure}[b]{0.24\textwidth}
        \includegraphics[width=\textwidth]{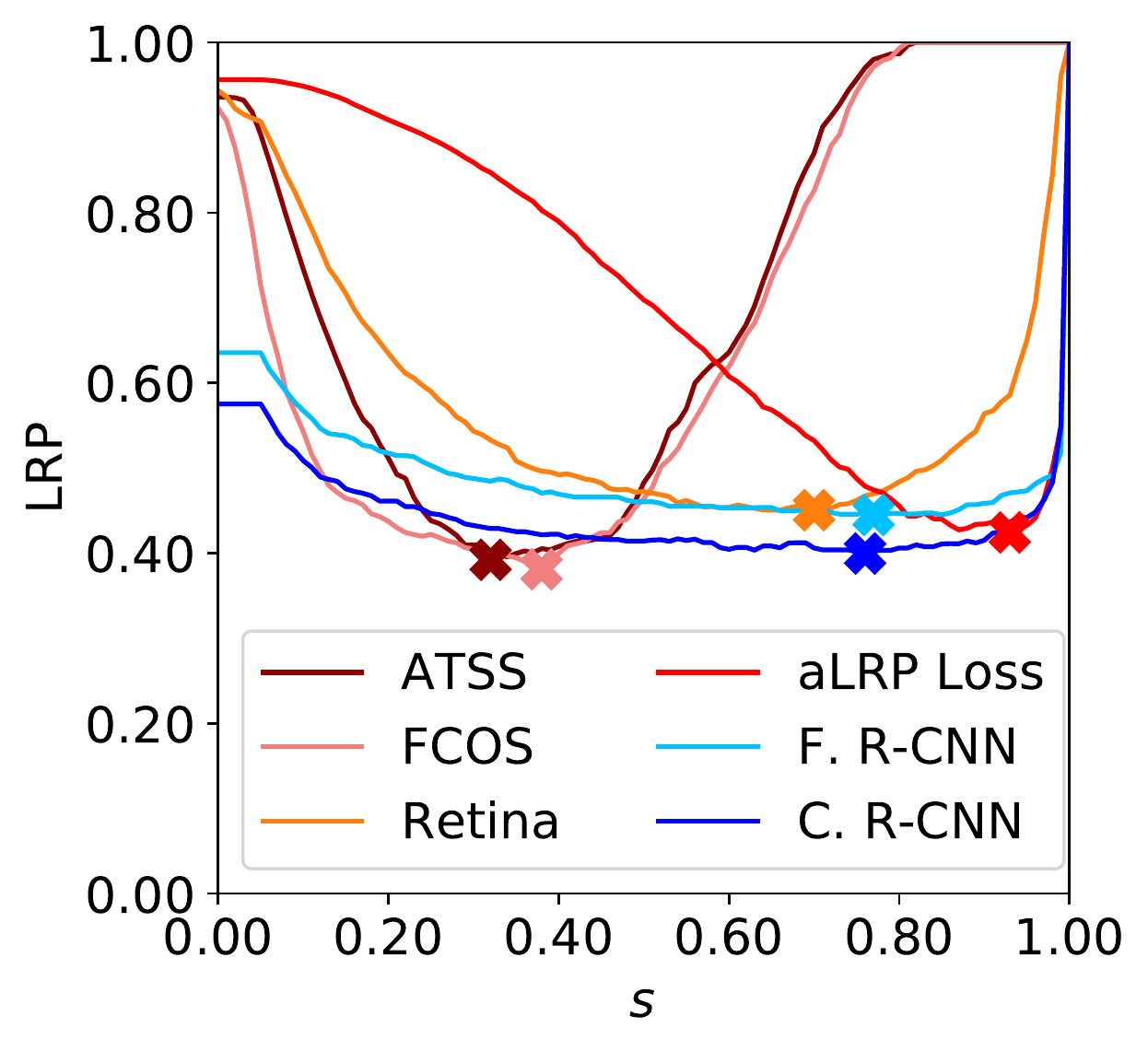}
        \caption{Zebra}
        \end{subfigure}
        \begin{subfigure}[b]{0.24\textwidth}
        \includegraphics[width=\textwidth]{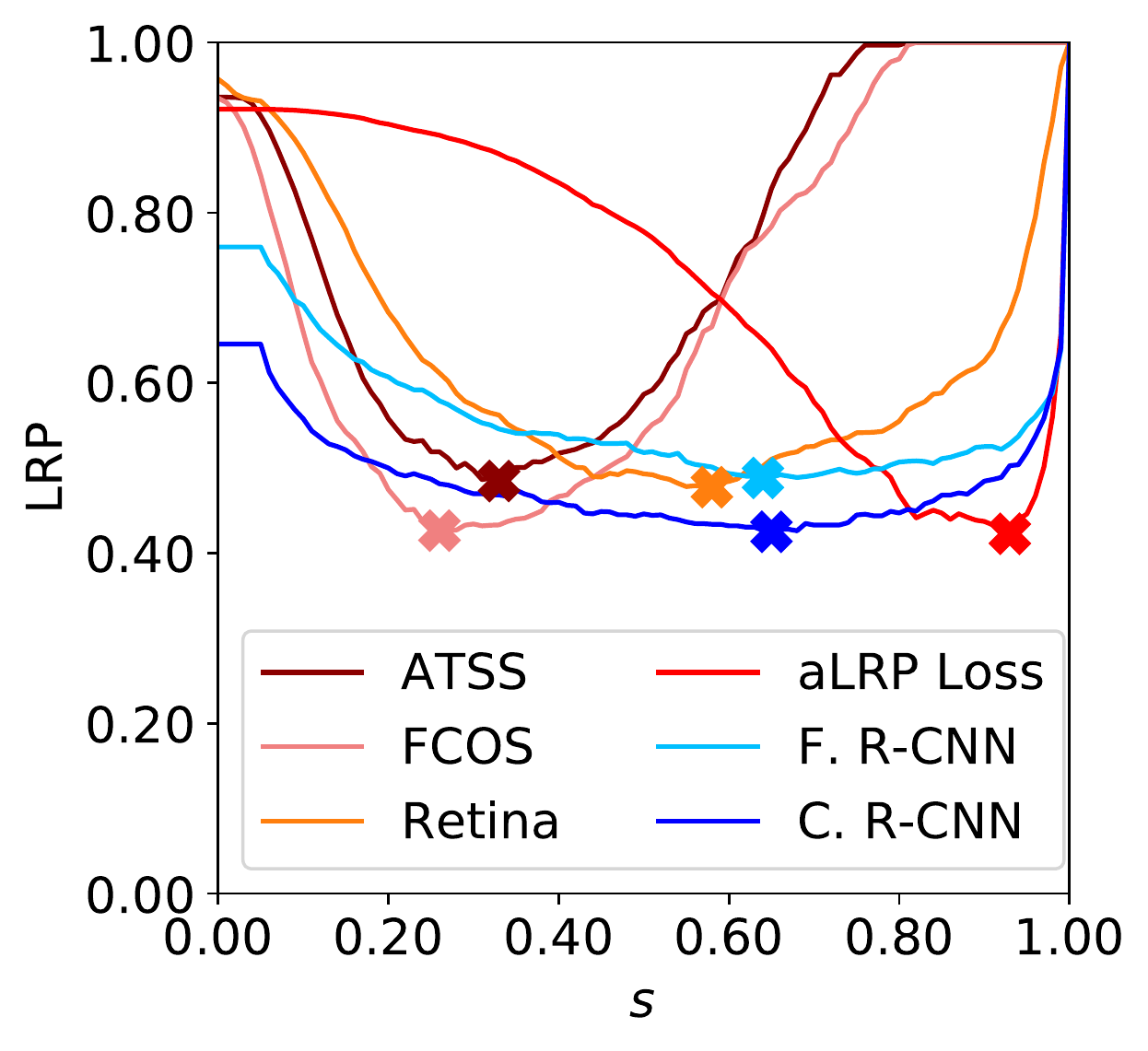}
        \caption{Bus}
        \end{subfigure}
        \caption{s-LRP curves of different object detectors for the ``zebra'' and ``bus'' classes from COCO dataset whose PR curve is demonstrated in Fig. 7 of the paper but not included in Fig. 10 due to space limitations. The lines with a different red tone represent a one-stage detector, while blue tones correspond to two-stage detector F. R-CNN: Faster R-CNN, C. R-CNN: Cascade R-CNN}
        \label{fig:LRPConfScoreCurves1}
\end{figure}

\begin{figure*}
        \captionsetup[subfigure]{}
        \centering
        \begin{subfigure}[b]{0.32\textwidth}
        \includegraphics[width=\textwidth]{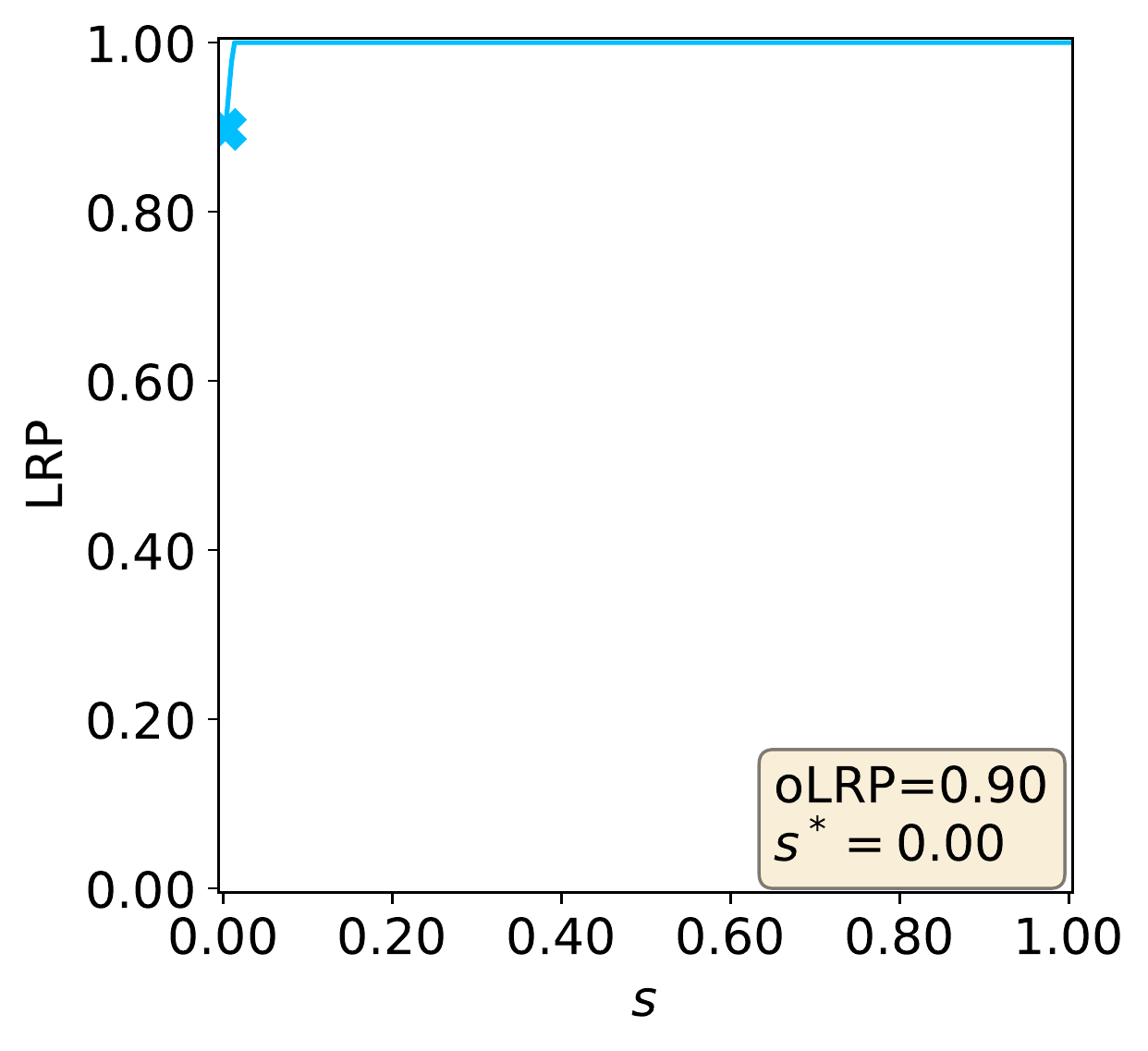}
        \caption{Walking stick}
        \end{subfigure}
        \begin{subfigure}[b]{0.32\textwidth}
        \includegraphics[width=\textwidth]{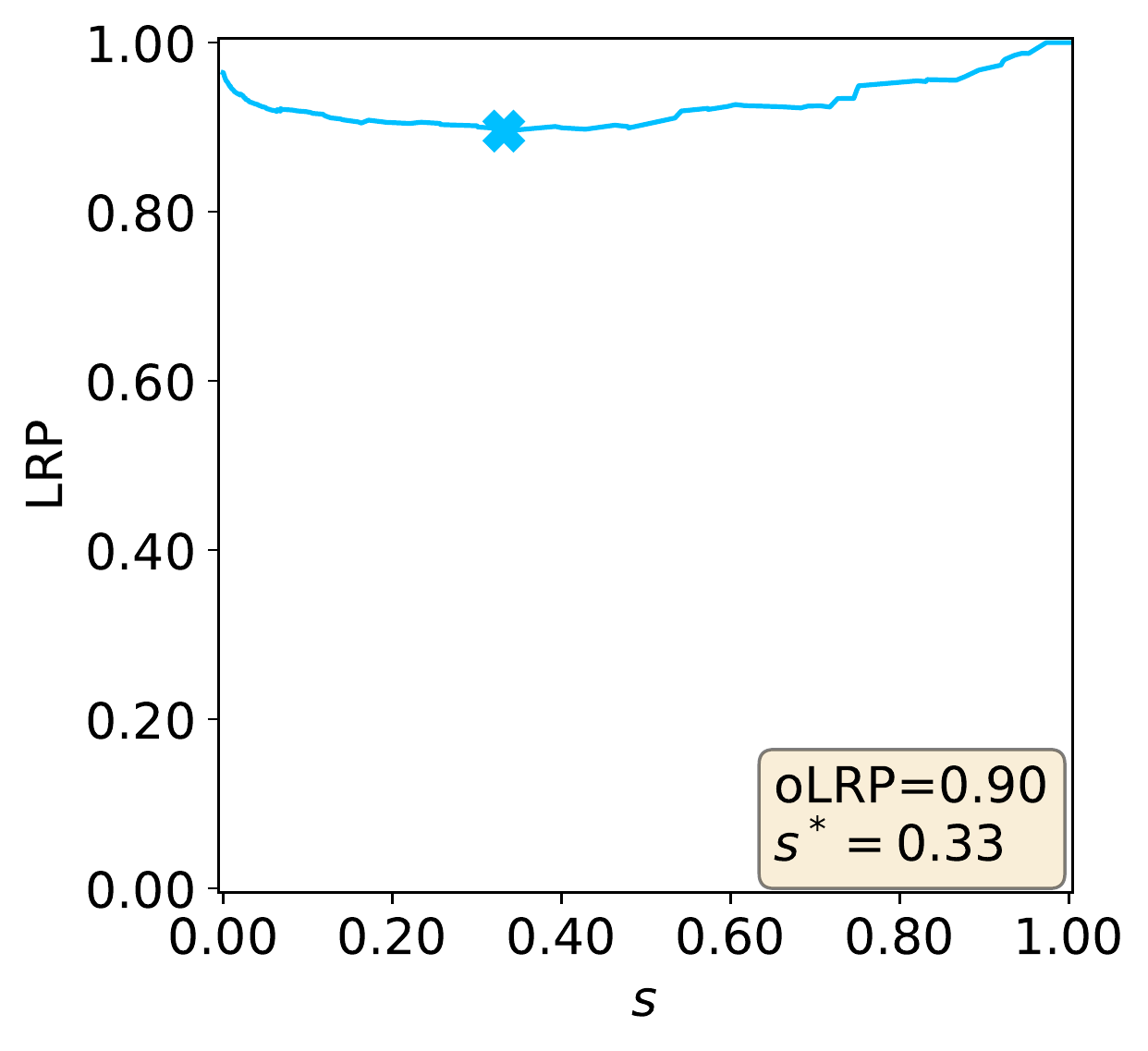}
        \caption{Weatherwane}
        \end{subfigure}
        \begin{subfigure}[b]{0.32\textwidth}
        \includegraphics[width=\textwidth]{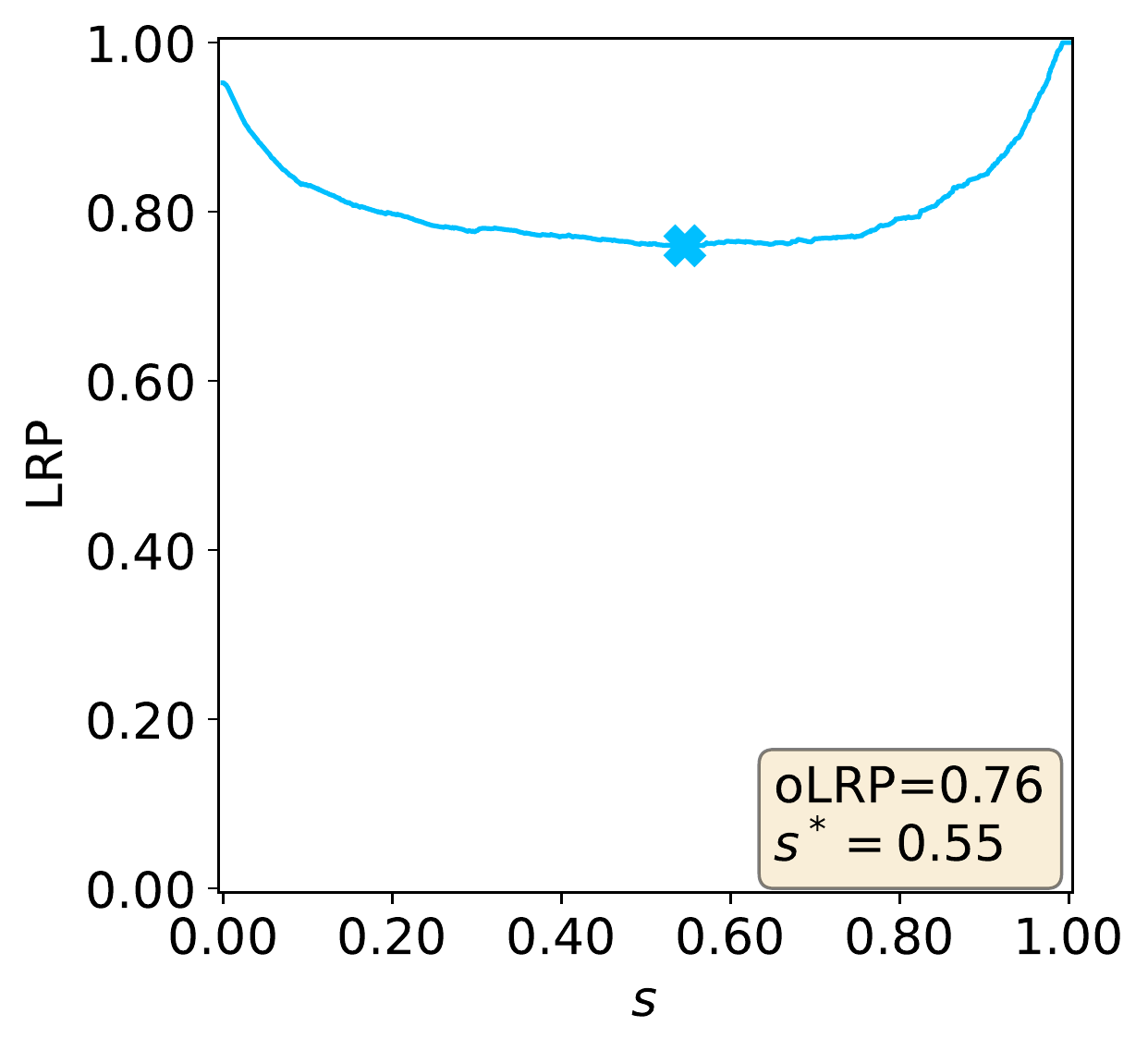}
        \caption{Baseball bat}
        \end{subfigure}
        
        \begin{subfigure}[b]{0.32\textwidth}
        \includegraphics[width=\textwidth]{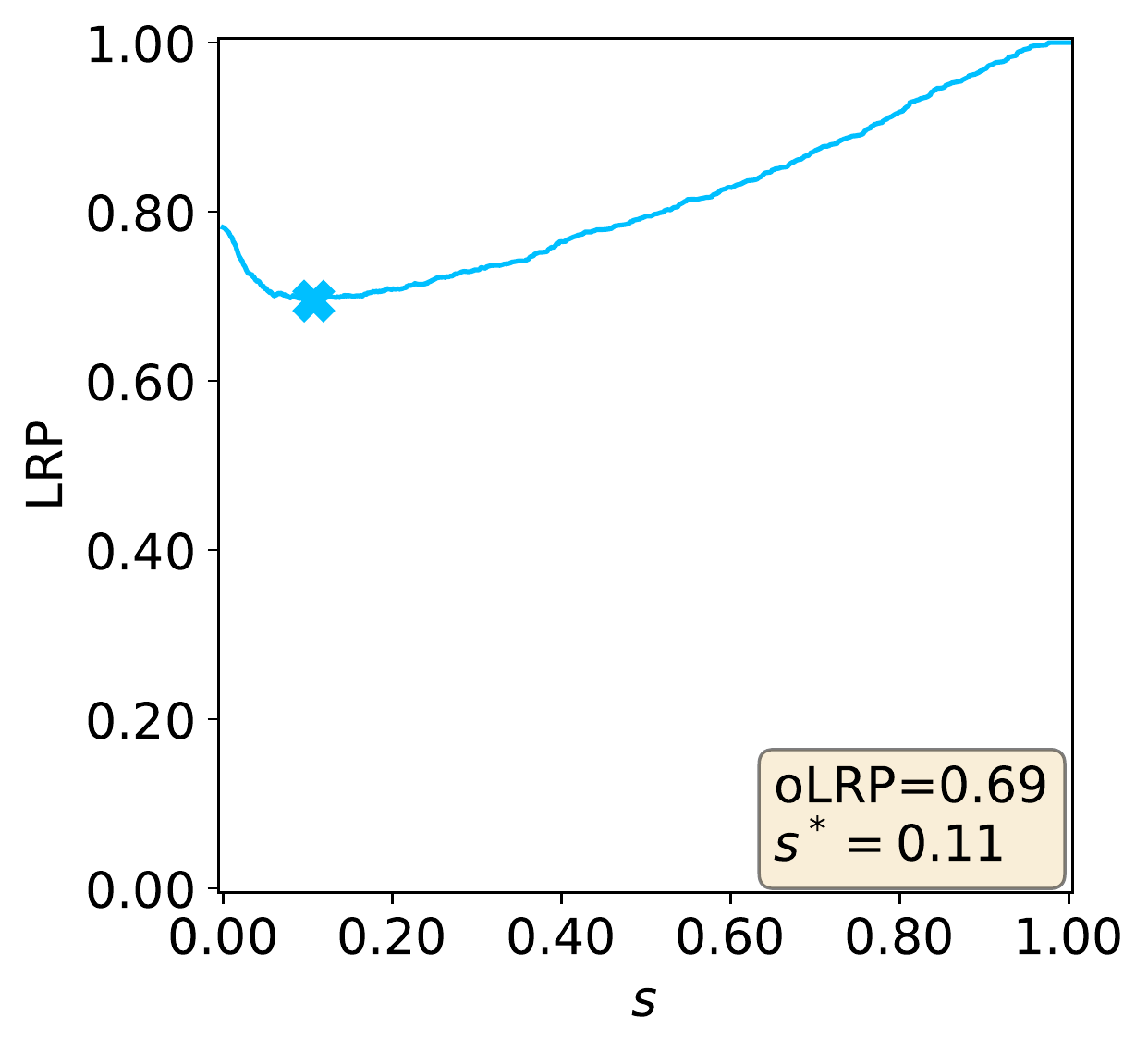}
        \caption{Bottle cap}
        \end{subfigure}
        \begin{subfigure}[b]{0.32\textwidth}
        \includegraphics[width=\textwidth]{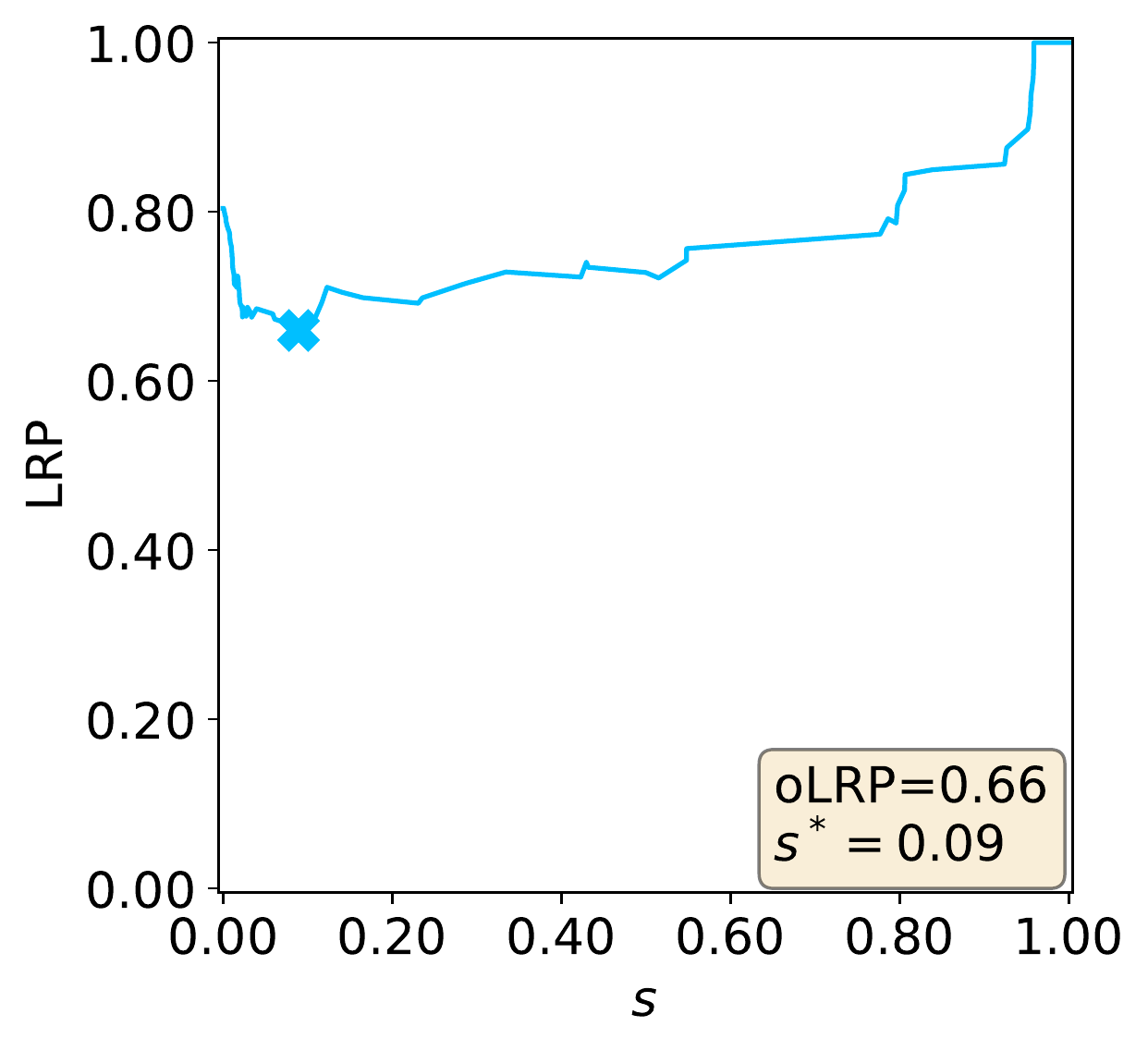}
        \caption{Radiator}
        \end{subfigure}
        \begin{subfigure}[b]{0.32\textwidth}
        \includegraphics[width=\textwidth]{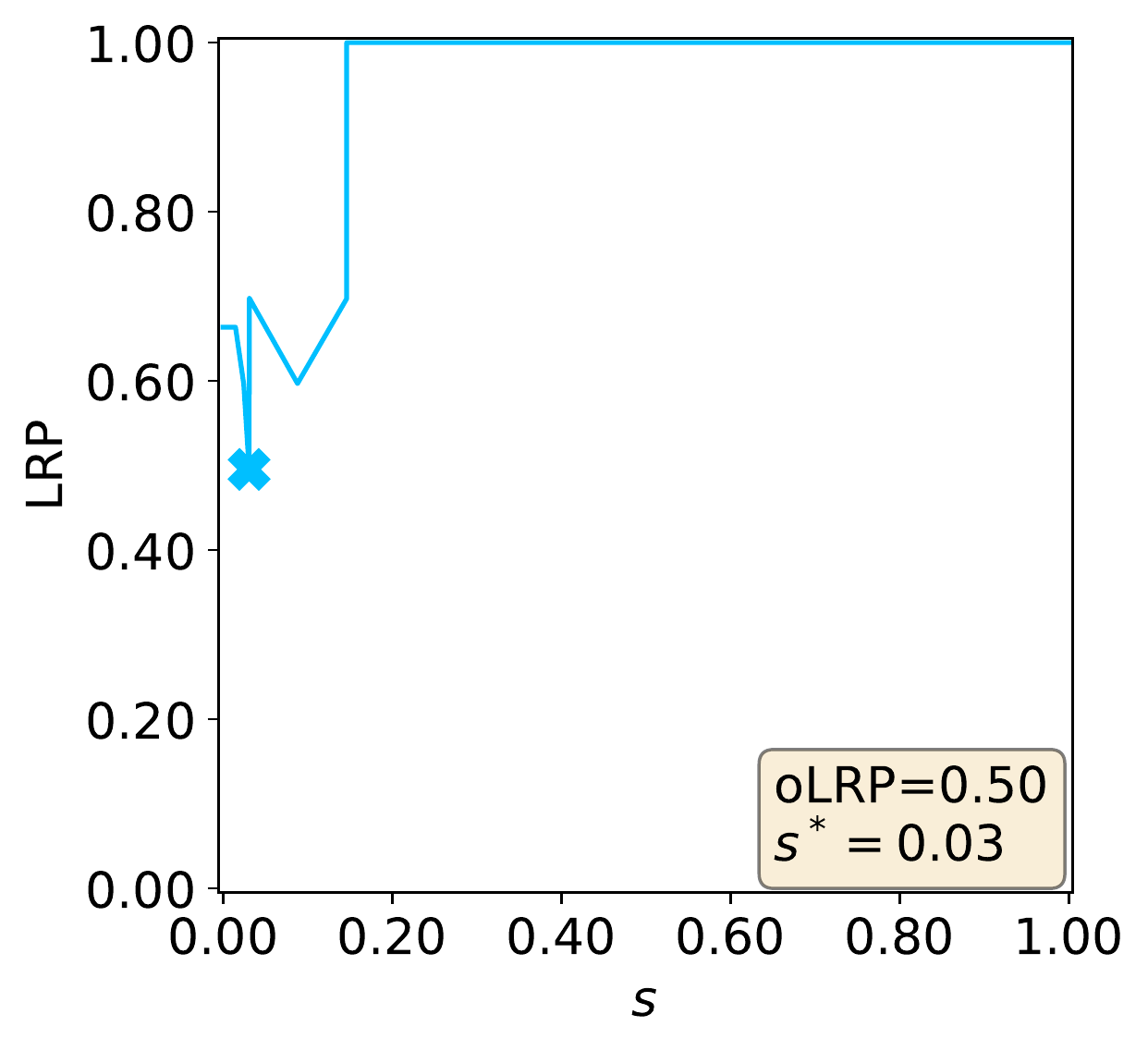}
        \caption{Sour cream}
        \end{subfigure}
        
        \begin{subfigure}[b]{0.32\textwidth}
        \includegraphics[width=\textwidth]{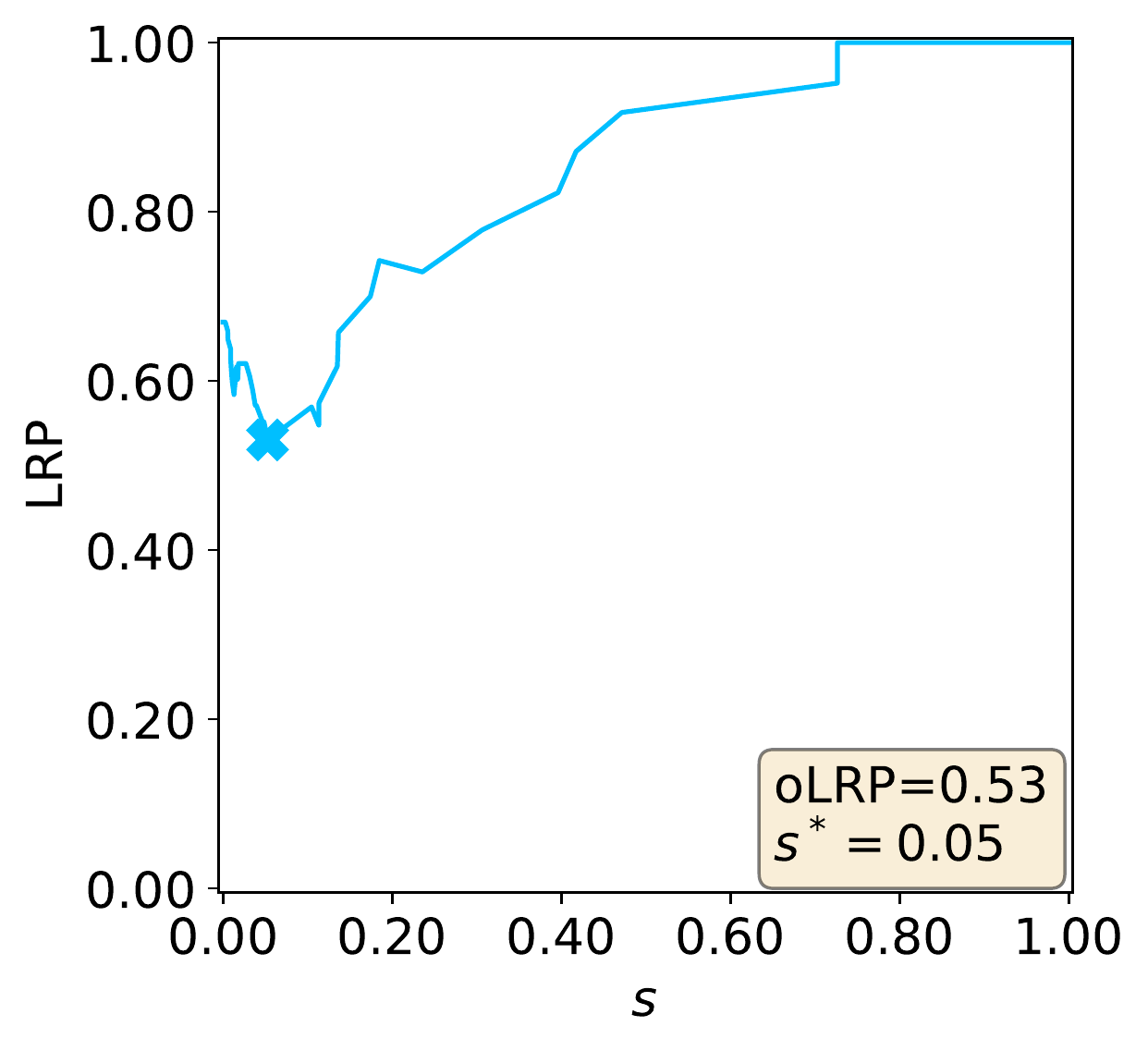}
        \caption{Cornice}
        \end{subfigure}
        \begin{subfigure}[b]{0.32\textwidth}
        \includegraphics[width=\textwidth]{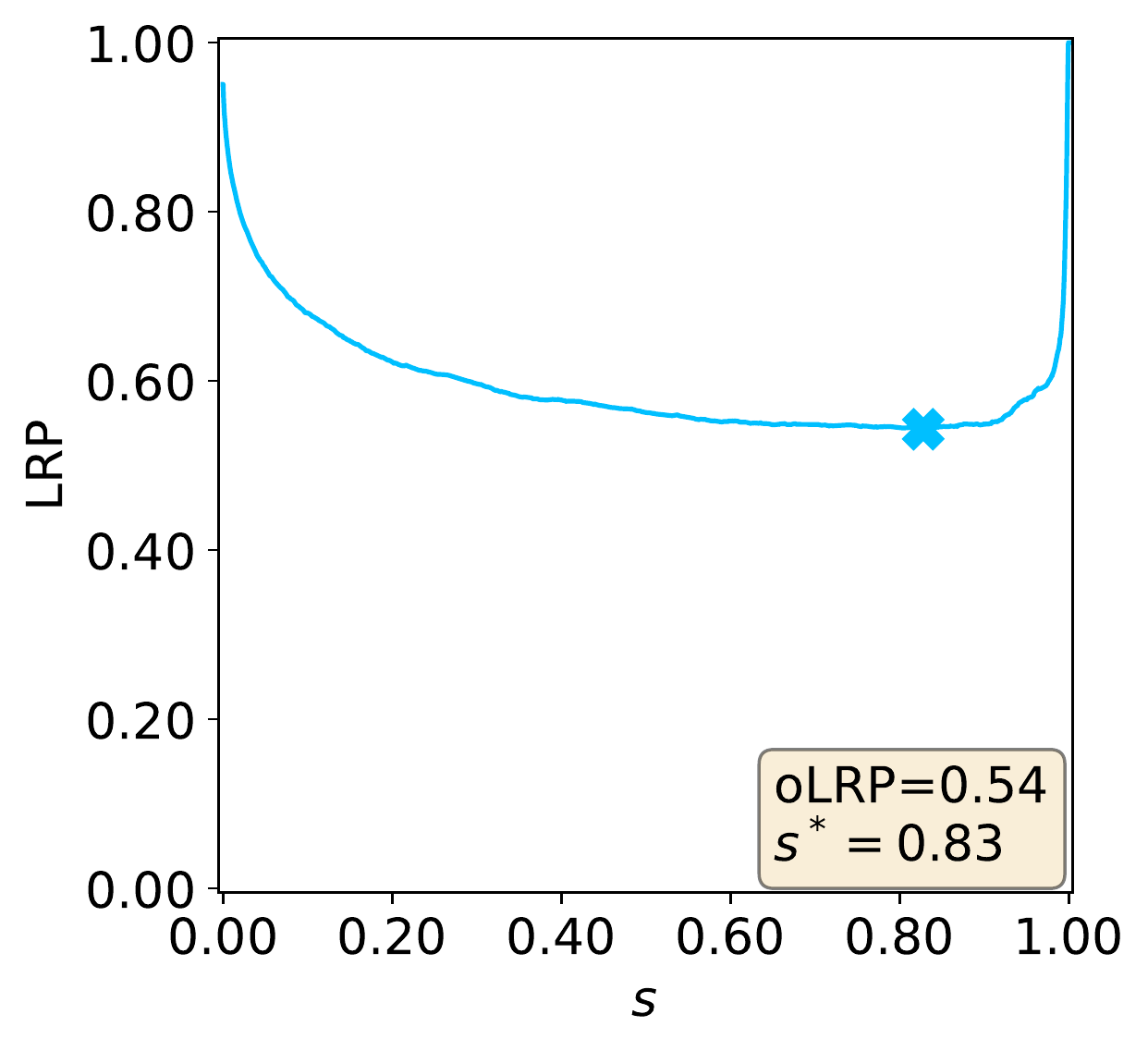}
        \caption{Zebra}
        \end{subfigure}
        \begin{subfigure}[b]{0.32\textwidth}
        \includegraphics[width=\textwidth]{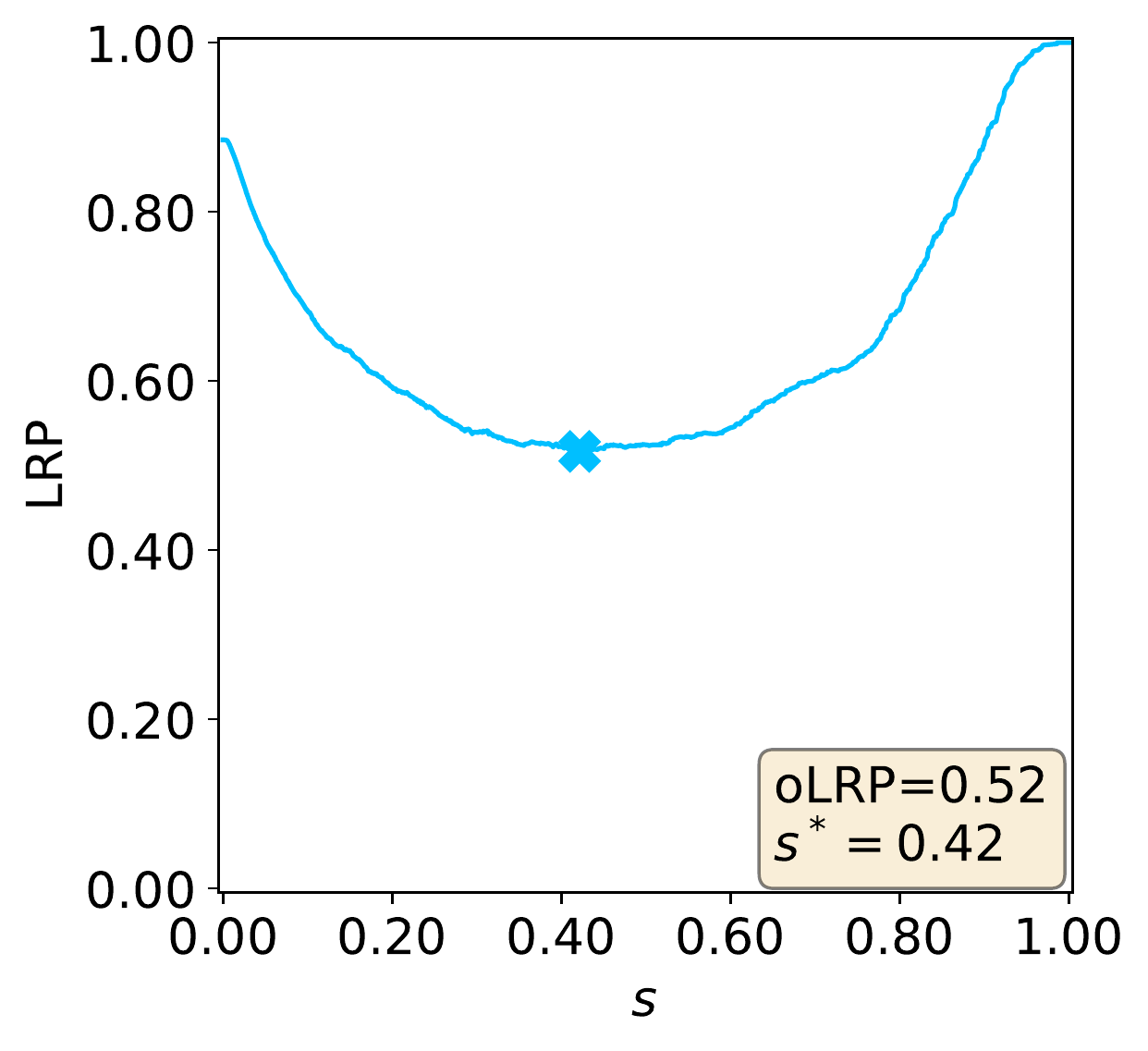}
        \caption{Bus}
        \end{subfigure}
        \caption{s-LRP curves of various classes for Mask R-CNN with ResNet-50 backbone on LVIS dataset.}
        \label{fig:LRPConfScoreCurves_LVIS}
\end{figure*}

\textbf{LVIS Dataset:} Using the s-LRP curves in \figref{\ref{fig:LRPConfScoreCurves_LVIS}}, one can observe the following:

\begin{itemize}
    \item While ``walking stick'' and ``weatherwane'' (c.f. (a) and (b)) have equal performance (i.e. $\mathrm{oLRP}=0.90$), ``walking stick'' is very sensitive to thresholding but ``weatherwane'' is not. In particular, if the threshold of ``walking stick'' is set slightly larger than $0.00$, then LRP Error increases to $1$ implying no true positives. On the other hand, this is not very critical for ``weatherwane'' since LRP Errors are similar for $s \in [0.00, 0.80]$.
    \item While ``bottle cap'' has better performance  than ``baseball bat'' (c.f. (c) and (d)), LRP-Optimal threshold of ``bottle cap'' is significantly lower, and besides ``bottle cap'' is more sensitive to thresholding. Hence, the performance superiority may not indicate neither less sensitivity to thresholding nor larger LRP-Optimal threshold.
    \item Since, unlike PR curves, we do not use interpolation in s-LRP curves (i.e. to demonstrate the performance as it is - c.f. practicality in Section 3 to see how PR curves are interpolated), s-LRP curves can include wiggles (c.f. (e), (f) and (g)).
    \item We provide s-LRP curves for ``zebra'' and ``bus'' classes (c.f. (h) and (i)) for LVIS dataset as well to allow comparison those with COCO dataset. Note that compared to Faster R-CNN in \figref{\ref{fig:LRPConfScoreCurves1}}, while the LRP-Optimal threshold of ``zebra'' is larger, that of ``bus'' is lower.
    \item Finally, note that, while ``sour cream'', ``cornice'', ``zebra'' and ``bus'' (c.f. (f)-(i)) have similar performance, their s-LRP curves are different, i.e. they have different sensitivity to threshold choice and different LRP-Optimal thresholds.
\end{itemize}

\renewcommand{\thesection}{H}
\section{The Repositories and Configurations of the Used Models in the Experiments}
\label{app:repos}
In order to ensure reproducibility and facilitate direct usage of our results, Table \ref{tbl:repositories} associates the models used in different tables in the paper with the repositories from which we downloaded the trained models. For rare cases, there may be multiple alternatives for the same backbone and number of trained epochs owing to different design choices (e.g. using a different layer normalisation such as group normalisation \cite{GroupNorm}). In such cases, one can infer the corresponding model by comparing COCO-style AP values in the corresponding repository and method.

\begin{table}
\setlength{\tabcolsep}{0.4em}
\caption{The repositories of models that we downloaded, evaluated and utilized for comparison.} 
 \centering
\begin{tabular}{|c|c|l|}
\hline
Table&Repository&Method\\\hline\hline
\multirow{24}{*}{Table 3}&\multirow{19}{*}{mmdetection \cite{mmdetection}}&ATSS \cite{ATSS}\\
& & Carafe \cite{carafe}\\
& & Cascade Mask R-CNN \cite{CascadeRCNN}\\
& & Cascade R-CNN \cite{CascadeRCNN}\\
& & DetectoRS \cite{detectors}\\
& & FCOS \cite{FCOS}\\
& & FreeAnchor \cite{FreeAnchor}\\
& & GHM \cite{gradientharmonizing}\\
& & Grid R-CNN \cite{GridRCNN}\\
& & GRoIE \cite{groie} \\
& & Guided Anchoring \cite{GuidedAnchoring}\\
& & Hybrid Task Cascade \cite{HTC}\\
& & Libra R-CNN \cite{LibraRCNN}\\
& & Mask R-CNN \cite{MaskRCNN}\\
& & Mask Scoring R-CNN \cite{MaskScoringRCNN}\\
& & NAS-FPN \cite{NASFPN}\\
& & PointRend \cite{PointRend}\\
& & RPDet\cite{RepPoints}\\
& & SSD \cite{SSD} \\\cline{2-3}
&\multirow{2}{*}{detectron \cite{Detectron2018}}& Faster R-CNN \cite{FasterRCNN}\\
& & RetinaNet\cite{FocalLoss}\\\cline{2-3}
&detectron2 \cite{Detectron2}& Keypoint R-CNN \cite{MaskRCNN}\\ \cline{2-3}
&official code \cite{aLRPLossRepo}&aLRP Loss \cite{aLRPLoss}\\ \hline
\multirow{2}{*}{Table 5}&\multirow{2}{*}{mmdetection \cite{mmdetection}}& Faster R-CNN \cite{FasterRCNN}\\
& & RetinaNet\cite{FocalLoss}\\\hline
Table 6&detectron2 \cite{Detectron2}&Panoptic FPN \cite{PanopticFPN}\\\hline
Table 7&mmdetection \cite{mmdetection}&Mask R-CNN \cite{MaskRCNN}\\\hline
Table 8&detectron2 \cite{Detectron2}&Mask R-CNN \cite{MaskRCNN}\\\hline
Table 9&tensorflow \cite{tensorflow}&Faster R-CNN \cite{FasterRCNN}\\\hline
\multirow{3}{*}{Table 10}&\multirow{3}{*}{mmdetection \cite{mmdetection}}& DC5 \cite{DeformableRoIPool}\\
& & FPN \cite{FeaturePyramidNetwork}\\
& & RFP \cite{detectors}\\\hline
Table 11&official code \cite{qpic_off}&QPIC \cite{qpic}\\\hline
Table 12&official code \cite{blc_off}&BLC \cite{BLC}\\\hline
Table S2&mmdetection \cite{mmdetection}&Mask R-CNN \cite{MaskRCNN}\\\hline
Table S3&mmdetection \cite{mmdetection}&Faster R-CNN \cite{FasterRCNN}\\\hline
\multirow{3}{*}{Table S4}&\multirow{3}{*}{mmdetection \cite{mmdetection}}& DETR \cite{detr}\\
& & Sparse R-CNN \cite{sparsercnn}\\
& & DDETR \cite{ddetr}\\\hline
Table S5&detectron2 \cite{Detectron2}&Panoptic FPN \cite{PanopticFPN}\\\hline
\end{tabular}
\label{tbl:repositories}
\end{table}


%

\begin{IEEEbiography}[{\includegraphics[width=1in,height=1.25in,clip,keepaspectratio]{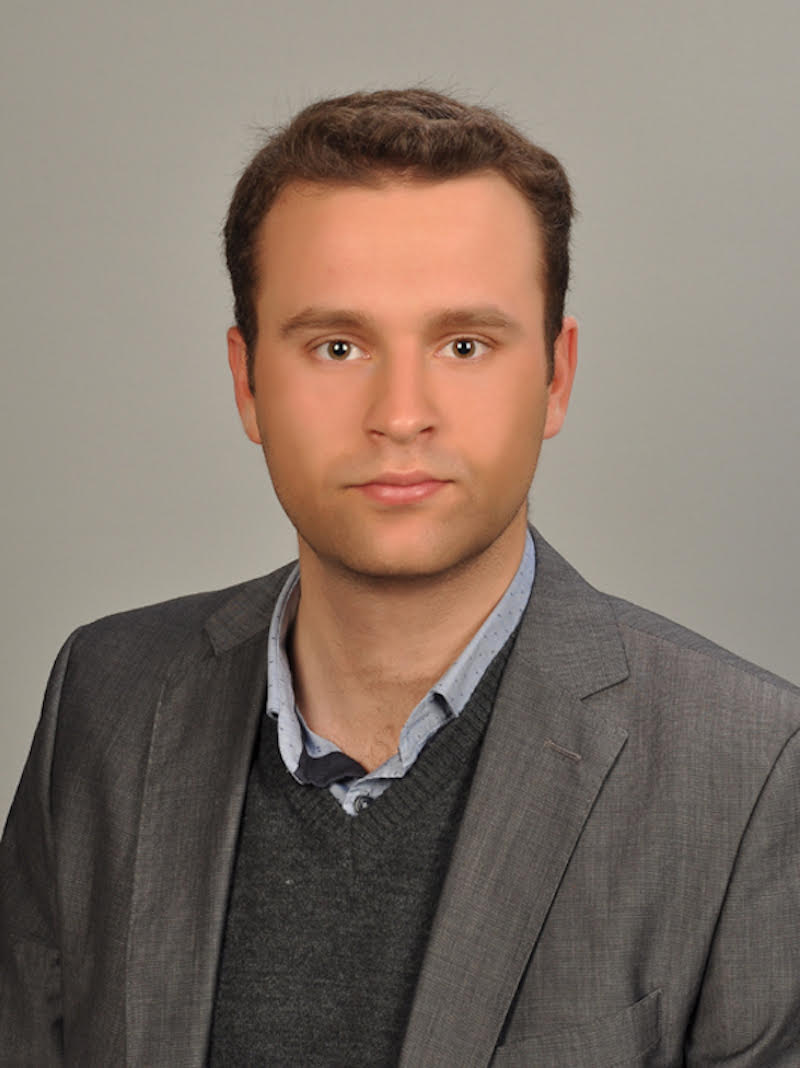}}]{Kemal Oksuz}
received his B.Sc. from the Land Forces Academy of Turkey in 2008 and his M.Sc. in 2016 from Bogazici University, Turkey. He received his Ph.D. degree in 2021 from the Dept. of Computer Eng., Middle East Technical University (METU), Turkey. Currently, he is working for the Turkish Armed Forces. He is also a research  associate at  ImageLab, METU. His research interests include computer vision with a focus on visual detection and imbalance problems.
\end{IEEEbiography}
\vspace{-1cm}
\begin{IEEEbiography}[{\includegraphics[width=1in,height=1.25in,clip,keepaspectratio]{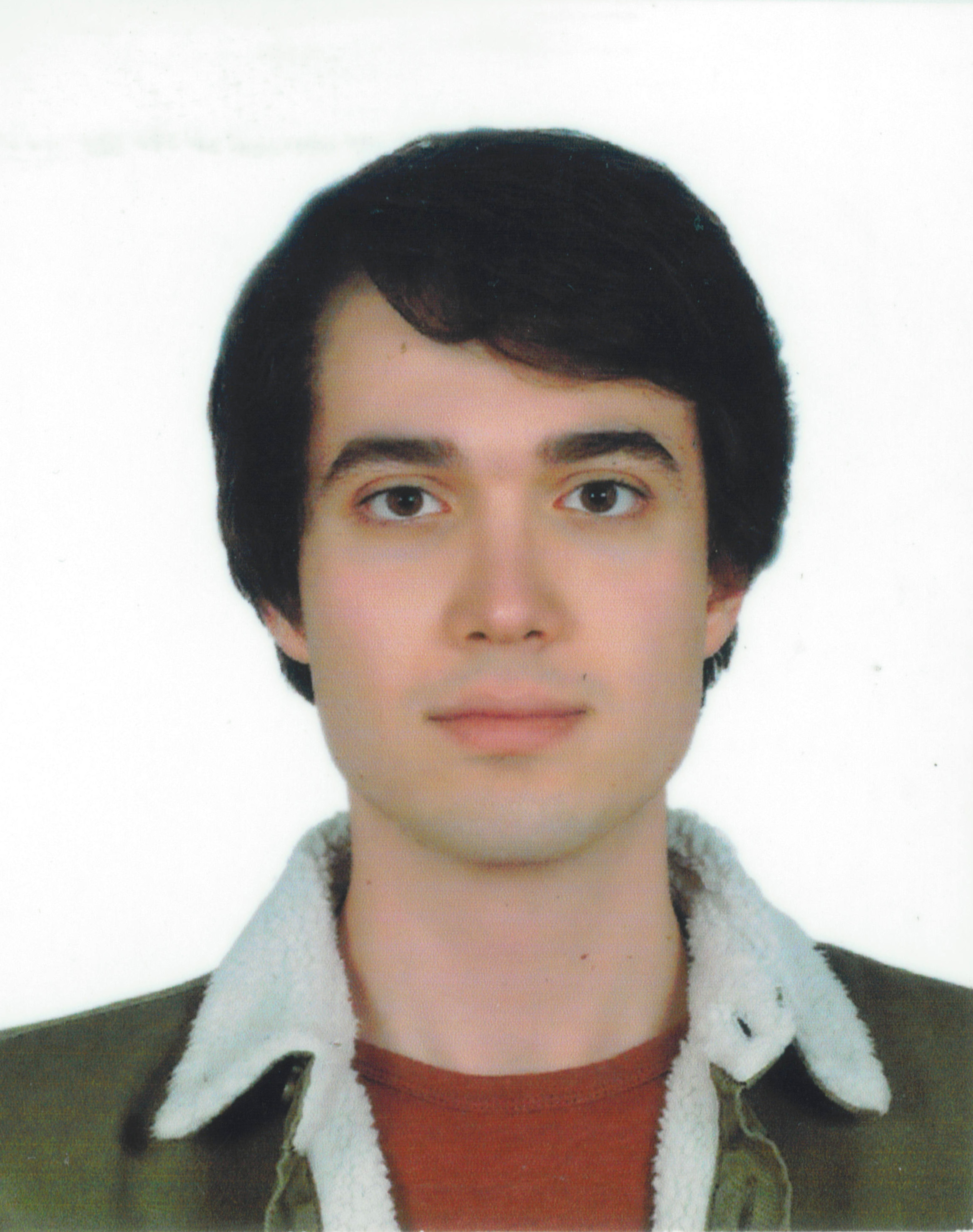}}]{Baris Can Cam}
received his B.Sc. degree in Electrical and Electronics Engineering from Eskisehir Osmangazi University, Turkey in 2016. He is currently pursuing his M.Sc. in Middle East Technical University, Ankara, Turkey. His research interests include computer vision with a focus on object detection.
\end{IEEEbiography}
\vspace{-1cm}

\begin{IEEEbiography}[{\includegraphics[width=1in,height=1.25in,clip,keepaspectratio]{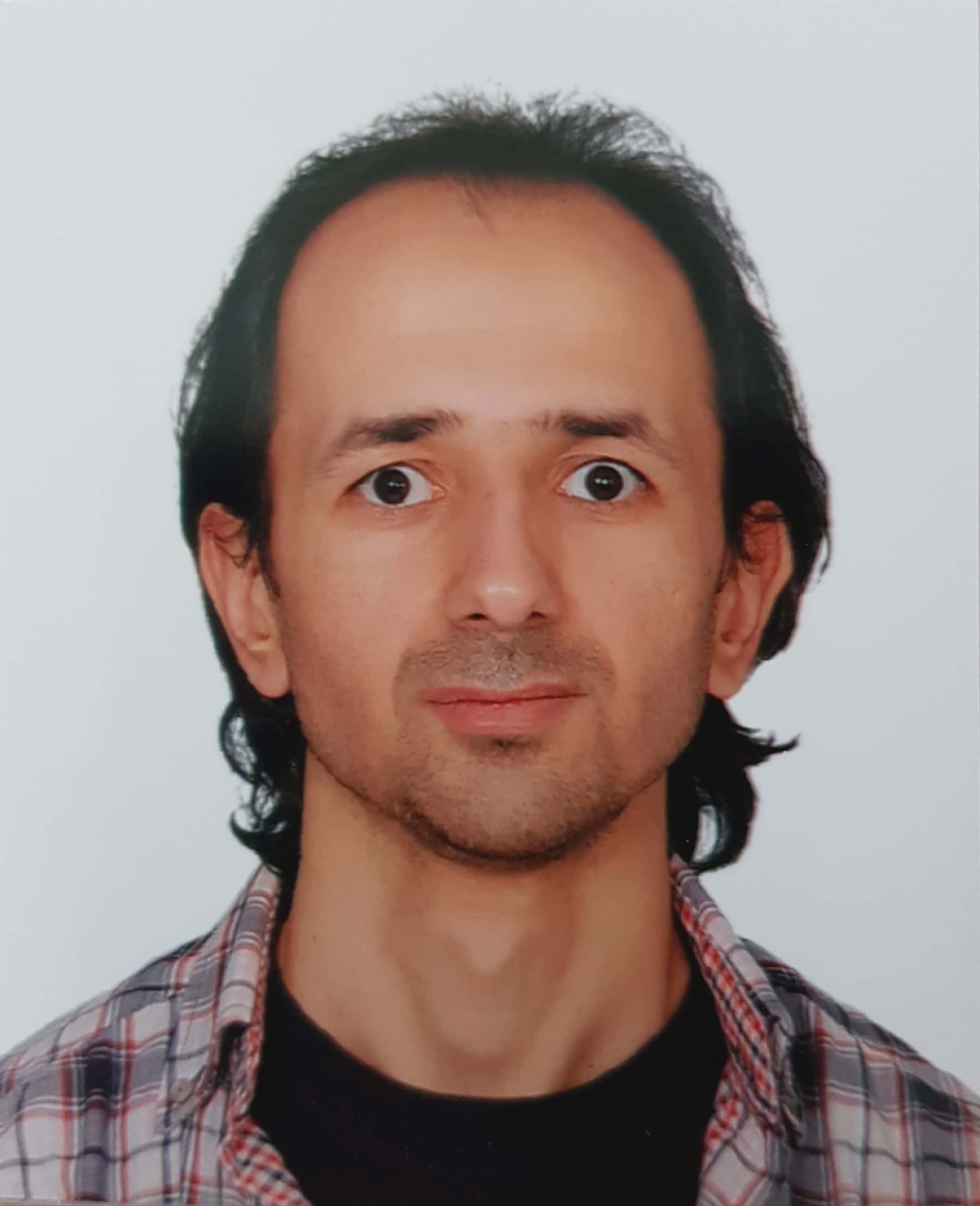}}]{Sinan Kalkan}
received his M.Sc. degree in Computer Eng. from Middle East Technical University (METU), Turkey in 2003, and his Ph.D. degree in Informatics from the Uni. of Göttingen, Germany in 2008. After working as a postdoctoral researcher at the Uni. of Göttingen and at METU, he joined METU in 2010 as a faculty member and since then, has been working as an assoc. prof. on problems within Computer Vision and Robotics.
\end{IEEEbiography}
\vspace{-1cm}
\begin{IEEEbiography}[{\includegraphics[width=1in,height=1.25in,clip,keepaspectratio]{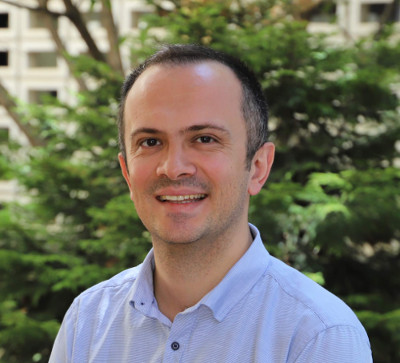}}]{Emre Akbas} is an asst. prof. at the Dept. of Computer Eng., METU, Turkey.
    He received his Ph.D. degree from the Dept. of Elect. and Comp.
    Eng., Uni. of Illinois at Urbana-Champaign in 2011. His M.Sc. and
    B.Sc. degrees in computer sci. are both from METU.
    Prior to joining METU, he
    was a postdoctoral researcher at the Dept. of Psychological and Brain
    Sciences, Uni. of California Santa Barbara.
    His research interests
    are in computer vision with a focus on object detection and human pose estimation.  
\end{IEEEbiography}





\end{document}